\documentclass{article}

\usepackage{microtype}
\usepackage{graphicx}
\usepackage{subcaption}
\usepackage{booktabs} 
\usepackage{siunitx}    

\usepackage{hyperref}
\usepackage{multirow}
\usepackage{natbib}
\usepackage{stfloats} 
\usepackage{enumitem}
\usepackage{fontawesome5}
\usepackage{multirow, stackengine, booktabs}

\usepackage[accepted]{icml2026}

\usepackage{amsmath}
\usepackage{amssymb}
\usepackage{mathtools}
\usepackage{amsthm}
\usepackage{graphicx}
\usepackage{subcaption}  
\usepackage{multirow}
\usepackage{xcolor}
\usepackage{tcolorbox}
\usepackage{makecell} 
\usepackage[capitalize,noabbrev]{cleveref}

\theoremstyle{plain}

\theoremstyle{definition}

\theoremstyle{remark}

\usepackage[textsize=tiny]{todonotes}

\icmltitlerunning{Interpretability Transfer from Language to Vision via Sparse Autoencoders}
\begin{document}

\twocolumn[
  \icmltitle{Interpretability Transfer from Language to Vision via Sparse Autoencoders}



  \icmlsetsymbol{corr}{\faEnvelope}

  \begin{icmlauthorlist}
    \icmlauthor{Alexey Kravets}{bath}
    \icmlauthor{Da Li}{samsung}
    \icmlauthor{Chuan Li}{lambda}
    \icmlauthor{Da Chen}{bath,corr}
    \icmlauthor{Vinay P. Namboodiri}{bath}
  \end{icmlauthorlist}

  \icmlaffiliation{bath}{University of Bath, UK}
  \icmlaffiliation{lambda}{Lambda, Inc.}
  \icmlaffiliation{samsung}{Samsung AI Centre Cambridge}

  \icmlcorrespondingauthor{Da Chen}{da.chen@bath.edu}

  \icmlkeywords{Machine Learning, ICML}

  \vskip 0.3in
]



\printAffiliationsAndNotice{}  

\begin{abstract}
  Recent advances in language model interpretability using sparse autoencoders (SAEs) have yet to effectively translate to the visual domain, mainly due to the difficulty and ambiguity of labeling visual concepts. In this paper, we introduce Visual Interpretability via SAE Transfer Alignment (VISTA), a framework that transfers interpretability from language to vision in a LLaVA-style vision-language model by constraining a visual projector to map visual tokens into an LLM's pre-existing, labeled textual SAE space. This approach enables visual interpretability without training dedicated vision SAEs. By regularizing the projector using the LLM's SAE reconstruction loss, VISTA achieves a threefold increase in the matching rate, which measures how accurately the most activating textual concepts in the SAE space correspond to semantic elements in the image. Using this framework, we further analyze spatial localization properties of different vision encoders and show that DINOv2 features have stronger localization abilities than other encoders. Leveraging this precision, we validate VISTA's cross-modal alignment through fine-grained, localized concept interventions, where specific objects are removed or replaced in the model's perception while preserving the surrounding scene. This results in improvements of 35\% in object removal and 47\% in object replacement tasks over vision-only baselines, providing causal evidence that visual tokens inhabit the text SAE manifold. These contributions are validated across multiple LLM architectures. Code is available on \href{https://github.com/akres001/Interpretability-Transfer-from-Language-to-Vision-via-Sparse-Autoencoders}{GitHub}.

\end{abstract}

\section{Introduction}

Interpreting internal representations of Large Multimodal Models (LMMs) remains a challenge despite their strong visual reasoning capabilities. While sparse autoencoders (SAEs) have successfully decomposed polysemantic language features into interpretable concepts \cite{bricken2023monosemanticity}, applying them to the visual modality is obstructed by limitations in the two primary solutions. The first solution involves training domain-specific SAEs on visual encoders \cite{lim2025patchsae,pach2025sparse,sjose_steeringclip} which yields monosemantic \textit{visual features} or \textit{visual concepts}. However, labeling these features is difficult and ambiguous. For instance, distinguishing whether a ``dog'' feature represents the dog concept, a more specific breed, or even an action may be hard. Furthermore, vision-only SAEs do not reveal how visual features are interpreted once they enter the LLM's residual stream. 
The second potential solution is to fine-tune existing LLM-based SAEs to incorporate visual information, but this approach incurs prohibitive computational costs. For example, Gemma Scope \cite{lieberum2024gemmascopeopensparse} required $\sim$20\% of GPT-3's total compute and would necessitate an extensive re-labeling effort (circa half a million LLM requests for Gemma-2-2B alone) to account for the shifted latent space.

To solve this, we ask the following question: \begin{tcolorbox}[colback=gray!15, colframe=gray!50, boxrule=0.3pt, arc=2pt, left=6pt, right=6pt, top=4pt, bottom=4pt]
\textbf{Q:} \textit{Can language sparse autoencoders be reused to interpret visual concepts?}
\end{tcolorbox}
In this paper, we answer this affirmatively through our solution \textbf{VISTA} (Visual Interpretability via SAE Transfer Alignment), that bypasses the need for new SAEs by leveraging existing, labeled feature space of text SAEs. Instead of training or fine-tuning SAEs, we introduce a training objective that forces the visual projector to map visual tokens directly into the LLM's residual stream, constrained such that they align with the semantic structure defined by the text SAEs. In this way, we achieve \textbf{interpretability transfer}, allowing visual inputs to be interpreted via pre-labeled linguistic SAE concepts with negligible additional computational cost.

VISTA is applied within the LLaVA paradigm \cite{liu_llava_2023}. A frozen vision encoder produces visual tokens, a trainable linear projector maps them into a frozen LLM's residual stream, and the LLM generates text conditioned on the resulting multimodal sequence. This design is grounded in the finding by \citet{limber} that linear mappings suffice for cross-modal alignment. VISTA augments this standard setup with a single addition of an SAE reconstruction loss constraining visual tokens to lie on the LLM's pre-trained text SAE manifold. All interpretations and interventions in this work operate on visual token representations within the LLM's residual stream.

Three metrics are applied to evaluate how well visual features align with the textual SAE space: reconstruction error, feature sparsity, and an LLM-verified matching rate \cite{venhoff2025how}. Reconstruction and sparsity dynamics (\cref{sec:recons_sparsity}) determine if the projected visual tokens match the sparse distribution and reconstructibility of the LLM's text features while matching rate measures how accurately the most activating text SAE concepts activating on the visual tokens correspond to the semantic elements visible in the image (\cref{sec:match_rate}). VISTA constraints lead the matching rate to increase more than a \textbf{threefold} on average.

\begin{figure*}[!t]
    \centering
    \includegraphics[width=0.7\linewidth]{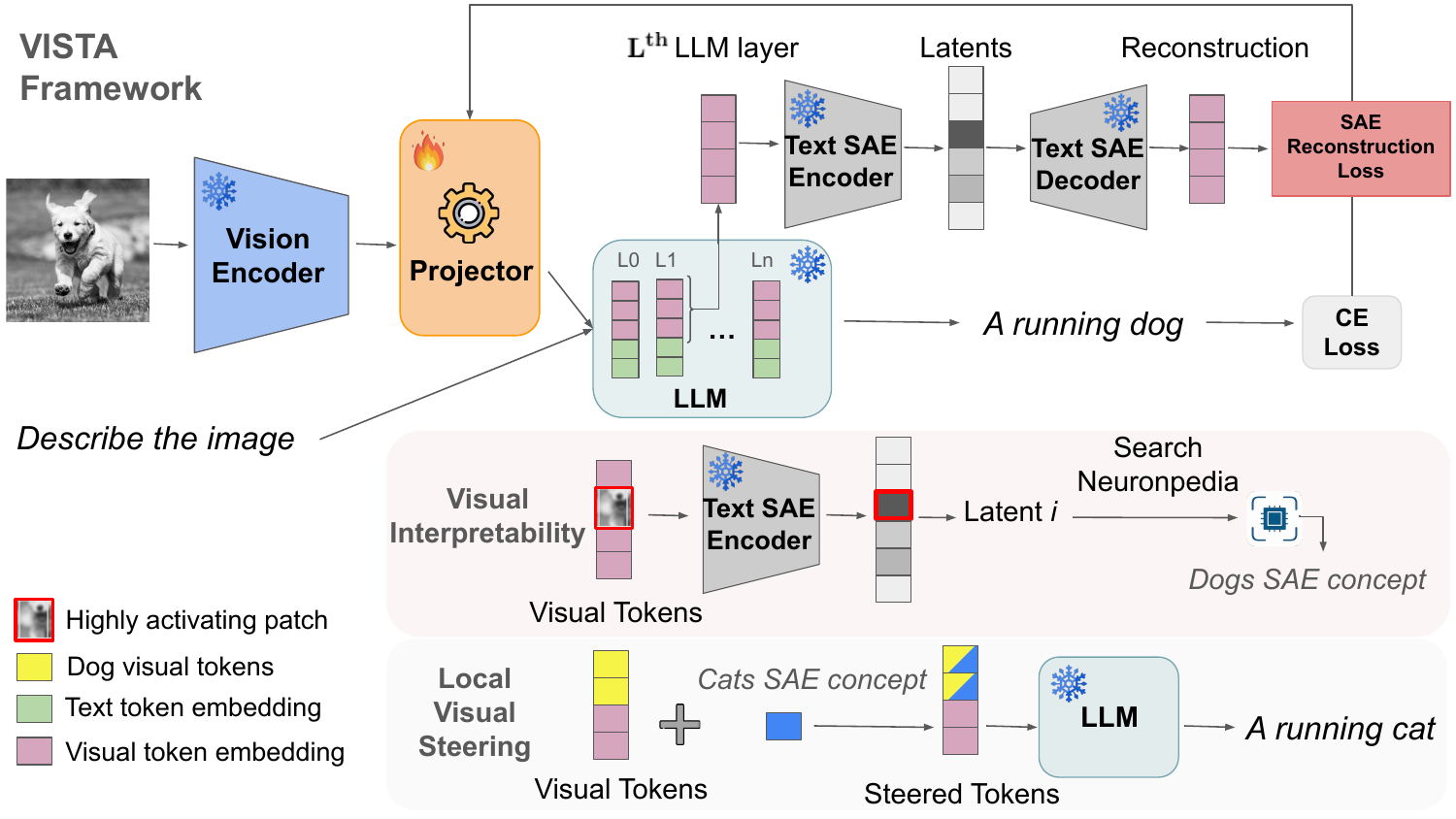}    
    \caption{\textbf{VISTA Framework.} \textbf{Training:} A trainable projector maps tokens from a frozen vision encoder to a frozen LLM, optimized via Cross-Entropy and an auxiliary SAE Reconstruction Loss to force alignment of the visual embeddings with the manifold defined by pre-trained Text SAEs. \textbf{Visual Interpretability:} Once aligned, highly activating visual tokens can be interpreted via the Text SAE. Activating latents are searched in Neuronpedia for their linguistic labels that correspond to the visual concept in the patch, such as ``Dog'' concept in the patch showing the dog. \textbf{Steering:} By manipulating specific SAE concepts in localized visual tokens, model's output can be precisely steered (changing ``A running dog'' to ``A running cat'') preserving global context.}
    \label{fig:mainfig}
\end{figure*}

Unlike previous beliefs \cite{venhoff2025how}, we find that ``delayed alignment'', a phenomenon where visual tokens initially diverge from the text SAE sparse linguistic manifold, showing high reconstruction error and low feature sparsity, and only naturally converge toward the model's text-based feature space in deeper layers is not a universal characteristic of all LLM architectures. While visual tokens in Gemma SAEs eventually align with the textual patterns in later layers, this natural alignment does not occur for the LLaMA-3.1-8B-Instruct SAEs. By applying our SAE constraints during training, alignment is achieved immediately across all architectures (\cref{sec:recons_sparsity}). 

Furthermore, we show that precise local interpretability and steering depend heavily on the choice of the vision encoder (\cref{sec:spatial_loc}). Local interpretability requires that activated SAE concepts are spatially aligned with specific image patches. For example, a patch covering the background might incorrectly activate a concept like ``cat'' (\cref{fig:spatial_location} bottom-left) that actually belongs to an object elsewhere in the scene. Our analysis reveals that while contrastive models like CLIP \cite{radford2021LearningTV} prioritize global summaries and lead to spatial confusion, self-supervised models like DINOv2 \cite{oquab2024dinov} preserve the fine-grained local information required for patch-level accuracy.
Specifically, DINOv2 maintains robust spatial localization accuracy across different LLM architectures, consistently outperforming CLIP, whose localization quality varies depending on the LLM backbone but remains substantially lower. This spatial fidelity is a key requirement for fine-grained interpretability, as it allows concept activations to be reliably localized to the correct image regions. The same property enables the localized causal interventions demonstrated in \cref{sec:applic_steering}. Our \textbf{contributions} are:
\begin{itemize}
    \item We introduce a training objective that aligns visual tokens with LLM SAE features, facilitating cross-modal interpretability transfer without the computational burden of training or labeling vision-specific SAEs.
    \item We provide a rigorous comparison of different visual encoders enabled by VISTA's interpretability transfer, revealing that DINOv2 achieves superior and architecture-stable spatial localization, maintaining high accuracy across all tested LLM backbones, whereas CLIP's localization accuracy is lower and varies substantially with the LLM. By tracing exactly which patch triggers which semantic SAE concept, VISTA exposes encoder-level spatial fidelity that prior interpretability methods could not measure.
    \item As a causal validation of our cross-modal alignment, we demonstrate that arithmetic interventions on visual tokens using text-SAE directions produce coherent, localized changes in model output, achieving \textbf{35\%} improvement in object removal and \textbf{47\%} in replacement tasks over best vision-only baselines.
\end{itemize}

\section{Related Work}

Sparse autoencoders address polysemanticity \cite{elhage2022superposition} by decomposing activations into monosemantic features \cite{bricken2023monosemanticity, huben2024sparse}. Many works in the field of mechanistic interpretability use this decomposition as a basis for further investigation, such as the causal role these features play in downstream tasks \cite{marks2025sparse}, the universality of features across different model scales \cite{wang2025towards}, and the potential for feature steering to mitigate hallucinations \cite{abdaljalil-etal-2025-safe}. While foundational for LLMs, visual applications are emerging. Recent work has applied SAEs to CLIP encoders to identify interpretable features \cite{pach2025sparse, sjose_steeringclip, daujotas_2023_case}, but these primarily utilize global [CLS] tokens, limiting their ability to explain or steer localized objects. While \citet{lim2025patchsae}  trained SAEs on patches for attribution, their method is primarily designed for the explanation of classification decisions rather than for the steering or control of model behavior.

The difficulty in labeling visual features suggests that a more effective path to interpretability lies in cross-modal alignment with the linguistic domain. While traditional alignment relies on contrastive losses, recent work shows that linear mappings often suffice. \citet{limber} showed that a linear projection can map frozen vision encoders to frozen LLMs. \citet{venhoff2025how} found that visual features naturally align with language features in deeper LLM layers. We extend this by enforcing alignment via SAE reconstruction. By constraining the projector to output vectors reconstructible by text SAEs, we achieve ``interpretability transfer'' leveraging the LLM's semantic structure to interpret visual tokens without training separate vision-based models.

This alignment enables more effective activation steering and control. Building on representation engineering, where model behavior is modulated by adding steering vectors to internal activations \cite{rimsky-etal-2024-steering, turner_2023_activation}, it has been shown \cite{pach2025sparse} that SAE-based directions to steer VLMs outperform traditional Difference-in-Means \cite{diffinmean} steering in preserving scene integrity. We demonstrate that by enforcing SAE constraints on the visual projector in VLMs, we can perform high-fidelity steering directly within the LLM's residual stream outperforming vision encoder based VLMs steering.

\section{Method}

In this paper, we introduce VISTA (Visual Interpretability via SAE Transfer Alignment), a framework that maps visual tokens into the pre-existing, labeled SAE feature space of a frozen LLM. By adding an SAE reconstruction loss to the training objective, we force visual features to lie in the text SAE manifold. This enables the analysis and steering of visual concepts using textual descriptors, bypassing the need for vision-specific SAE training or the fine-tuning of existing SAEs. This section details the VISTA training objective, the localized steering mechanism enabled by this alignment, and the metrics used to evaluate the precision of this transfer.

\subsection{VISTA}

To disentangle polysemantic LLM representations \cite{elhage2022superposition}, SAEs use an encoder-decoder architecture to map activations $x$ into a high-dimensional, sparse latent space $E(x) \in \mathbb{R}^{d_{SAE}}$ \cite{bricken2023monosemanticity}, where $d_{SAE}$ is typically 10 to 100 times the size of the original input. The model is trained to generate a reconstruction $\hat{x}$ by linearly combining a small subset of features in $E(x)$. By minimizing the error $\|x - \hat{x}\|_2^2$ alongside an $L_1$ sparsity penalty, the SAE is forced to find a near-perfect reconstruction using only a few human-interpretable ``concepts''. See \cref{appendix:sae_details} for the full mathematical derivation.

As depicted in \cref{fig:mainfig}, our VLM architecture follows the LLaVA paradigm \cite{liu_llava_2023} augmented by the VISTA alignment constraint. Let $V$ be a frozen vision encoder (e.g., DINOv2) and $L$ be a frozen language model (e.g., Gemma). We introduce a learnable linear projector $P_\theta$. Given an image $I$, the vision encoder produces visual tokens $v = V(I) \in \mathbb{R}^{N \times D_v}$ where $N$ is the number of visual tokens and $D_v$ is their embedding dimension. $P_\theta$ maps these to the LLM's embedding space:
\begin{equation}
    e_v = P_\theta(v) \in \mathbb{R}^{N \times D_{llm}}
\end{equation}
These visual embeddings are concatenated with text embeddings $e_t$ corresponding to the user input (e.g., a question or caption prompt) and passed to the LLM.

Traditionally, the projector $P_\theta$ is trained solely by minimizing the cross-entropy loss ($\mathcal{L}_{CE}$) on the target text sequences, such as answers or image descriptions \cite{venhoff2025how}. We propose augmenting this with a constraint derived from a pre-trained LLM sparse autoencoder. Let $SAE$ be an SAE trained on the residual stream (or activations) of the language model $L$ using only text data. The SAE consists of an encoder and a decoder, such that $\hat{x} \approx SAE(x)$. We hypothesize that if visual tokens $e_v$ are to be interpretable to the LLM, they should be representable by the sparse features learned from text. To ensure visual tokens are interpretable within this linguistic framework, we define the SAE Reconstruction Loss ($\mathcal{L}_{SAE}$) as the Mean Squared Error between the projected visual tokens and their reconstruction by the text SAE:
\begin{equation}
    \mathcal{L}_{SAE} = \frac{1}{|M|} \sum_{l \in M} || SAE_l(e_{v}^l) - e_{v}^l ||^2_2
\end{equation}
where $e_{v}^l$ denote the hidden state of the visual tokens at layer $l$ of the frozen LLM and $M$ is the set of layers at which the SAE reconstruction loss is applied to the visual tokens. 
The total training objective becomes:
\begin{equation}
    \mathcal{L}_{total} = \mathcal{L}_{CE} + \mathcal{L}_{SAE}
\end{equation}
Minimizing this joint loss forces the projector to satisfy both the functional requirements of the multimodal task and visual token reconstruction via textual SAEs. We train the visual projector in two stages keeping the vision encoder and LLM frozen, adopting the training setup from \citet{venhoff2025how}. Pre-training uses 595k image-text pairs from Conceptual Captions \cite{sharma2018conceptual}, and the fine-tuning stage uses 665k examples from the LLaVA-1.5 pipeline \cite{liu_llava_2023}. Additional details are in \cref{appendix:implm_details}.

\subsection{Fine-grained Visual Steering Mechanism}
The VISTA alignment forces visual tokens to inhabit the same latent space as text. This transforms the task of visual manipulation into a direct arithmetic operation within the linguistic manifold. As the training objective ensures visual tokens are represented as sparse combinations of pre-existing textual SAE features, we can modify the model's perception of specific token positions $v_{\text{source}}$ by adding or subtracting the decoder directions (concepts) of the text SAE. This allows us to apply removal and replacement operations to specific visual tokens.

\paragraph{Concept Removal and Replacement} For source patches $v_{\text{source}}$, removal subtracts a concept direction $v_{\text{remove}}$ to suppress it, while replacement adds a target direction $v_{\text{add}}$ to overwrite the local semantics:
\begin{equation}
    v_{\text{steered}} = v_{\text{source}} - \beta \cdot v_{\text{remove}},
    \qquad
    v_{\text{steered}} = v_{\text{source}} + \alpha \cdot v_{\text{add}}.
\end{equation}
The non-negative scalars $\alpha$ and $\beta$ control intensity, and restricting updates to selected token positions yields localized steering that leaves the rest of the scene intact.

\subsection{Spatial Localization and Evaluation Framework}
\label{sec:spatial_loc_framework}

We evaluate SAE-constrained alignment through multiple metrics. We first conduct a spatial localization accuracy analysis to verify if visual tokens activating specific semantic latents correspond to the object's physical location in the image. We then analyze reconstruction and sparsity to determine if projected visual tokens match the sparse distribution and reconstructibility of the LLM's text features. Finally, we utilize an LLM-verified matching rate, employing an oracle LLM to confirm that activated concepts on the visual features correspond to semantic concepts present in the scene. Formal definitions for sparsity, reconstructibility and matching rate are provided in \cref{sec:res_findings}.

\textbf{Spatial Localization Accuracy} To evaluate the precision of the aligned visual tokens, we define \textit{Spatial Localization Accuracy (SLA)}. This metric verifies whether the visual token that maximally activates a specific SAE latent is spatially located at the corresponding object in the image. The evaluation proceeds in two stages: we first confirm that an activated SAE latent corresponds to a recognizable object in the image, then verify that the visual token activating it covers that object. We use a subset of images sampled from COCO dataset \cite{lin2015microsoft}. For each image, we identify the three latents with the highest activations and record their visual tokens for every layer. These latent descriptions are retrieved from Neuronpedia\footnote{https://www.neuronpedia.org/}. We employ an oracle LLM to retain only latents whose descriptions correspond to localizable objects in the image, filtering out both irrelevant concepts and abstract or broad descriptions (e.g., ``building structure and layout'') that do not match specific COCO classes. Using COCO ground-truth bounding boxes, we then verify whether the visual token associated with each matched concept falls within the corresponding object's bounding box.

\section{Findings \& Results}
\label{sec:res_findings}

\subsection{How do VISTA Constraints Affect Performance?}
\label{sec:performance}

We evaluate the impact of SAE constraints on model performance across standard benchmarks, such as MME \cite{fu2025mme}, GQA \cite{hudson2019gqa}, POPE \cite{pope_bench}, and LLaVA-Bench \cite{liu_llava_2023}. In smaller models like Gemma-2-2B-it, we generally observe a slight performance regression, though DINOv2 maintains or slightly improves performance on specific benchmarks like LLaVA-Bench and POPE. On the other hand, as model scale increases, the impact of these interpretability constraints becomes mixed. For the LLaMA-3.1-8B-Instruct architecture, the SAE constraint acts as a beneficial regularizer, leading to improved performance on the majority of benchmarks for every visual encoder. Conversely, the Gemma-2-9B-it backbone shows encoder-dependent behavior where DINOv2 maintains competitive performance both with and without the SAE constraints, whereas CLIP and I-JEPA show varying degrees of performance degradation when the SAE constraints are introduced. Performance tables are provided in \cref{appendix:performance}. We determine the optimal number of SAE-constrained layers based on a performance-interpretability trade-off analysis, detailed in \cref{appendix:perf_interp}.

\subsection{Spatial Localization Accuracy: CLIP vs. DINOv2 vs. I-JEPA}
\label{sec:spatial_loc}

\begin{figure}[!t]
    \centering
    \small
    \begin{minipage}{0.04\columnwidth} \quad \end{minipage} 
    \begin{minipage}{0.34\columnwidth} \centering \textbf{Cat} \end{minipage}
    \hspace{5pt}
    \begin{minipage}{0.34\columnwidth} \centering \textbf{Cookie} \end{minipage}

    \vspace{2pt}

    \begin{tabular}{cll}
        \rotatebox{90}{\scriptsize \textbf{DINOv2}} &
        \begin{subfigure}[c]{0.35\columnwidth}
            \centering
            \includegraphics[width=0.9\linewidth]{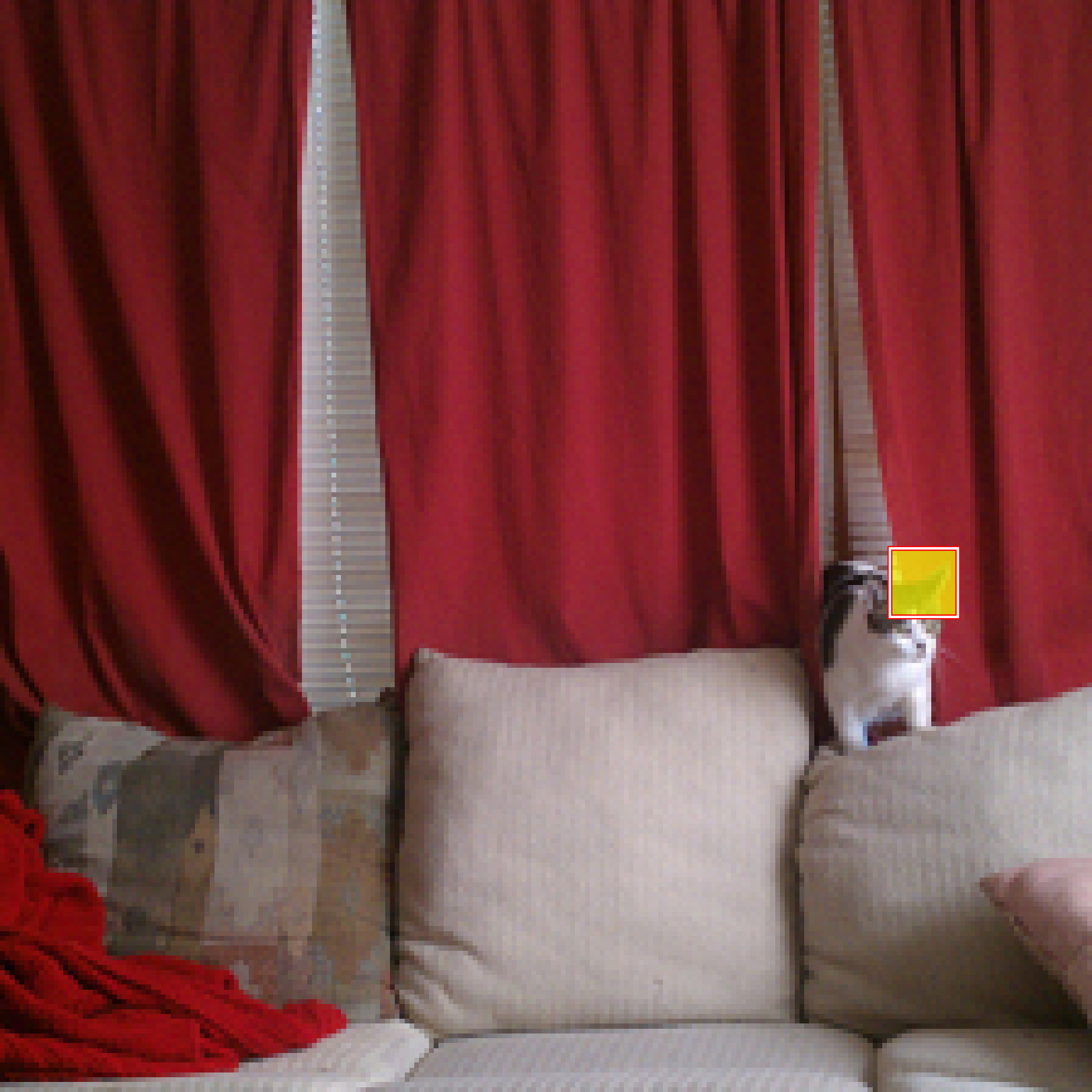}
        \end{subfigure} &
        \begin{subfigure}[c]{0.35\columnwidth}
            \centering
            \includegraphics[width=0.9\linewidth]{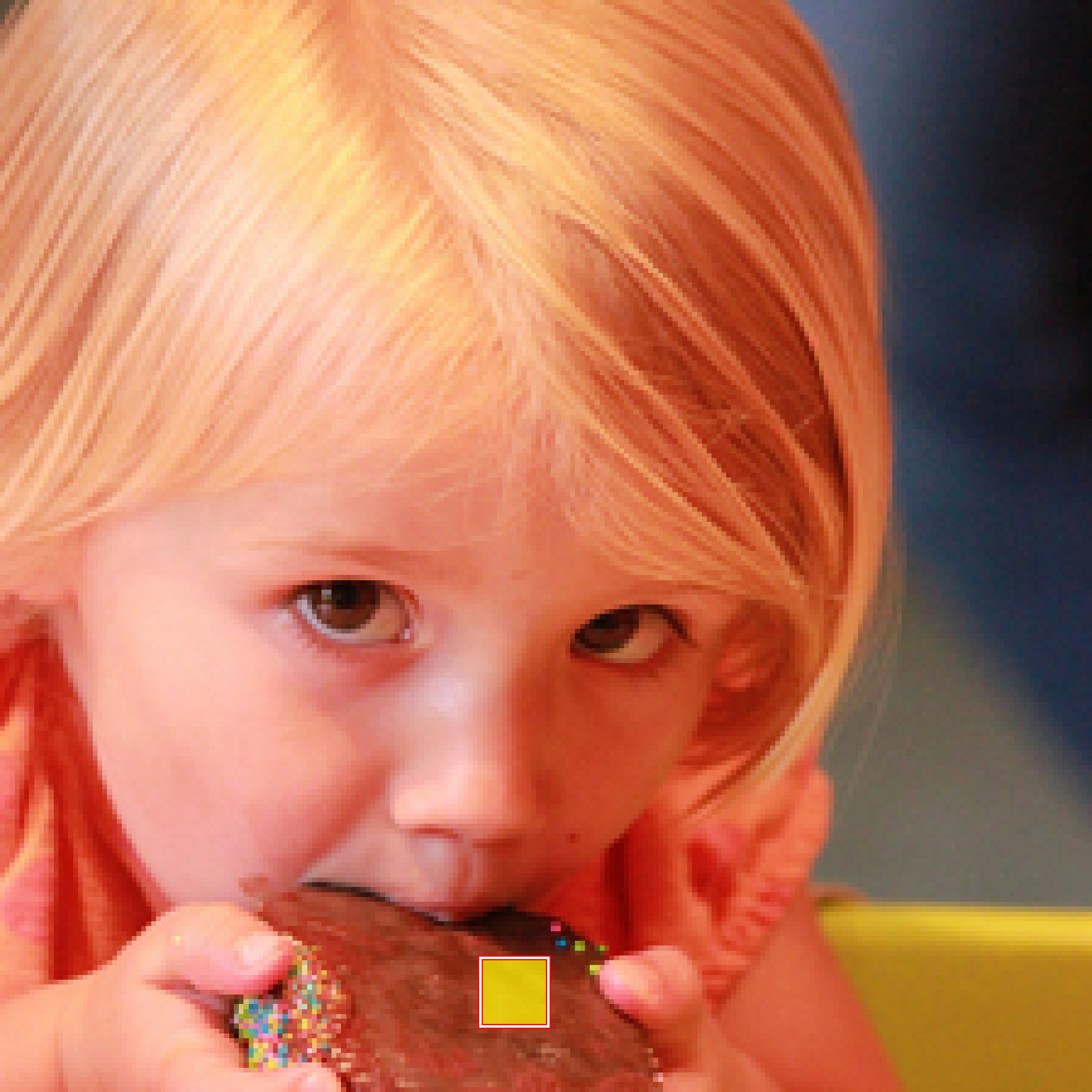}
        \end{subfigure} \\
        
        \addlinespace[2pt] 

        \rotatebox{90}{\scriptsize \textbf{CLIP}} &
        \begin{subfigure}[c]{0.35\columnwidth}
            \centering
            \includegraphics[width=0.9\linewidth]{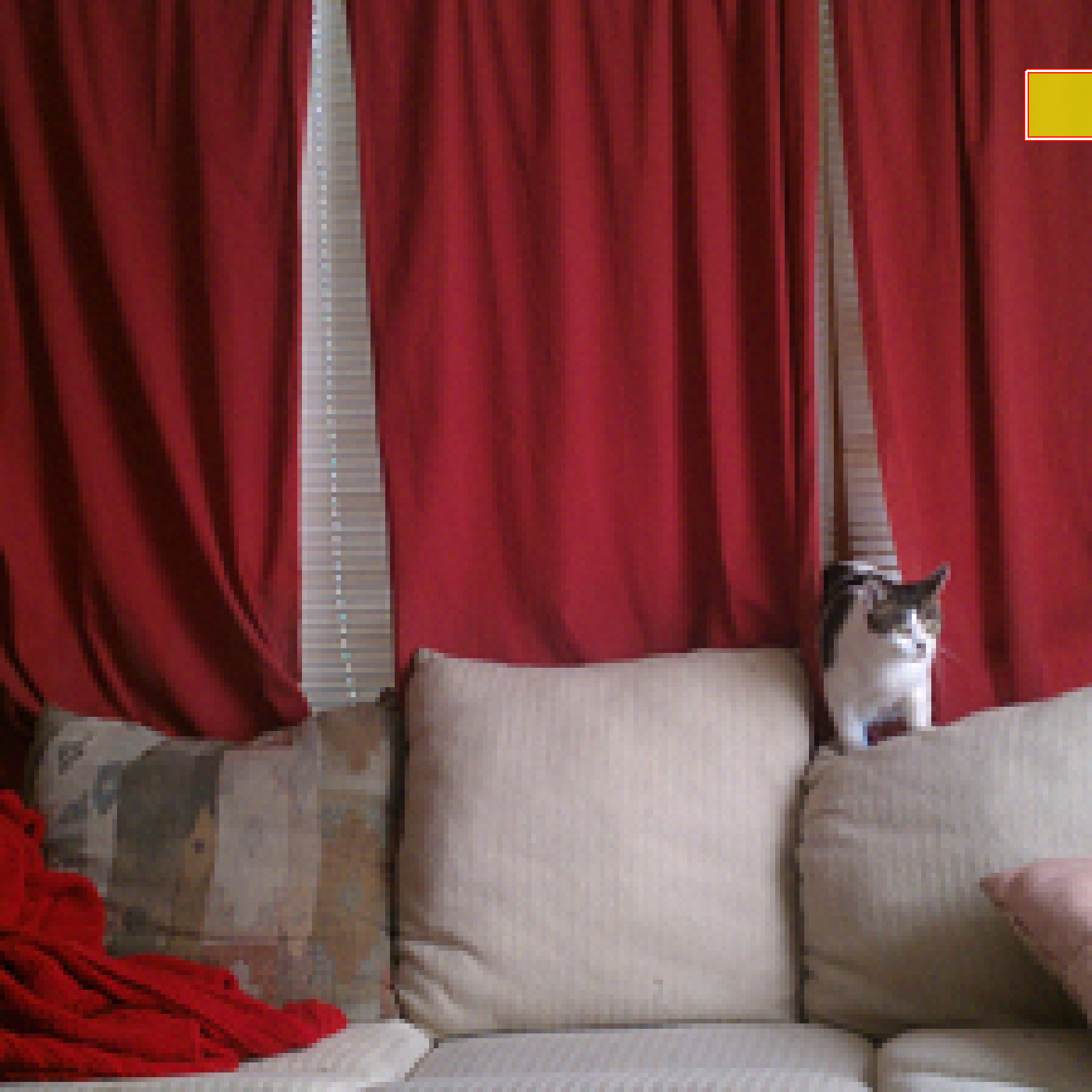}
        \end{subfigure} &
        \begin{subfigure}[c]{0.35\columnwidth}
            \centering
            \includegraphics[width=0.9\linewidth]{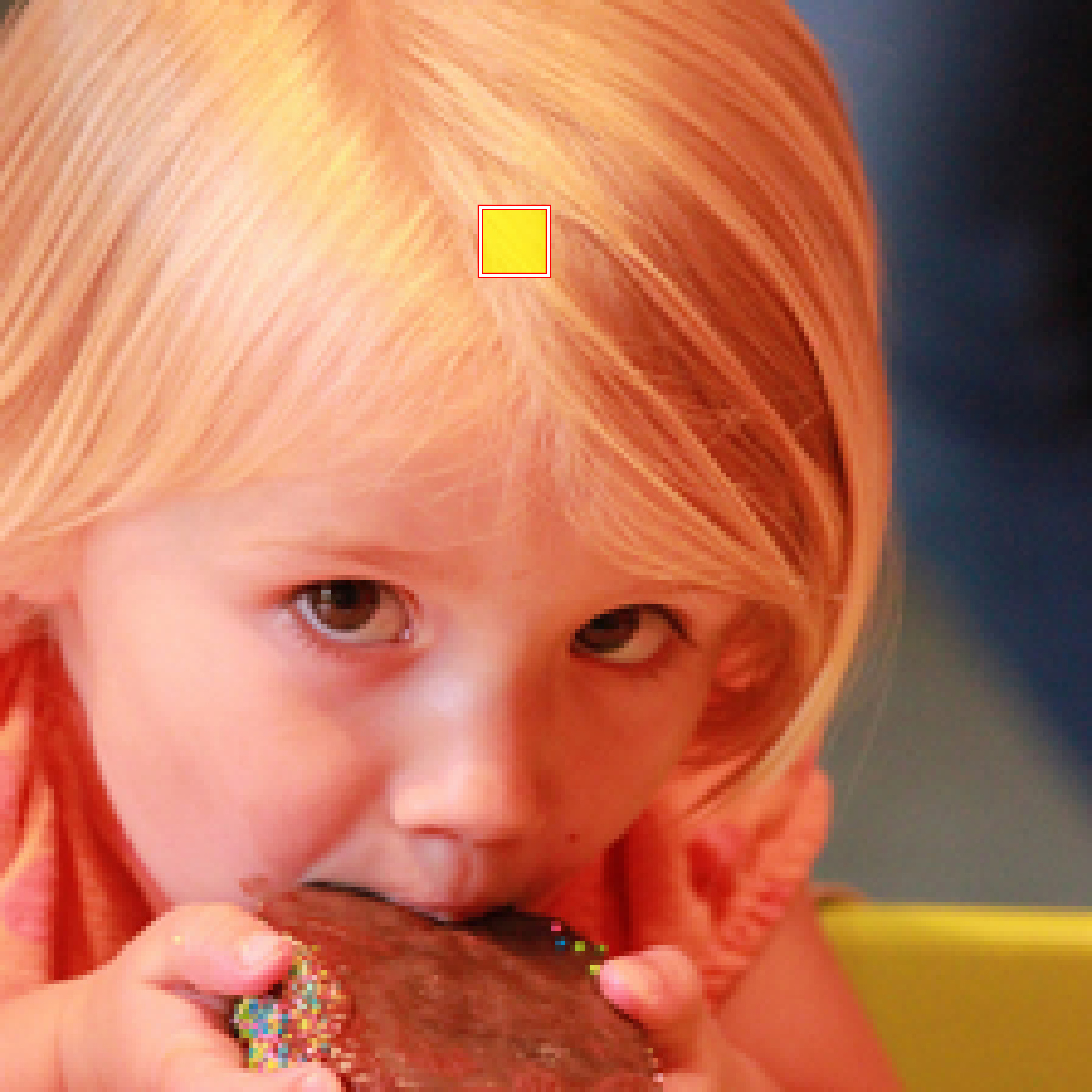}
        \end{subfigure}
    \end{tabular}

    \caption{\textbf{Spatial Location Analysis} of activating tokens. \textbf{Top Row:} DINOv2 precisely localizes visual concepts (cat and cookie). \textbf{Bottom Row:} CLIP activations exhibit spatial confusion, often activating tokens far from the relevant object. This high spatial fidelity in DINOv2 is a prerequisite for localized concept steering.}
    \label{fig:spatial_location}
\end{figure}

We observe a large disparity in spatial activation patterns across vision encoders, suggesting that VLM interpretability is primarily influenced by the visual backbone, but also depends on the LLM architecture used for alignment.
As shown in \cref{tab:localization_accuracy}, for the Gemma-2-2B-it backbone, DINOv2 achieves superior spatial fidelity with a localization accuracy of 0.927, whereas I-JEPA and CLIP have scores of 0.702 and 0.201, respectively.

DINOv2 shows good precision; visual tokens at specific indices correspond to the semantic content of their patches. For example, as shown in \cref{fig:spatial_location} (top-left) the highlighted visual token specifically activates the ``cat'' feature, which aligns perfectly with the cat's physical position in the image. This high level of spatial preservation likely stems from DINOv2's local-to-global discriminative objective, which forces the model to encode each patch's local content rather than a global summary. This patch-level locality is what enables our localized steering, allowing for arithmetic interventions on specific patch vectors without triggering unintended global hallucinations.

CLIP exhibits weaker spatial localization compared to DINOv2. Its global contrastive objective incentivizes the model to discard spatial details that do not contribute to global alignment, leading to patch-level feature collapse. Qualitatively, CLIP activations do not preserve patch location; as illustrated in \cref{fig:spatial_location} (bottom-left) the highlighted visual token activates the ``cat'' feature even though the cat is physically located in different patches. I-JEPA occupies a middle ground, retaining more spatial structure than CLIP but failing to achieve the precise feature-to-patch alignment of DINOv2. Its predictive objective requires capturing high-level scene context, causing local features to be slightly smeared to remain consistent with their neighbours.

We further validate this by ablating spatial localization accuracy across different LLM backbones in \cref{appendix:spatial_loc_abl}. We observe that CLIP’s localization accuracy modestly improves with LLaMA-3.1-8B and more noticeably with Gemma-9B, suggesting that cross-modal alignment quality partially depends on the language model size as well. However, DINOv2 significantly outperforms CLIP across all tested configurations, with I-JEPA also achieving higher localization accuracy than CLIP.

\begin{table}[!t]
\centering
\small
\setlength{\tabcolsep}{6pt} 
\caption{\textbf{Spatial Localization Accuracy Scores} (Gemma-2-2B-it backbone). The score reflects the proportion of top-activating tokens that fall within the ground-truth bounding box of the corresponding concept. DINOv2 demonstrates the highest alignment.}
\label{tab:localization_accuracy}
\begin{tabular}{lccc}
\toprule
\textbf{Encoder}    & CLIP  & I-JEPA & DINOv2 \\
\midrule
\textbf{SLA} & 0.201 & 0.702  & 0.927  \\
\bottomrule
\end{tabular}
\end{table}

\subsection{Reconstruction and Sparsity Dynamics}
\label{sec:recons_sparsity}

To quantify the alignment between the projected visual tokens and the LLM's internal feature space, we follow \citet{venhoff2025how} using two primary metrics. \textit{SAE Reconstruction Error} measures how accurately visual representations can be reconstructed by the text SAE. For a given layer $l$, we compute the mean squared error between the visual token $e_{v,i}^l$ and its reconstruction:
\begin{equation}
    \mathrm{Err}_l = \frac{1}{N} \sum_{i=1}^{N} ||e_{v,i}^l - \text{SAE}_l(e_{v,i}^l)||_2^2
\end{equation}
where $N$ is the number of visual tokens and $\text{SAE}_l$ represents the SAE at layer $l$.  Then, we analyze \textit{SAE Feature Sparsity}, defined as the average fraction of language features activated by the visual tokens. Let $E_l(x)$ (see \cref{appendix:sae_details}) denote the SAE encoder at layer $l$. Sparsity $S_l$ is calculated as:
\begin{equation}
S_l = \frac{1}{d_{SAE}} \frac{1}{N} \sum_{i=1}^{N} ||E_l(e_{v,i}^l)||_0
\end{equation}
where $||\cdot||_0$ is the $L_0$ norm and $d_{SAE}$ is the dictionary size. As textual activations are sparse \citep{bricken2023monosemanticity}, we hypothesize that visual tokens should exhibit similar sparsity.

\begin{figure}[!t]
    \centering
    \begin{subfigure}[b]{0.5\textwidth}
        \centering
        \includegraphics[width=\linewidth]{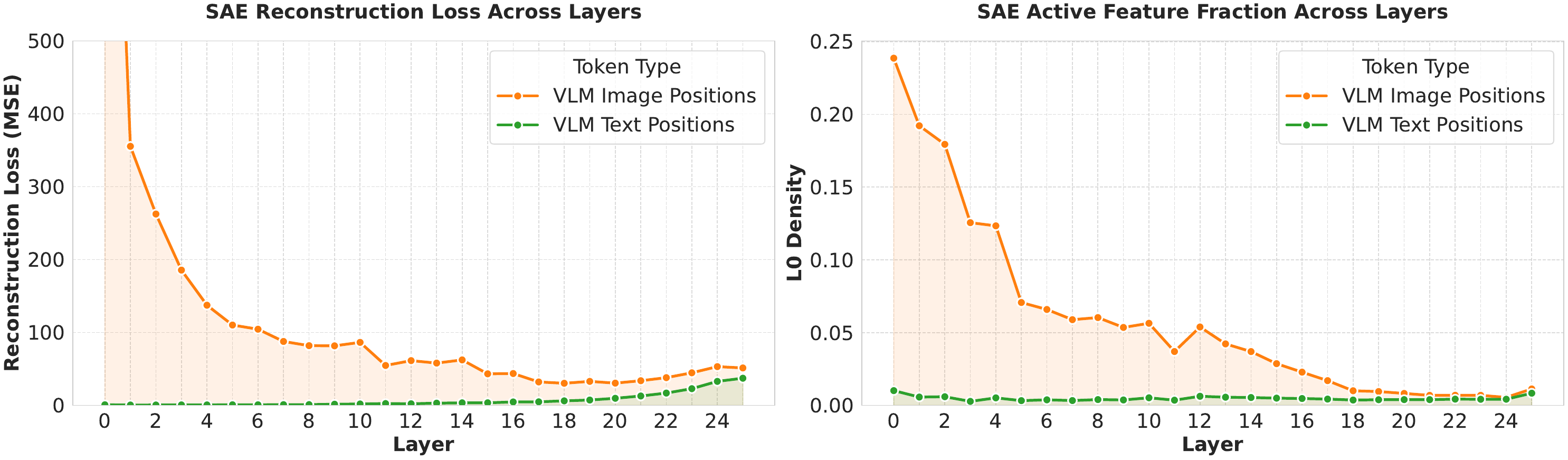}
        \caption{Without SAE constraints}
        \label{fig:reconstr_sparsity_dinov2_nosae}
    \end{subfigure}
    \hfill
    \begin{subfigure}[b]{0.5\textwidth}
        \centering
        \includegraphics[width=\linewidth]{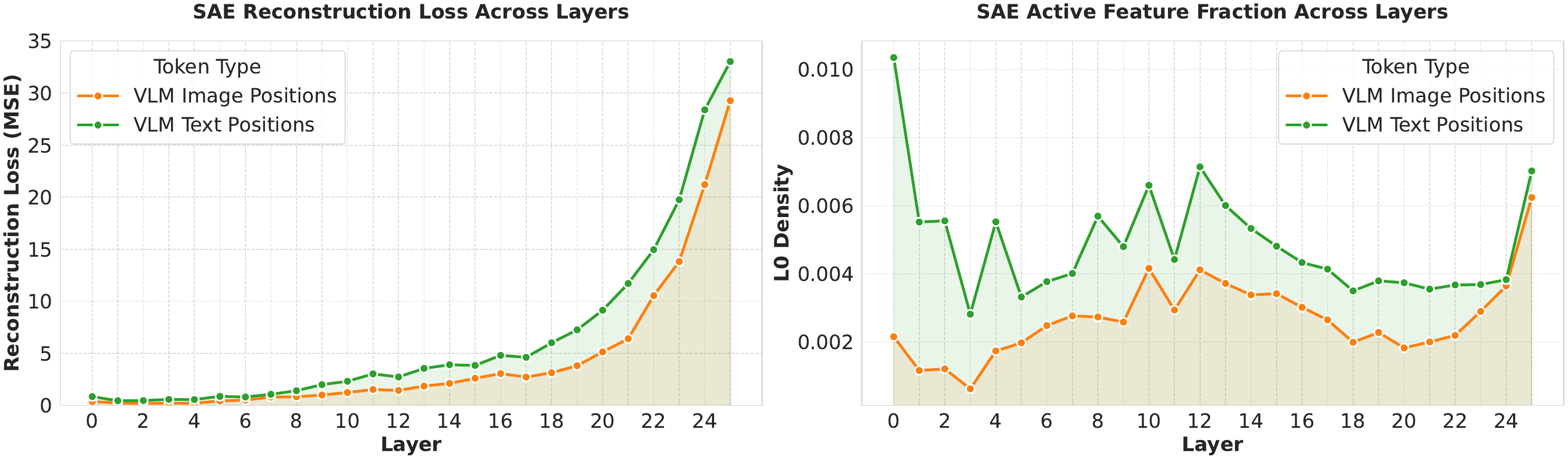}
        \caption{With SAE constraints}
        \label{fig:reconstr_sparsity_dinov2_wsae}
    \end{subfigure}
    \caption{\textbf{DINOv2 Reconstruction and Sparsity} across layers, (a) without and (b) with SAE constraints. The gap between image-token (orange) and text-token (green) curves measures alignment where closer curves mean visual tokens inhabit the same sparse manifold as text. Without constraints, curves converge only near the end of the network; with constraints, convergence is immediate.}
\end{figure}

We analyze these metrics across different LLM backbones. Our results reveal a disparity in how different architectures achieve cross-modal alignment. For the Gemma models (2B and 9B) trained without SAE constraints, we replicate the findings of \citet{venhoff2025how}: visual tokens exhibit high reconstruction error and density in early layers, with alignment occurring only in the final layers (\cref{fig:reconstr_sparsity_dinov2_nosae,fig:reconstr_sparsity_dinov2_gemma9b_nosae}). This ``delayed alignment'' suggests that standard training struggles to map visual data to the LLM's sparse feature manifold. However, when training with SAE constraints, we observe a shift. As shown in the reconstruction graphs (\cref{fig:reconstr_sparsity_dinov2_wsae,fig:reconstr_sparsity_dinov2_gemma9b_wsae}), visual tokens achieve immediate convergence, matching the low reconstruction error and low density of text tokens from the very first layer.

When analyzing the LLaMA-3.1-8B-Instruct architecture, we observe a notable divergence in behavior. Unlike Gemma, where unconstrained visual tokens eventually align in deeper layers, the unconstrained LLaMA model fails to converge (\cref{fig:reconstr_sparsity_dinov2_llama_nosae}); visual tokens remain dense and exhibit high reconstruction error throughout the entire network depth. This indicates that LLaMA's internal representation space is harder for the visual projector to target without guidance, making SAE constraints even more important for the visual features to be interpretable via textual SAEs. Applying our SAE constraints successfully forces alignment, reducing reconstruction error and inducing sparsity (\cref{fig:reconstr_sparsity_dinov2_llama_wsae}).

\begin{figure}[!t]
    \centering
    
    \includegraphics[width=0.48\textwidth]{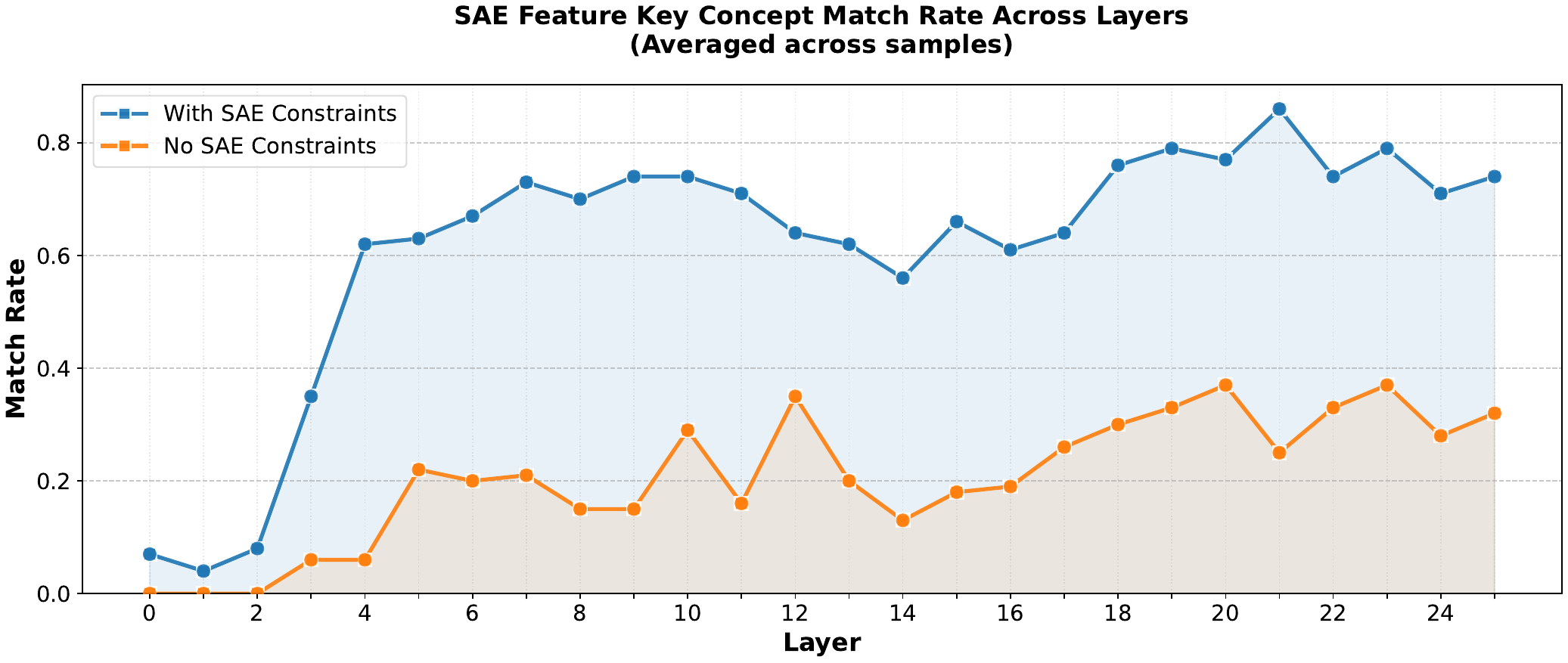}
    \caption{\textbf{DINOv2 Match Rate} for visual tokens with and without SAE constraints with Gemma-2-2B-it LLM model. }

    \label{fig:matchrate_dinov2}
\end{figure}

\subsection{Interpretability Matching Rate}
\label{sec:match_rate}

We quantify interpretability via an LLM-Verified Matching Rate \cite{venhoff2025how}. For a given image, we first project its visual tokens through the text SAE. At each layer, we identify the three latents with the highest activation values. To verify their relevance, we employ an oracle LLM, presenting it with both the original image and the textual descriptions associated with these top latents from Neuronpedia. If the LLM confirms that at least one latent description accurately represents a concept visible in the image, we record a successful ``match'' for that layer. 

For Gemma, the baseline (no SAE) matching rate gradually improves in later layers as natural alignment occurs (\cref{fig:matchrate_dinov2} and \cref{fig:matchrate_dinov2_gemma9b}). This also holds for different visual encoders such as CLIP and I-JEPA (\cref{fig:matchrate_clip_gemma2b,fig:matchrate_jepa_gemma2b}). In contrast, for LLaMA, the baseline matching rate remains stagnant and low across all layers, reinforcing that natural alignment is not guaranteed across all LLM architectures (\cref{fig:matchrate_dinov2_llama}).
Adding SAE constraints not only improves the matching rate (e.g., by a threefold increase on average for DINOv2 with Gemma-2-2B-it as shown in \cref{fig:matchrate_dinov2}) but also bridges the gap between different LLMs, achieving high matching rates in LLaMA equivalent to those seen in Gemma. Full results across all visual encoders and LLM backbones are provided in \cref{appendix:matchrate_llms}.

To verify that matching rate improvements generalize beyond photo-centric training data, we evaluate VISTA 500 OOD images from DomainNet \cite{peng2019moment}. The matching rate is comparable to the in-distribution baseline despite the substantial distribution shift, indicating that VISTA's alignment is a structural property rather than a training-set artifact. Full results are provided in \cref{appendix:ood_robustness}.

\begin{figure}[!t]
    \centering
    \scriptsize
    \begin{subfigure}{\linewidth}
        \centering
        \begin{minipage}[t]{0.32\linewidth}
            \vspace{0pt}
            \includegraphics[width=\linewidth]{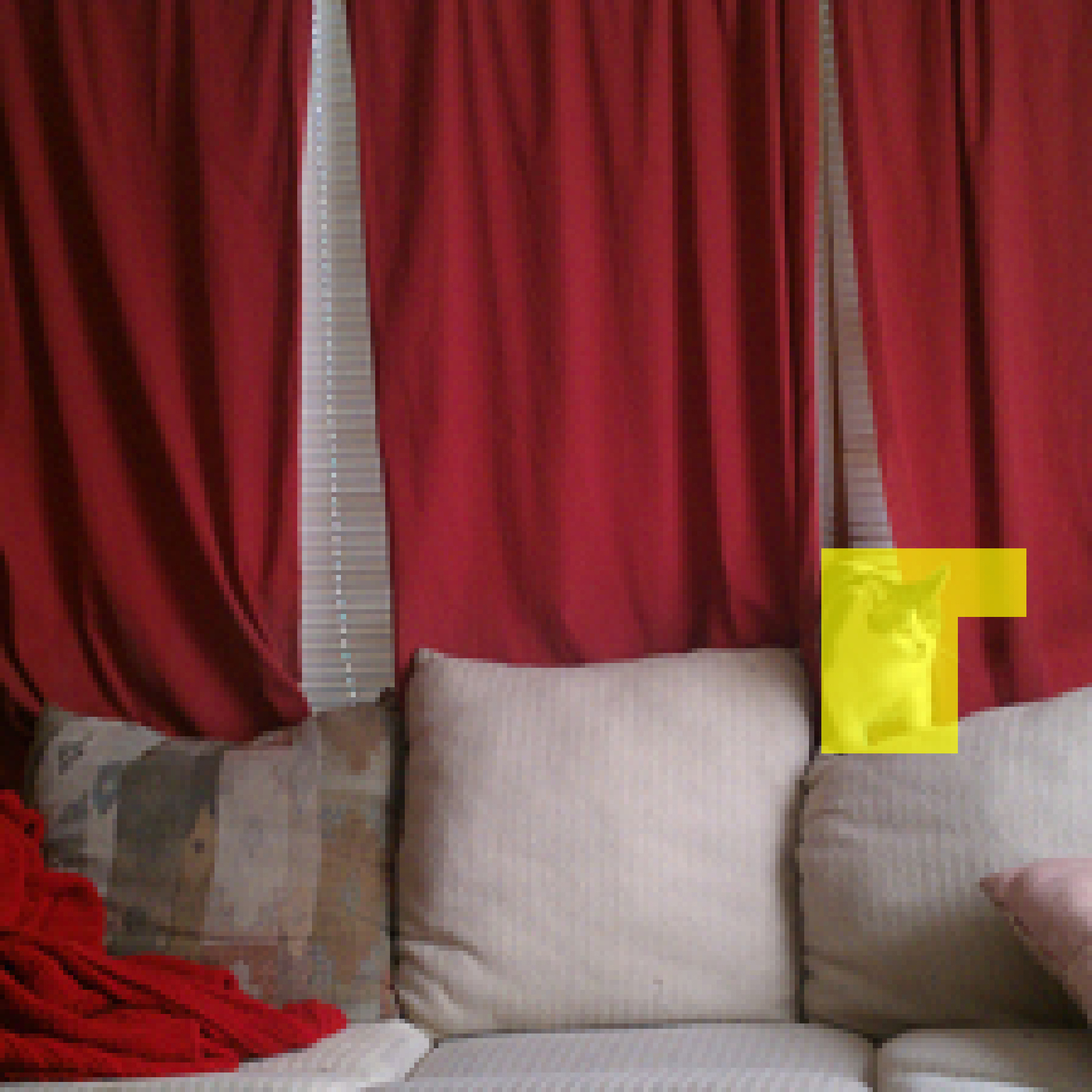}
        \end{minipage}
        \hfill
        \begin{minipage}[t]{0.65\linewidth}
            \vspace{0pt}
            \textbf{Q: What is shown in this image?}  \\
            \textbf{Original:} The image shows a cat sitting on a couch with a pillow. The cat is looking at the camera.  \\
            \textbf{Remove ``cat'':} The image shows a couch with a person sitting on it. The person is wearing a white shirt and has a red scarf around their neck.  \\
            \textbf{Replace with ``dog'':} The image shows a dog sitting on a couch with a pillow. The dog is a white and tan mix. 
        \end{minipage}
    \end{subfigure}

    \begin{subfigure}{\linewidth}
        \centering
        \begin{minipage}[t]{0.32\linewidth}
            \vspace{0pt}
            \includegraphics[width=\linewidth]{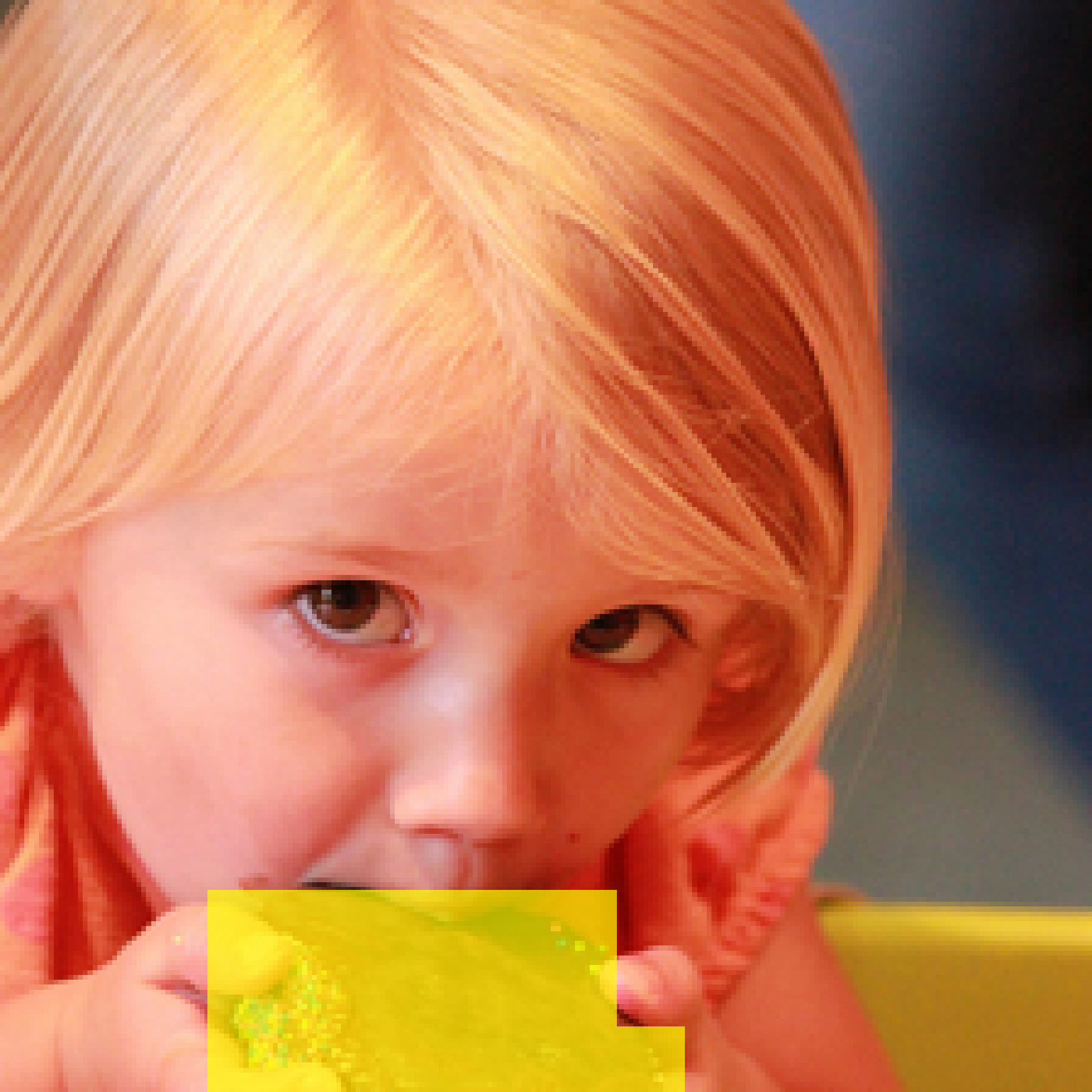}
        \end{minipage}
        \hfill
        \begin{minipage}[t]{0.65\linewidth}
            \vspace{0pt}
            \textbf{Q: What is shown in this image?} \\
            \textbf{Original:} The image shows a girl eating a chocolate chip cookie. The girl is holding a cookie in her hand and is eating it.  \\
            \textbf{Remove ``chocolate'':} This image shows a girl with a large amount of food on her face. The girl is eating a sandwich.  \\
            \textbf{Replace with ``bread'':} This image shows a girl with shirt. She is holding a piece of bread in her hand. 
        \end{minipage}
    \end{subfigure}

    \caption{Visual steering with DINOv2, selecting precise patches and changing model's understanding of what is contained in them.}
    \label{fig:steering_example_dinov2}
\end{figure}

\section{Insights on Visual Steering}
\label{sec:applic_steering}

The interpretability transfer established in the previous section raises the question of whether visual tokens truly inhabit the text SAE manifold or whether the alignment is merely an artifact of the matching rate metric. We address this through a causal test. If the alignment is genuine then arithmetic operations using text-SAE concept directions that are meaningful in the text domain, should steer the model's output towards the concept of intervention when applied to visual tokens. The experiments in this section serve as this causal validation. We use object removal and replacement as a controlled testbed because they admit clear success criteria (the target concept either disappears or changes).  \cref{sec:abstract_manip} (abstract concept manipulation) and \cref{sec:clustering} (image clustering by activated concepts) provide complementary downstream validations of the same alignment property.

As our first validation test, we examine whether localized arithmetic interventions on visual tokens produce spatially coherent changes in model output. Unlike standard steering, which alters the targeted concept globally across the entire image, fine-grained steering allows us to modify specific token positions while leaving the rest of the scene untouched.

\subsection{Qualitative Results: Spatial Modification}
As an example, we use an image of ``a girl eating a cookie'', first identifying the specific visual tokens (patches) where the cookie is located. We then apply the steering vectors only to those specific patches. As shown in \cref{fig:steering_example_dinov2} (with more examples in \cref{fig:steering_example_dinov2_appendix}), the results are highly localized; the original description of a ``girl eating a chocolate cookie'' shifts to a ``sandwich'' after removing the ``chocolate'' direction, or ``bread'' after replacing it with the ``bread'' direction in the relevant patches. This precise control is uniquely enabled by DINOv2's preservation of spatial alignment. I-JEPA also supports steering, but with lower consistency than DINOv2 (see \cref{tab:steering_results_full} for quantitative evaluation). 

In contrast, steering with CLIP-based models is unstable. As shown in \cref{fig:steering_example_clip}, since CLIP does not reliably preserve spatial topology, selecting specific patches for steering often yields no effect. While steering can be achieved by applying vectors to \textbf{all} tokens, this causes the model to hallucinate. For instance, when attempting to change a ``cat on a sofa'' to a ``dog'' using global CLIP steering, the model describes unrelated concepts such as ``a person in a blue jacket standing by a fence'', elements not present in the image.

\subsection{Quantitative Evaluation of Steering}

To assess the precision of fine-grained steering, we implement a quantitative evaluation framework. We use an LLM judge to determine whether modifications are localized correctly and whether the global scene context remains intact.

\subsubsection{Dataset and Patch Selection}
For this evaluation, we sample 100 images from the COCO validation set, spanning diverse categories including animals, food, and vehicles. To ensure the steering task requires high spatial precision, we filter the dataset for images where the target object occupies between 2\% and 30\% of the total image area. We use COCO ground-truth bounding boxes to select the specific visual patch tokens corresponding to the object targeted for steering.

\subsubsection{Scoring Protocol}

For each steered image, the judge LLM evaluates the generated text against the original image and the baseline description (output prior to steering). We define \textit{Source Concept} as the class initially present in the baseline and \textit{Target Keyword} as the concept we intend to replace or remove. For removals, Source Concept corresponds to the Target Keyword. For replacements, the Target Keyword is sampled from available classes. Success is determined by the operation type:

\begin{enumerate}
    \item \textbf{Steering Success Check:} The LLM assigns a score of 0 if the steering operation fails:
    \begin{itemize}
        \item \textbf{Replace:} The Target Keyword is missing, \textbf{OR} the Source Concept is still present in the text (indicating a conceptual collision where the object was not fully replaced).
        \item \textbf{Remove:} The Target Keyword is still present in the description.
    \end{itemize}
    
    \item \textbf{Baseline Comparison:} If the steering succeeds, the description is compared to the original baseline. If the model successfully modifies the object without adding new environmental details (e.g., background or scene layout) beyond what is present in the baseline description, it receives a Score of 2. 

    \item \textbf{Image Grounding:} If the steered text introduces new descriptive details not found in the baseline text, the LLM verifies them against the source image:
    \begin{itemize}
        \item If the new details are physically present in the image, it receives a score of 2. 
        \item If the new details are hallucinated (not in the image), it gets a score of 1. 
    \end{itemize}
\end{enumerate}

\subsubsection{Steering Metric}

We quantify steering performance using a weighted mean score, defined as:
\begin{equation}
    \text{Mean Score} = \frac{2 \cdot N_{S2} + 1 \cdot N_{S1}}{N_{\text{total}}}
\end{equation}
where $N_{S2}$ and $N_{S1}$ denote the number of samples classified as Strong Success (score 2) and Partial Success (score 1), respectively, and $N_{\text{total}}$ is the total number of evaluation samples. Thus, the score is bounded between 0 and 2. 

\subsubsection{Results and Analysis}

\begin{table}[!t]
\centering
\caption{Results of steering with Gemma-2-2B-it. We report the sample counts for \textbf{S2} (Strong Success), \textbf{S1} (Partial Success), and \textbf{S0} (Failure). VISTA with DINOv2 achieves best results. }
\label{tab:steering_results_full}
\resizebox{0.49\textwidth}{!}{
\begin{tabular}{@{}llcccc c cccc@{}}
\toprule
& & \multicolumn{4}{c}{\textbf{Remove Operation}} & & \multicolumn{4}{c}{\textbf{Replace Operation}} \\ 
\cmidrule(lr){3-6} \cmidrule(lr){8-11}
\textbf{Method} & \textbf{Visual Encoder} & \textbf{S2} & \textbf{S1} & \textbf{S0} & \textbf{MS} & & \textbf{S2} & \textbf{S1} & \textbf{S0} & \textbf{MS} \\ 
\midrule
\textbf{VisionSAE} & CLIP Global & 11 & 84 & 5 & 1.06 & & 8 & 49 & 43 & 0.65 \\
& CLIP Local & 24 & 23 & 53 & 0.71 & & 23 & 14 & 63 & 0.60 \\
& CLIP Global (Rem+Rep) & --- & --- & --- & --- & & 0 & 1 & 99 & 0.01 \\
& CLIP Local (Rem+Rep) & --- & --- & --- & --- & & 22 & 2 & 76 & 0.46 \\
\midrule
\textbf{PatchSAE} & CLIP Local & 5 & 55 & 40 & 0.65 & & 34 & 31 & 35 & 0.99 \\
\midrule
\multirow{3}{*}{\shortstack[l]{\textbf{Ablation} \\ \textbf{(No SAE)}}} & DINOv2  & 10 & 2 & 88 & 0.22 & & 3 & 0 & 97 & 0.06 \\
& CLIP  & 13 & 20 & 67 & 0.46 & & 16 & 6 & 78 & 0.38 \\
& I-JEPA  & 8 & 18 & 74 & 0.34 & & 20 & 9 & 71 & 0.49 \\
\midrule
\multirow{3}{*}{\textbf{VISTA}} & DINOv2  & 47 & 49 & 4 & \textbf{1.43} & & 59 & 28 & 13 & \textbf{1.46} \\
 & CLIP - Gemma 2B   & 18 & 4 & 78 & 0.40 & & 7 & 4 & 89 & 0.18 \\
 & I-JEPA - Gemma 2B   & 8 & 25 & 67 & 0.41 & & 17 & 47 & 36 & 0.81 \\
\bottomrule
\end{tabular}
}
\end{table}

\paragraph{Baselines}

\cref{tab:steering_results_full} presents the results of our evaluation. We evaluate our approach against two state-of-the-art visual steering baselines: VisionSAE \cite{pach2025sparse}, which utilizes SAEs trained on global CLIP tokens, and PatchSAE \cite{lim2025patchsae}, which trains SAEs on individual CLIP patches. We specifically focus on SAE-based baselines as they have been shown to outperform traditional activation steering methods. \citet{pach2025sparse} demonstrated that SAE-based directions surpass the Difference-in-Means \cite{diffinmean} approach in both concept insertion and removal. 

While effective for broad concept removal, VisionSAE lacks spatial precision. As detailed in \cref{tab:steering_results_full}, it frequently alters the background or global context, resulting in a higher rate of score 1 (S1), where hallucinated details are added, compared to score 2 (S2), which requires steering with no alterations beyond the targeted object. This limitation is particularly acute in replacement tasks: because the method relies on global semantic vectors, the model struggles with fine-grained swaps, often superimposing the new concept or modifying unrelated image regions. To address this, in addition to the default (Global) implementation, we evaluate two modifications to this baseline:

\begin{enumerate}
    \item \textbf{Local Patch Steering:} We restrict steering to specific local patches. This yields a lower Mean Score. 
    \item \textbf{Sequential Edit (Remove + Replace):} We attempt to first remove the target concept and subsequently inject the new one. This approach also underperforms the global steering according to the Mean Score. 
\end{enumerate}

We also evaluate PatchSAE\footnote{We adapted the original PatchSAE to enable steering abilities.}, which theoretically offers better granularity by training on CLIP's patch tokens directly. However, as shown in \cref{tab:steering_results_full}, even with patch-level training it underperforms our LLM-based steering which provides better control than steering the visual encoder and is particularly superior in targeted removal, where PatchSAE often hallucinates extra details (as evidenced by its significantly higher S1 rate relative to S2).

Our method achieves substantially higher overall steering quality on both replacement and removal while preserving the scene. Specifically, VISTA with DINOv2 improves the Mean Score from 1.06 to 1.43 on removal (a 35\% gain over the strongest vision-only baseline, VisionSAE) and from 0.99 to 1.46 on replacement (a 47\% gain over PatchSAE). Furthermore, visual token steering fails without SAE constraints as evidenced by the low Mean Scores in our ablation. The ``No SAE'' baseline applies the identical steering procedure to VISTA, but on a projector trained without SAE reconstruction loss. Moreover, steering is less reliable when CLIP and I-JEPA encoders are used due to their lower spatial fidelity. 

For VISTA, the replace operation generally achieves a higher S2 rate than remove (\cref{tab:llm_ablation_vista_allencoders}). This asymmetry is intrinsic to the steering mechanism as localized subtraction does not inpaint the affected patches but leaves modified embeddings that the LLM must decode coherently. As shown in \cref{fig:steering_example_dinov2}, removing a cat from ``a cat sitting on a couch'' creates a semantic void at those patch positions, which the LLM fills with a plausible substitute (``a couch with a person sitting on it'') to maintain scene coherence. Replacement avoids this issue as injecting a target concept gives the LLM coherent content to describe at the steered patches.

We also demonstrate the robustness of our steering mechanism by evaluating it across different LLM backbones; these ablation results are detailed in \cref{appendix:steering_backbones_abl}. While CLIP's spatial localization accuracy increases with larger LLM backbones as shown in \cref{tab:localization_accuracy_sae_only}, this improvement is not sufficient to enable reliable fine-grained steering. As shown in \cref{tab:llm_ablation_vista_allencoders}, CLIP-based models continue to underperform DINOv2 in both object removal and replacement tasks.

We provide details on steering parameters and layers used in \cref{appendix:implm_details}. Finally, we validate that VISTA's effect stems from intervention on visual tokens (rather than linguistic hijacking) through a controlled comparison against text-only steering, detailed in \cref{appendix:visual_vs_text_steering}.

\begin{figure}[!htbp]
    \centering
    
    \begin{subfigure}[t]{0.15\textwidth}
        \centering
        \includegraphics[width=\textwidth]{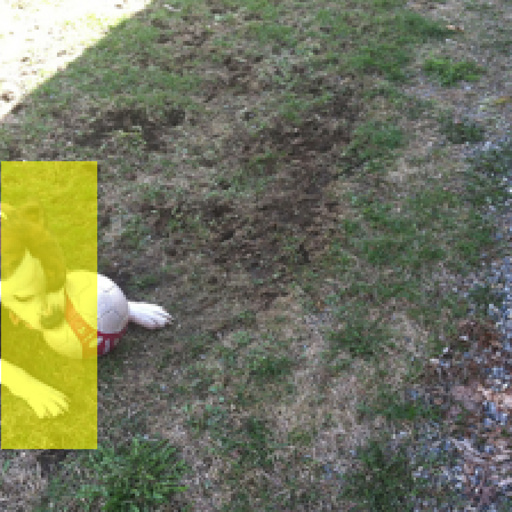}
        \caption{\scriptsize \textbf{Steering: Sadness.} \\ \textit{Q: Is the dog happy?} \\ \textbf{Baseline:} "The dog is happy because it is playing with a soccer ball." \\ \textbf{Steered:} "No, the dog is not happy."}
        \label{fig:steering_sad}
    \end{subfigure}
    \hfill
    \begin{subfigure}[t]{0.15\textwidth}
        \centering
        \includegraphics[width=\textwidth]{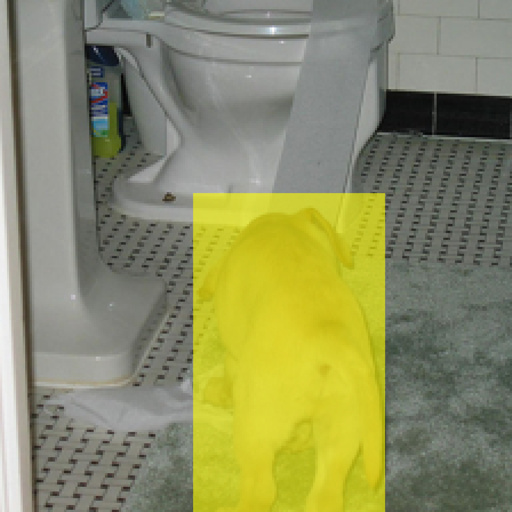}
        \caption{\scriptsize \textbf{Steering: Sleeping.} \\ \textit{Q: Is the dog awake or sleeping?} \\ \textbf{Baseline:} "The dog is awake and standing on the bathroom floor." \\ \textbf{Steered:} "The dog is sleeping on the bathroom floor."}
        \label{fig:steering_sleep}
    \end{subfigure}
    \hfill
    \begin{subfigure}[t]{0.15\textwidth}
        \centering
        \includegraphics[width=\textwidth]{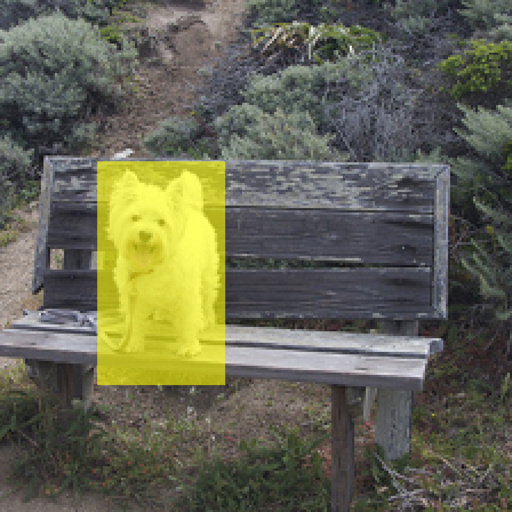}
        \caption{\scriptsize \textbf{Steering: Danger.} \\ \textit{Q: Is the dog safe to approach?} \\ \textbf{Baseline:} "Yes, the dog is safe to approach." \\ \textbf{Steered:} "No, the dog shown is not safe to approach."}
        \label{fig:steering_danger}
    \end{subfigure}
    \caption{\textbf{Qualitative results.} We steer (a) \textit{Sadness}, (b) \textit{Sleeping}, and (c) \textit{Danger}. In all cases, the steering vector inverts the model's interpretation (Steered) vs the original (Baseline).}
    \label{fig:steering_qualitative}
    
\end{figure}

\subsection{Manipulating High-Level Concepts}
\label{sec:abstract_manip}

We also target high-level concepts. Though abstract notions like emotion or threat lack bounding boxes, they remain anchored to identifiable subjects. We select patches covering the subject exhibiting the concept (highlighted in yellow in \cref{fig:steering_qualitative}), and manipulate latents for emotions, physiological state, and threat level. Such concepts are binary in language but lack a fixed visual signature. For instance, ``danger'' can manifest through aggression, instability, or environmental hazards. This demonstrates another advantage of steering with text-aligned SAEs over visual ones.

\subsection{Clustering Images by Activated SAE Concepts}
\label{sec:clustering}

\begin{figure}[!htbp]
    \centering
    \includegraphics[width=\linewidth]{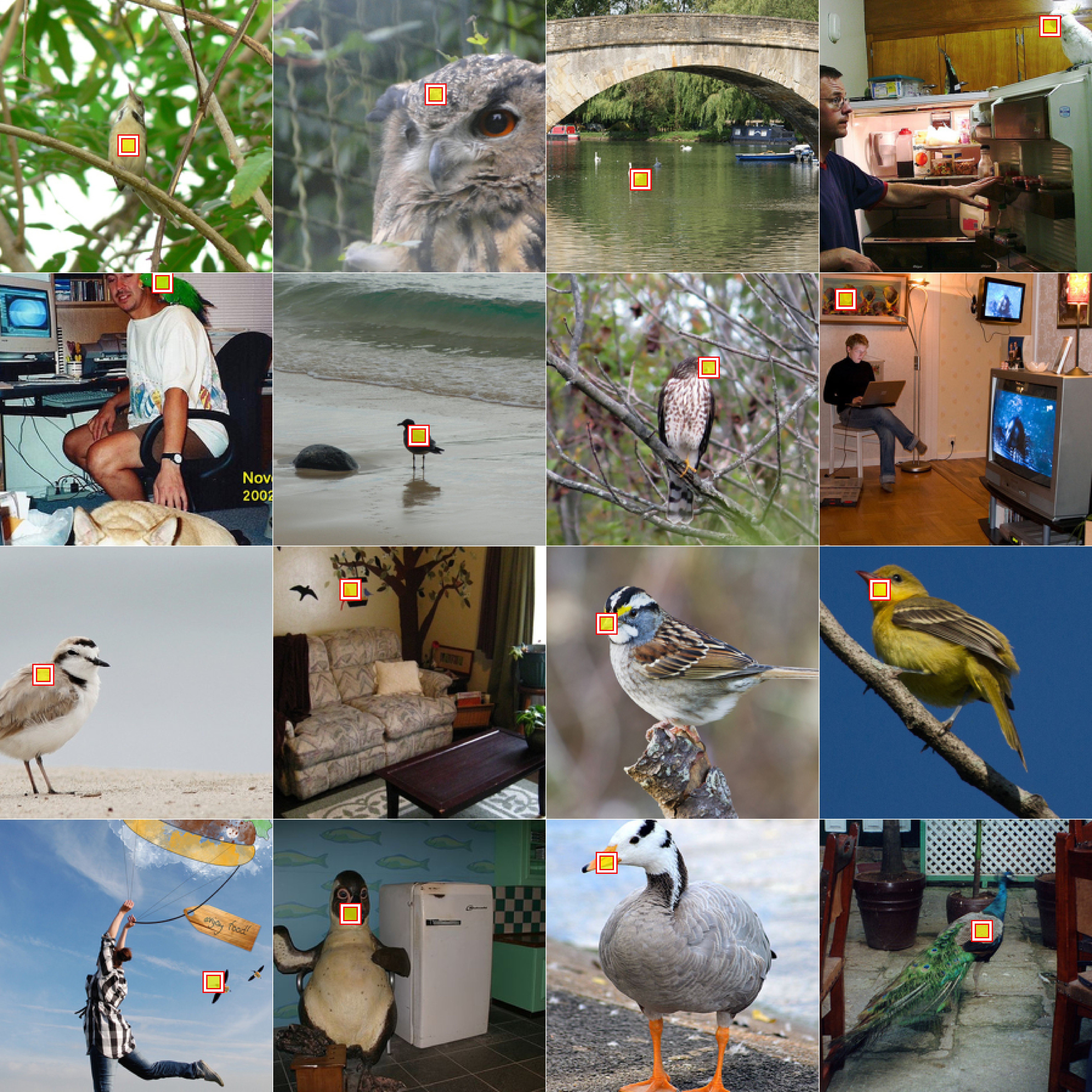}
    \caption{\textbf{Image Clustering.} Sample of images that activate the latent ``terms and concepts related to birds''. The patch that activates the latent is shown in yellow with red bounding box. }
    \label{fig:cluster_birds}
\end{figure}

We further demonstrate that the aligned SAE feature space can be utilized to cluster images based on shared semantic attributes. To perform clustering, we first select the top activating tokens from each image, and then identify the images that activate the same concept. As shown in \cref{fig:cluster_birds}, this yields coherent semantic clusters, confirming that visual tokens inherit the language model's structured organization.

\section{Limitations and Future Work}
\label{sec:limitations}

\paragraph{Dependence on Pre-trained Text SAEs} VISTA requires the availability of SAE trained on the target LLM. We believe that this is becoming less of a constraint as foundational models are increasingly released with paired SAEs.

\paragraph{SAE Concept Coverage} Text SAEs can only interpret visual concepts with corresponding features in the LLM's representation. Rare concepts may activate broader, less precise latents. VISTA inherits future improvements ``for free'' by swapping in larger SAEs.

\paragraph{Future Work} VISTA establishes the cross-modal alignment necessary for a broader research agenda we leave to future work: (i) feature circuit analysis \cite{marks2025sparse} to trace the causal effect of visual inputs through the LLM's internal computation; (ii) hallucination detection \cite{abdaljalil-etal-2025-safe} by identifying when visual tokens trigger ungrounded textual latents. Each requires substantial future work, but VISTA provides the foundational alignment they require. Also, VISTA's interpretability transfer succeeds across all encoders; only spatial precision degrades with weaker encoders. Incorporating self-supervised localized losses into contrastive encoders could mitigate this.

\section{Conclusion}
In this work, we demonstrated that interpretability is a transferable property across modalities. By enforcing SAE constraints during the alignment of vision encoders to LLMs, we mapped visual tokens into a pre-existing, labeled linguistic feature space, eliminating the computational and labeling costs of training vision-specific SAEs. Our evaluation revealed that the success of localized interpretability and steering is coupled with the spatial fidelity of the vision backbone; DINOv2’s local objective provides the necessary precision for this, significantly outperforming contrastive models like CLIP. This ``interpretability transfer'' paradigm provides a scalable path for understanding and controlling complex multimodal systems.

\clearpage

\section*{Acknowledgements}
We’d like to gratefully acknowledge Lambda Labs compute support and the support from University of Bath for the studentship.

\section*{Impact Statement}

This paper presents work whose goal is to advance the field of Machine
Learning. We specifically intend to enhance the interpretability aspect. There are many potential societal consequences of our work, none
which we feel must be specifically highlighted here.

\bibliography{paper_noUrl}

@article{bricken2023monosemanticity,
       title={Towards Monosemanticity: Decomposing Language Models With Dictionary Learning},
       author={Bricken, Trenton and Templeton, Adly and Batson, Joshua and Chen, Brian and Jermyn, Adam and Conerly, Tom and Turner, Nick and Anil, Cem and Denison, Carson and Askell, Amanda and Lasenby, Robert and Wu, Yifan and Kravec, Shauna and Schiefer, Nicholas and Maxwell, Tim and Joseph, Nicholas and Hatfield-Dodds, Zac and Tamkin, Alex and Nguyen, Karina and McLean, Brayden and Burke, Josiah E and Hume, Tristan and Carter, Shan and Henighan, Tom and Olah, Christopher},
       year={2023},
       journal={Transformer Circuits Thread}
    }

@article{elhage2022superposition,
   title={Toy Models of Superposition},
   author={Elhage, Nelson and Hume, Tristan and Olsson, Catherine and Schiefer, Nicholas and Henighan, Tom and Kravec, Shauna and Hatfield-Dodds, Zac and Lasenby, Robert and Drain, Dawn and Chen, Carol and Grosse, Roger and McCandlish, Sam and Kaplan, Jared and Amodei, Dario and Wattenberg, Martin and Olah, Christopher},
   year={2022},
   journal={Transformer Circuits Thread}
}

@misc{rajamanoharan_2024_jumping,
  author = {Rajamanoharan, Senthooran and Lieberum, Tom and Sonnerat, Nicolas and Conmy, Arthur and Varma, Vikrant and Kramár, János and Nanda, Neel},
  title = {Jumping Ahead: Improving Reconstruction Fidelity with JumpReLU Sparse Autoencoders},
  year = {2024},
  organization = {arXiv.org}
}

@inproceedings{liu_llava_2023,
 author = {Liu, Haotian and Li, Chunyuan and Wu, Qingyang and Lee, Yong Jae},
 booktitle = {Advances in Neural Information Processing Systems},
 editor = {A. Oh and T. Naumann and A. Globerson and K. Saenko and M. Hardt and S. Levine},
 pages = {34892--34916},
 publisher = {Curran Associates, Inc.},
 title = {Visual Instruction Tuning},
 volume = {36},
 year = {2023}
}

@article{venhoff2025how,
  title     = {How Visual Representations Map to Language Feature Space in Multimodal LLMs},
  author    = {Constantin Venhoff and Ashkan Khakzar and Sonia Joseph and Philip Torr and Neel Nanda},
  booktitle = {The 4th Explainable AI for Computer Vision (XAI4CV) Workshop at CVPR 2025},
  year      = {2025}
}

@inproceedings{
fu2025mme,
title={{MME}: A Comprehensive Evaluation Benchmark for Multimodal Large Language Models},
author={Chaoyou Fu and Peixian Chen and Yunhang Shen and Yulei Qin and Mengdan Zhang and Xu Lin and Jinrui Yang and Xiawu Zheng and Ke Li and Xing Sun and Yunsheng Wu and Rongrong Ji and Caifeng Shan and Ran He},
booktitle={The Thirty-ninth Annual Conference on Neural Information Processing Systems Datasets and Benchmarks Track},
year={2025}
}

@inproceedings{hudson2019gqa,
  title={Gqa: A new dataset for real-world visual reasoning and compositional question answering},
  author={Hudson, Drew A and Manning, Christopher D},
  booktitle={Proceedings of the IEEE/CVF conference on computer vision and pattern recognition},
  pages={6700--6709},
  year={2019}
}

@inproceedings{pope_bench,
    title = "Evaluating Object Hallucination in Large Vision-Language Models",
    author = "Li, Yifan  and
      Du, Yifan  and
      Zhou, Kun  and
      Wang, Jinpeng  and
      Zhao, Xin  and
      Wen, Ji-Rong",
    editor = "Bouamor, Houda  and
      Pino, Juan  and
      Bali, Kalika",
    booktitle = "Proceedings of the 2023 Conference on Empirical Methods in Natural Language Processing",
    month = dec,
    year = "2023",
    address = "Singapore",
    publisher = "Association for Computational Linguistics",
    doi = "10.18653/v1/2023.emnlp-main.20",
    pages = "292--305"
}

@inproceedings{sharma2018conceptual,
  title = {Conceptual Captions: A Cleaned, Hypernymed, Image Alt-text Dataset For Automatic Image Captioning},
  author = {Sharma, Piyush and Ding, Nan and Goodman, Sebastian and Soricut, Radu},
  booktitle = {Proceedings of ACL},
  year = {2018}
}

@inproceedings{
pach2025sparse,
title={Sparse Autoencoders Learn Monosemantic Features in Vision-Language Models},
author={Mateusz Pach and Shyamgopal Karthik and Quentin Bouniot and Serge Belongie and Zeynep Akata},
booktitle={The Thirty-ninth Annual Conference on Neural Information Processing Systems},
year={2025}}

@article{sjose_steeringclip,
  title     = {Steering CLIP’s vision transformer with sparse autoencoders},
  author    = {Sonia Joseph and Praneet Suresh and Ethan Goldfarb and Lorenz Hufe and Yossi Gandelsman and Robert Graham and Danilo Bzdok and Wojciech Samek and Blake Aaron Richards
},
  booktitle = {CVPR 2025 Workshop on Mechanistic Interpretability for Vision (MIV)},
  year      = {2025}
}

@inproceedings{radford2021LearningTV,
  title={Learning Transferable Visual Models From Natural Language Supervision},
  author={Alec Radford and Jong Wook Kim and Chris Hallacy and Aditya Ramesh and Gabriel Goh and Sandhini Agarwal and Girish Sastry and Amanda Askell and Pamela Mishkin and Jack Clark and Gretchen Krueger and Ilya Sutskever},
  booktitle={International Conference on Machine Learning},
  year={2021}
}

@article{
oquab2024dinov,
title={{DINO}v2: Learning Robust Visual Features without Supervision},
author={Maxime Oquab and Timoth{\'e}e Darcet and Th{\'e}o Moutakanni and Huy V. Vo and Marc Szafraniec and Vasil Khalidov and Pierre Fernandez and Daniel HAZIZA and Francisco Massa and Alaaeldin El-Nouby and Mido Assran and Nicolas Ballas and Wojciech Galuba and Russell Howes and Po-Yao Huang and Shang-Wen Li and Ishan Misra and Michael Rabbat and Vasu Sharma and Gabriel Synnaeve and Hu Xu and Herve Jegou and Julien Mairal and Patrick Labatut and Armand Joulin and Piotr Bojanowski},
journal={Transactions on Machine Learning Research},
issn={2835-8856},
year={2024},
note={Featured Certification}
}

@article{Assran2023SelfSupervisedLF,
  title={Self-Supervised Learning from Images with a Joint-Embedding Predictive Architecture},
  author={Mahmoud Assran and Quentin Duval and Ishan Misra and Piotr Bojanowski and Pascal Vincent and Michael G. Rabbat and Yann LeCun and Nicolas Ballas},
  journal={2023 IEEE/CVF Conference on Computer Vision and Pattern Recognition (CVPR)},
  year={2023},
  pages={15619-15629}
}

@inproceedings{
  lim2025patchsae,
  title={Sparse autoencoders reveal selective remapping of visual concepts during adaptation},
  author={Hyesu Lim and Jinho Choi and Jaegul Choo and Steffen Schneider},
  booktitle={The Thirteenth International Conference on Learning Representations},
  year={2025}
}

@inproceedings{
huben2024sparse,
title={Sparse Autoencoders Find Highly Interpretable Features in Language Models},
author={Robert Huben and Hoagy Cunningham and Logan Riggs Smith and Aidan Ewart and Lee Sharkey},
booktitle={The Twelfth International Conference on Learning Representations},
year={2024}
}

@misc{daujotas_2023_case,
  author = {Daujotas, Gytis},
  title = {Case Study: Interpreting, Manipulating, and Controlling CLIP With Sparse Autoencoders},
  urldate = {2026-01-12},
  year = {2024},
  organization = {Lesswrong.com}
}

@inproceedings{
limber,
title={Linearly mapping from image to text space},
author={Jack Merullo and Louis Castricato and Carsten Eickhoff and Ellie Pavlick},
booktitle={In The Eleventh International Conference on Learning Representations},
year={2023}
}

@inproceedings{diffinmean,
  author={Andy Arditi and Oscar Obeso and Aaquib Syed and Daniel Paleka and Nina Panickssery and Wes Gurnee and Neel Nanda},
  title={Refusal in Language Models Is Mediated by a Single Direction},
  year={2024},
  cdate={1704067200000},
  booktitle={NeurIPS}
}

@inproceedings{rimsky-etal-2024-steering,
    title = "Steering Llama 2 via Contrastive Activation Addition",
    author = "Rimsky, Nina  and
      Gabrieli, Nick  and
      Schulz, Julian  and
      Tong, Meg  and
      Hubinger, Evan  and
      Turner, Alexander",
    editor = "Ku, Lun-Wei  and
      Martins, Andre  and
      Srikumar, Vivek",
    booktitle = "Proceedings of the 62nd Annual Meeting of the Association for Computational Linguistics (Volume 1: Long Papers)",
    month = aug,
    year = "2024",
    address = "Bangkok, Thailand",
    publisher = "Association for Computational Linguistics",
    doi = "10.18653/v1/2024.acl-long.828",
    pages = "15504--15522",
    }

@misc{turner_2023_activation,
  author = {Turner, Alexander Matt and Thiergart, Lisa and Udell, David and Leech, Gavin and Mini, Ulisse and MacDiarmid, Monte},
  month = {08},
  title = {Activation Addition: Steering Language Models Without Optimization},
  year = {2023},
  organization = {arXiv.org}
}

@misc{lin2015microsoft,
      title={Microsoft COCO: Common Objects in Context},
      author={Tsung-Yi Lin and Michael Maire and Serge Belongie and Lubomir Bourdev and Ross Girshick and James Hays and Pietro Perona and Deva Ramanan and C. Lawrence Zitnick and Piotr Dollár},
      year={2015},
      eprint={1405.0312},
      archivePrefix={arXiv},
      primaryClass={cs.CV}
}

@misc{lieberum2024gemmascopeopensparse,
      title={Gemma Scope: Open Sparse Autoencoders Everywhere All At Once on Gemma 2}, 
      author={Tom Lieberum and Senthooran Rajamanoharan and Arthur Conmy and Lewis Smith and Nicolas Sonnerat and Vikrant Varma and János Kramár and Anca Dragan and Rohin Shah and Neel Nanda},
      year={2024},
      eprint={2408.05147},
      archivePrefix={arXiv},
      primaryClass={cs.LG} 
}

@inproceedings{
wang2025towards,
title={Towards Universality: Studying Mechanistic Similarity Across Language Model Architectures},
author={Junxuan Wang and Xuyang Ge and Wentao Shu and Qiong Tang and Yunhua Zhou and Zhengfu He and Xipeng Qiu},
booktitle={The Thirteenth International Conference on Learning Representations},
year={2025}
}

@inproceedings{
marks2025sparse,
title={Sparse Feature Circuits: Discovering and Editing Interpretable Causal Graphs in Language Models},
author={Samuel Marks and Can Rager and Eric J Michaud and Yonatan Belinkov and David Bau and Aaron Mueller},
booktitle={The Thirteenth International Conference on Learning Representations},
year={2025}
}

@inproceedings{abdaljalil-etal-2025-safe,
    title = "{SAFE}: A Sparse Autoencoder-Based Framework for Robust Query Enrichment and Hallucination Mitigation in {LLM}s",
    author = "Abdaljalil, Samir  and
      Pallucchini, Filippo  and
      Seveso, Andrea  and
      Kurban, Hasan  and
      Mercorio, Fabio  and
      Serpedin, Erchin",
    editor = "Christodoulopoulos, Christos  and
      Chakraborty, Tanmoy  and
      Rose, Carolyn  and
      Peng, Violet",
    booktitle = "Findings of the Association for Computational Linguistics: EMNLP 2025",
    month = nov,
    year = "2025",
    address = "Suzhou, China",
    publisher = "Association for Computational Linguistics",
    doi = "10.18653/v1/2025.findings-emnlp.496",
    pages = "9335--9346",
    ISBN = "979-8-89176-335-7",
}

@article{he2024llamascope,
  title={Llama Scope: Extracting Millions of Features from Llama-3.1-8B with Sparse Autoencoders},
  author={He, Zhengfu and Shu, Wentao and Ge, Xuyang and Chen, Lingjie and Wang, Junxuan and Zhou, Yunhua and Liu, Frances and Guo, Qipeng and Huang, Xuanjing and Wu, Zuxuan and others},
  journal={arXiv preprint arXiv:2410.20526},
  year={2024}
}

@inproceedings{peng2019moment,
  title={Moment matching for multi-source domain adaptation},
  author={Peng, Xingchao and Bai, Qinxun and Xia, Xide and Huang, Zijun and Saenko, Kate and Wang, Bo},
  booktitle={Proceedings of the IEEE/CVF International Conference on Computer Vision},
  pages={1406--1415},
  year={2019}
}
\bibliographystyle{icml2026}

\newpage
\appendix
\onecolumn



{
\Large
\centering
\vspace{0.5em}
Supplementary Material \\ 
\vspace{1.5em}
\raggedright 
\textbf{Table of Contents} 
\vspace{1em}
\large
\begin{enumerate}[label=(\Alph*)]
    \item \textsc{Sparse Autoencoders} \hfill \pageref{appendix:sae_details}\\
    \item \textsc{Performance} \hfill \pageref{appendix:performance}\\
    \item \textsc{Match Rate for Different LLM Backbones and Visual Encoders} \hfill \pageref{appendix:matchrate_llms}\\
    \item \textsc{Implementation Details} \hfill \pageref{appendix:implm_details}
    \item \textsc{Reconstruction and Sparsity Comparisons} \hfill \pageref{appendix:reconst_spars}
    \item \textsc{Active Visual Tokens across Layers} \hfill \pageref{appendix:active_vtokens}
    \item \textsc{Qualitative Examples of Visual Steering with CLIP and DINOv2} \hfill \pageref{appendix:qualitative_steering}
    \item \textsc{Performance-Interpretability Trade-off} \hfill \pageref{appendix:perf_interp}
    \item \textsc{Spatial Localization with Different LLM Backbones} \hfill \pageref{appendix:spatial_loc_abl}\\
    \item \textsc{Steering with Different LLM Backbones} \hfill \pageref{appendix:steering_backbones_abl}\\
    \item \textsc{OOD Robustness on DomainNet} \hfill \pageref{appendix:ood_robustness}\\
    \item \textsc{Visual vs Text-Only Steering Comparison} \hfill \pageref{appendix:visual_vs_text_steering}\\

\end{enumerate}
}

\section{Sparse Autoencoders}
\label{appendix:sae_details}

Internal representations in large language models are difficult to interpret because they are often polysemantic. This means a single neuron might activate for many unrelated concepts, a phenomenon known as superposition \cite{elhage2022superposition, bricken2023monosemanticity}. To solve this, sparse autoencoders (SAEs) are used to break down these polysemantic activations into monosemantic features.

SAEs take an activation vector $x$ from the model, such as the residual stream, and processes it through two stages - encoding and decoding. The encoder $E(\cdot)$ maps the input into a high-dimensional, sparse latent space $E(x) \in \mathbb{R}^{d_{SAE}}$. This space is ``overcomplete'' meaning it has many more dimensions than the original input, typically 10 to 100 times more. The encoder uses a ReLU activation, or other alternatives like JumpReLU activation \cite{rajamanoharan_2024_jumping}, to ensure that feature values are non-negative:
\begin{equation}
    E(x) = \text{ReLU}(W_{enc}(x - b_{dec}) + b_{enc})
\end{equation}
where $W_{enc}$ represents the encoder weights and $b_{enc}, b_{dec}$ are learned biases.

The decoder $D(\cdot)$ then attempts to reconstruct the original activation $x$ by linearly combining the discovered sparse features $E(x)$:
\begin{equation}
    \hat{x} = D(E(x)) = W_{dec}E(x) + b_{dec}
\end{equation}
Each column in the decoder weight matrix $W_{dec}$ often represents a monosemantic feature that corresponds to a specific, human-interpretable concept.

The SAE is trained to minimize a dual-objective loss function consisting of a reconstruction error and an $L_1$ sparsity penalty:
\begin{equation}
    \mathcal{L} = |x - \hat{x}|_2^2 + \lambda |E(x)|_1
\end{equation}
The first term ensures that the SAE preserves the information content of the original activations, requiring the reconstructed version $\hat{x}$ to be as close to the original $x$ as possible. The second term, scaled by the hyperparameter $\lambda$, forces the model to represent the data using the smallest possible subset of features. By enforcing this sparsity, the SAE is forced to disentangle the overlapping concepts within the residual stream into discrete, monosemantic latent directions.

\clearpage
\section{Performance}
\label{appendix:performance}

\begin{table}[h]
\centering
\caption{Performance comparison of different visual encoders with and without SAE constraints on Gemma-2-2B-it.}
\label{tab:model_comparison}
\resizebox{0.64\columnwidth}{!}{%
\begin{tabular}{@{}llcc cc cc@{}}
\toprule
& & \multicolumn{2}{c}{\textbf{CLIP-Gemma-2-2B}} & \multicolumn{2}{c}{\textbf{DINOv2-Gemma-2-2B}} & \multicolumn{2}{c}{\textbf{I-JEPA-Gemma-2-2B}} \\
\cmidrule(lr){3-4} \cmidrule(lr){5-6} \cmidrule(lr){7-8}
\textbf{Benchmark} & \textbf{Metric} & \textbf{No SAE} & \textbf{SAE} & \textbf{No SAE} & \textbf{SAE} & \textbf{No SAE} & \textbf{SAE} \\
\midrule
MME & Cognition & 243.92 & 251.42 & 261.78 & 226.42 & 218.21 & 219.64 \\
& Perception & 993.73 & 942.89 & 942.13 & 881.30 & 907.13 & 719.31 \\
\midrule
LLaVA Bench & Conversation & 53.5 & 47.3 & 52.5 & 59.7 & 49.6 & 41.5 \\
& Complex & 32.1 & 32.6 & 36.2 & 33.3 & 30.6 & 27.9 \\
& Detailed & 30.0 & 24.0 & 29.4 & 30.4 & 26.7 & 21.8 \\
& All & 38.6 & 34.5 & 39.2 & 41.1 & 35.6 & 30.3 \\
\midrule
GQA & Exact Match & 42.95 & 41.00 & 43.50 & 42.10 & 40.46 & 39.20 \\
\midrule
\multirow{4}{*}{POPE} 
    & Accuracy & 81.6 & 79.3 & 76.63 & 76.33 & 74.1 & 73.57 \\
    & Precision & 86.8 & 94.2 & 88.16 & 79.74 & 70.2 & 83.19 \\
    & Recall & 74.53 & 62.4 & 61.53 & 70.6 & 83.6 & 59.07 \\
    & F1-score & 80.2 & 75.1 & 72.48 & 74.89 & 76.4 & 69.08 \\
\bottomrule
\end{tabular}
}
\end{table}

\begin{table}[!h]
\centering
\caption{Performance comparison of different visual encoders with and without SAE constraints on LLaMA-3.1-8B-Instruct.}
\resizebox{0.64\textwidth}{!}{%
\begin{tabular}{@{}llcccccc@{}}
\toprule
 &  & \multicolumn{2}{c}{\textbf{CLIP-LLaMA-3.1-8B}} & \multicolumn{2}{c}{\textbf{DINOv2-LLaMA-3.1-8B}} & \multicolumn{2}{c}{\textbf{I-JEPA-LLaMA-3.1-8B}} \\
\cmidrule(lr){3-4} \cmidrule(lr){5-6} \cmidrule(lr){7-8}
\textbf{Benchmark} & \textbf{Metric} & \textbf{No SAE} & \textbf{SAE} & \textbf{No SAE} & \textbf{SAE} & \textbf{No SAE} & \textbf{SAE} \\
\midrule
\multirow{2}{*}{MME} 
    & Cognition & 260.71 & 295.00 & 286.07 & 302.50 & 238.93 & 257.50 \\
    & Perception & 996.06 & 1062.32 & 1015.27 & 1175.09 & 883.94 & 972.33 \\
\midrule
\multirow{4}{*}{LLaVA Bench} 
    & Conversation & 59.9 & 67.9 & 56.8 & 69.1 & 44.5 & 60.7 \\
    & Complex & 35.9 & 33.5 & 25.9 & 29.5 & 33.6 & 31.5 \\
    & Detailed & 27.6 & 31.7 & 27.3 & 30.1 & 19.3 & 31.0 \\
    & All & 40.9 & 44.1 & 36.6 & 42.6 & 32.3 & 41.0 \\
\midrule
GQA  & Exact Match & 41.31 & 44.47 & 41.03 & 51.61 & 39.64 & 44.46 \\
\midrule
\multirow{4}{*}{POPE} 
    & Accuracy & 76.27 & 79.23 & 76.63 & 84.17 & 75.60 & 77.73 \\
    & Precision & 76.09 & 78.31 & 72.79 & 87.99 & 71.79 & 77.19 \\
    & Recall & 76.96 & 80.87 & 85.07 & 79.13 & 84.33 & 78.73 \\
    & F1-score & 76.52 & 79.57 & 78.45 & 83.33 & 77.56 & 77.95 \\
\bottomrule
\end{tabular}}
\label{tab:comparison_llama_encoders}
\end{table}

\begin{table}[h]
\centering
\caption{Performance comparison of different visual encoders with and without SAE constraints on Gemma-2-9B-it.}
\resizebox{0.64\textwidth}{!}{%
\begin{tabular}{@{}llcccccc@{}}
\toprule
 &  & \multicolumn{2}{c}{\textbf{CLIP-Gemma-2-9B}} & \multicolumn{2}{c}{\textbf{DINOv2-Gemma-2-9B}} & \multicolumn{2}{c}{\textbf{I-JEPA-Gemma-2-9B}} \\
\cmidrule(lr){3-4} \cmidrule(lr){5-6} \cmidrule(lr){7-8}
\textbf{Benchmark} & \textbf{Metric} & \textbf{No SAE} & \textbf{SAE} & \textbf{No SAE} & \textbf{SAE} & \textbf{No SAE} & \textbf{SAE} \\
\midrule
\multirow{2}{*}{MME} 
    & Cognition & 310.71 & 259.29 & 275.36 & 295.36 & 248.57 & 301.07 \\
    & Perception & 1198.37 & 1042.00 & 1170.90 & 1085.64 & 1063.82 & 1032.49 \\
\midrule
\multirow{4}{*}{LLaVA Bench} 
    & Conversation & 63.4 & 65.4 & 58.9 & 70.2 & 52.2 & 47.4 \\
    & Complex & 34.1 & 34.0 & 30.7 & 32.0 & 33.1 & 34.7 \\
    & Detailed & 33.8 & 32.4 & 33.1 & 32.3 & 27.8 & 20.8 \\
    & All & 43.9 & 42.7 & 41.0 & 44.2 & 37.7 & 34.2 \\
\midrule
GQA  & Exact Match & 49.62 & 44.16 & 47.43 & 47.27 & 44.85 & 42.44 \\
\midrule
\multirow{4}{*}{POPE} 
    & Accuracy & 83.13 & 76.40 & 83.77 & 83.17 & 78.73 & 77.80 \\
    & Precision & 85.55 & 75.19 & 87.32 & 87.95 & 79.40 & 76.44 \\
    & Recall & 79.73 & 78.80 & 79.16 & 76.87 & 77.60 & 81.68 \\
    & F1-score & 82.54 & 76.95 & 83.04 & 82.03 & 78.49 & 78.97 \\
\bottomrule
\end{tabular}}
\label{tab:comparison_gemma_encoders}
\end{table}

We evaluate the impact of SAE constraints on projector training across four benchmarks: MME (perception and cognition) \cite{fu2025mme}, GQA (compositional reasoning) \cite{hudson2019gqa}, POPE (hallucination) \cite{pope_bench}, and LLaVA-Bench (instruction following) \cite{liu_llava_2023}. As shown in \cref{tab:model_comparison}, enforcing interpretability constraints typically induces a marginal accuracy regression in smaller models. We hypothesize that forcing visual tokens to align strictly with text features taxes the limited representational capacity of smaller architectures. However, this effect is not universal; for instance, the DINOv2 encoder with SAE alignment yields slight improvements on LLaVA-Bench and POPE F1 scores. 

Scaling the LLM backbone fundamentally alters how the model handles these interpretability constraints (\cref{tab:comparison_gemma_encoders,tab:comparison_llama_encoders}). For the LLaMA-3.1-8B-Instruct architecture, enforcing SAE constraints results in performance gains across the majority of metrics for all visual encoders. This suggests that for certain larger architectures, SAE interpretability constraints can act as a beneficial regularizer for cross-modal alignment. Conversely, the Gemma-2-9B-it backbone demonstrates highly encoder-dependent results. While DINOv2 combined with the SAE constraint maintains competitive performance, improving in cognition and LLaVA-Bench while slightly regressing in perception, the CLIP encoder suffers a degradation across almost all metrics when the SAE is introduced. The addition of the SAE to the I-JEPA encoder yields mixed results across the benchmarks for Gemma-2-9B-it backbone.


\section{Match Rate for Different LLM Backbones and Visual Encoders}
\label{appendix:matchrate_llms}



\begin{figure}[!h]
    \centering
    
    \begin{subfigure}[b]{.7\textwidth}
        \centering
        \includegraphics[width=\textwidth]{figures/matchrate_dinov2.pdf}
        \caption{DINOv2 match rate for visual tokens with and without SAE constraints with Gemma-2-2B-it LLM model. }
        \label{fig:matchrate_dinov2_gemma2b}
    \end{subfigure}

    \begin{subfigure}[b]{.7\textwidth}
        \centering
        \includegraphics[width=\textwidth]{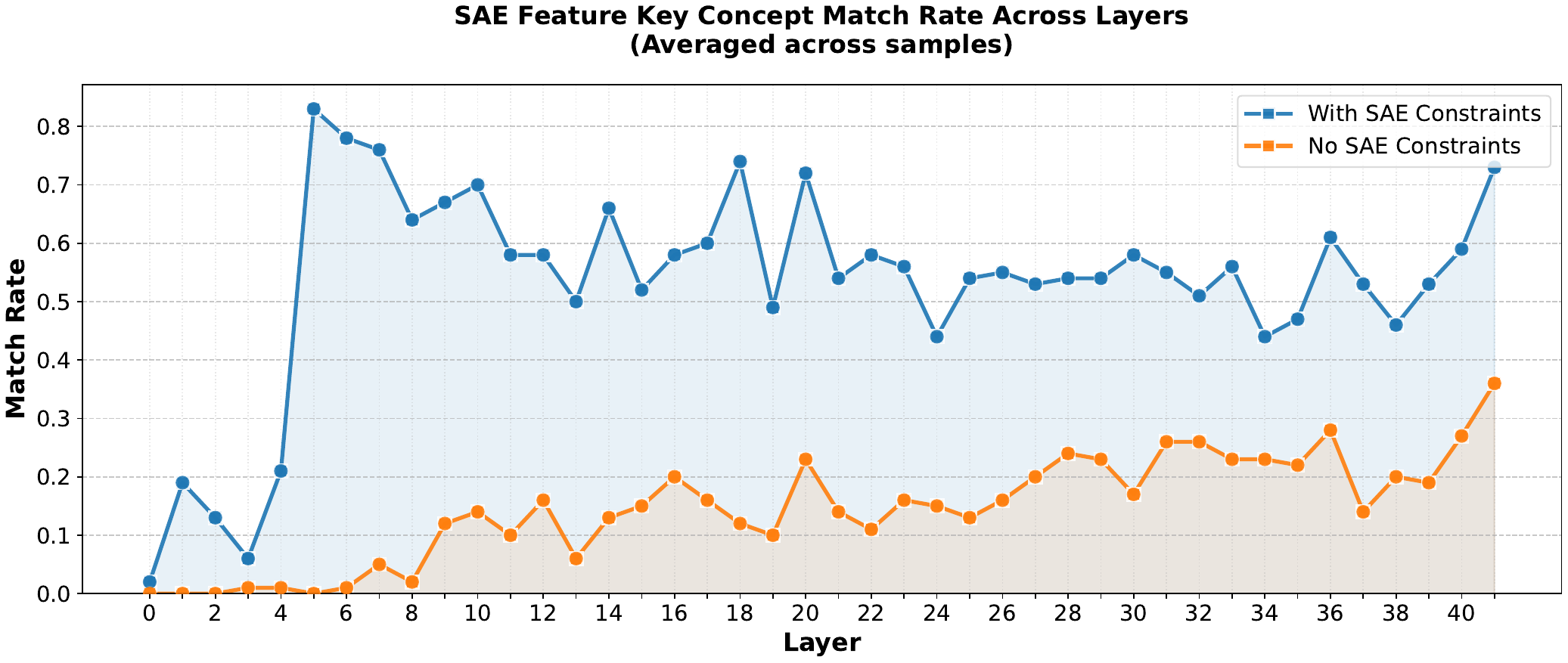}
        \caption{DINOv2 match rate for visual tokens with and without SAE constraints with Gemma-2-9B-it model. }
        \label{fig:matchrate_dinov2_gemma9b}
    \end{subfigure}

    \begin{subfigure}[b]{.7\textwidth}
        \centering
        \includegraphics[width=\textwidth]{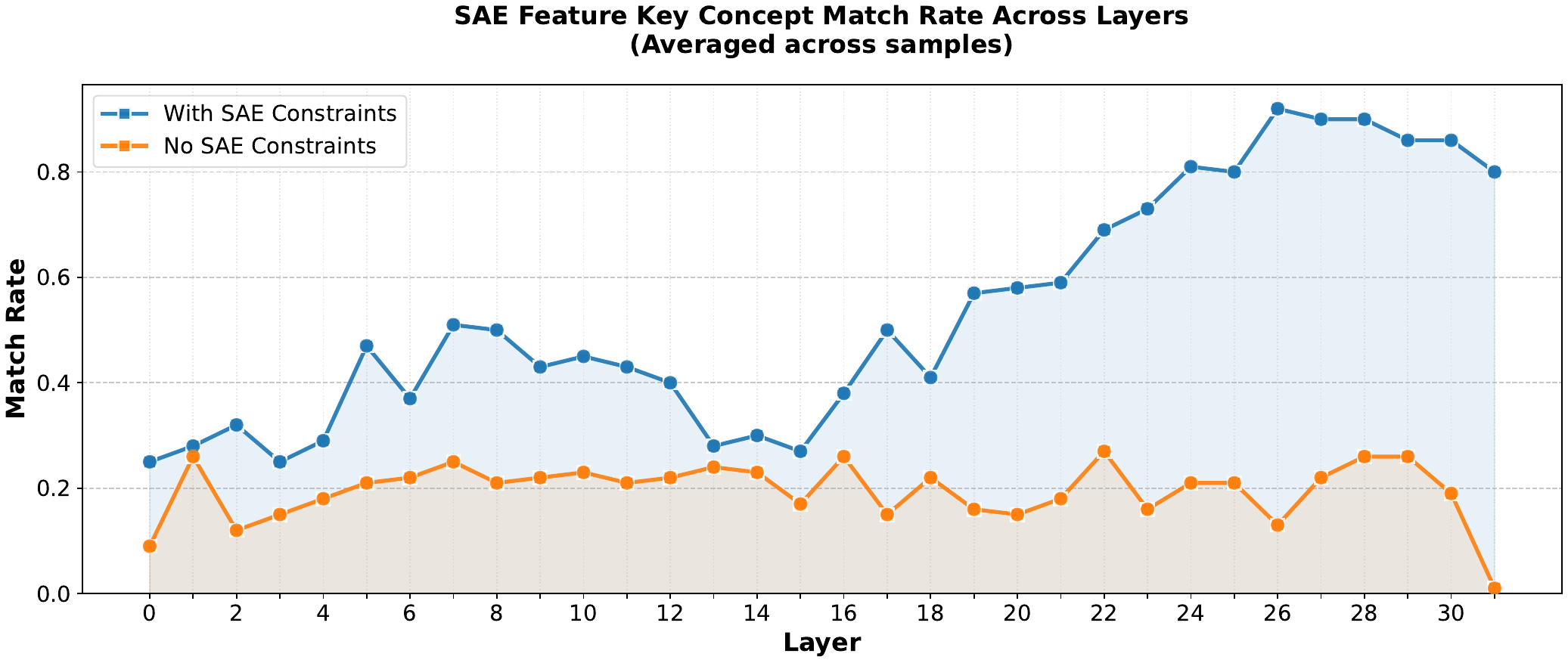}
        \caption{DINOv2 match rate for visual tokens with and without SAE constraints with LLaMA-3.1-8B-Instruct model. }
        \label{fig:matchrate_dinov2_llama}
    \end{subfigure}

    \caption{Matching rate for DINOv2 visual encoder.}
\end{figure}

\begin{figure}[!h]
    \centering
    
    \begin{subfigure}[b]{.7\textwidth}
        \centering
        \includegraphics[width=\textwidth]{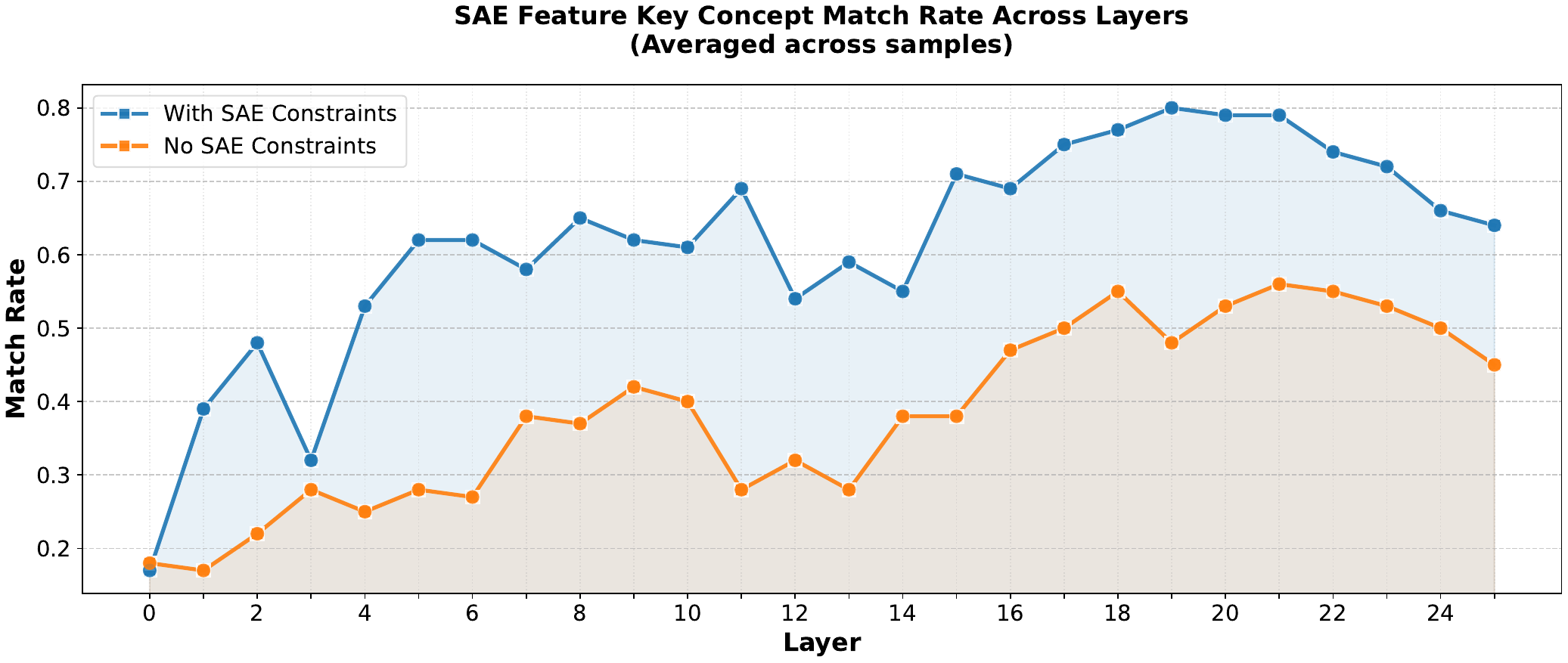}
        \caption{CLIP match rate for visual tokens with and without SAE constraints with Gemma-2-2B-it LLM model. }
        \label{fig:matchrate_clip_gemma2b}
    \end{subfigure}

    \begin{subfigure}[b]{.7\textwidth}
        \centering
        \includegraphics[width=\textwidth]{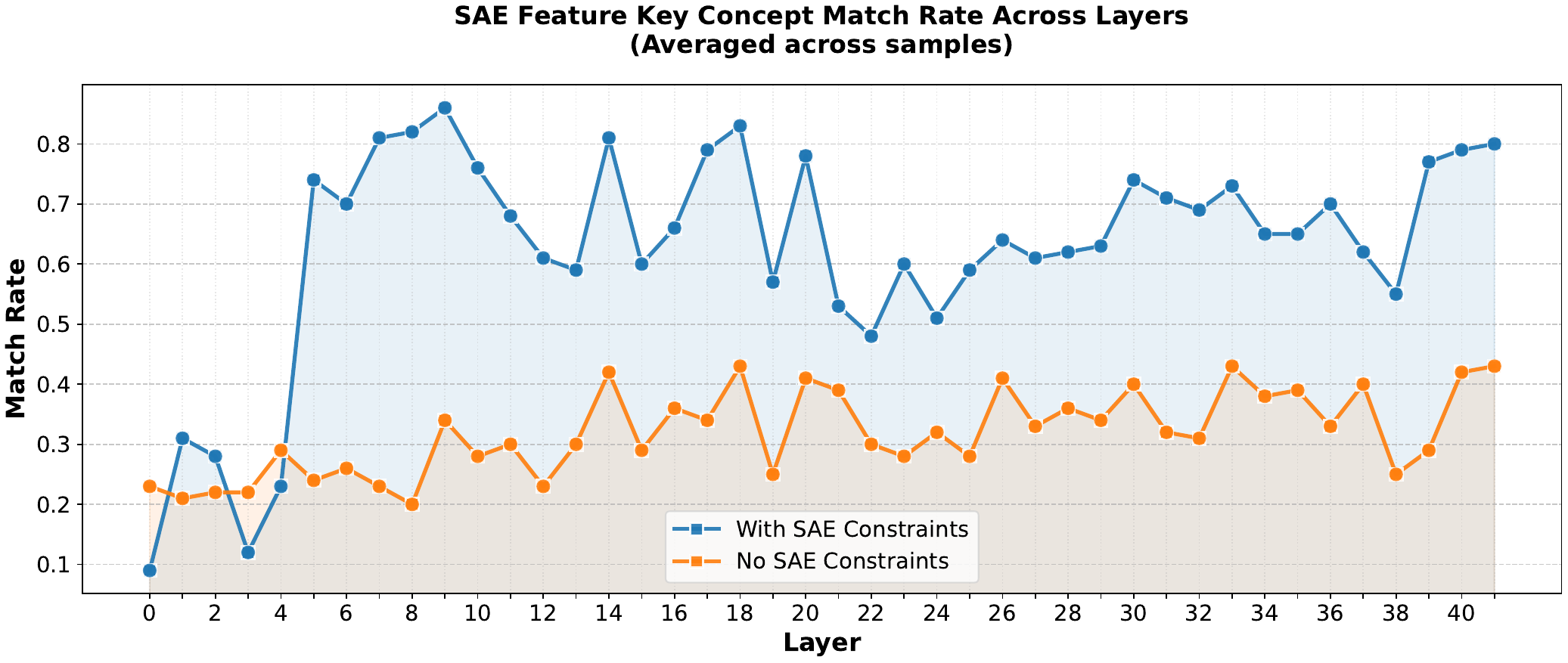}
        \caption{CLIP match rate for visual tokens with and without SAE constraints with Gemma-2-9B-it model. }
        \label{fig:matchrate_clip_gemma9b}
    \end{subfigure}

    \begin{subfigure}[b]{.7\textwidth}
        \centering
        \includegraphics[width=\textwidth]{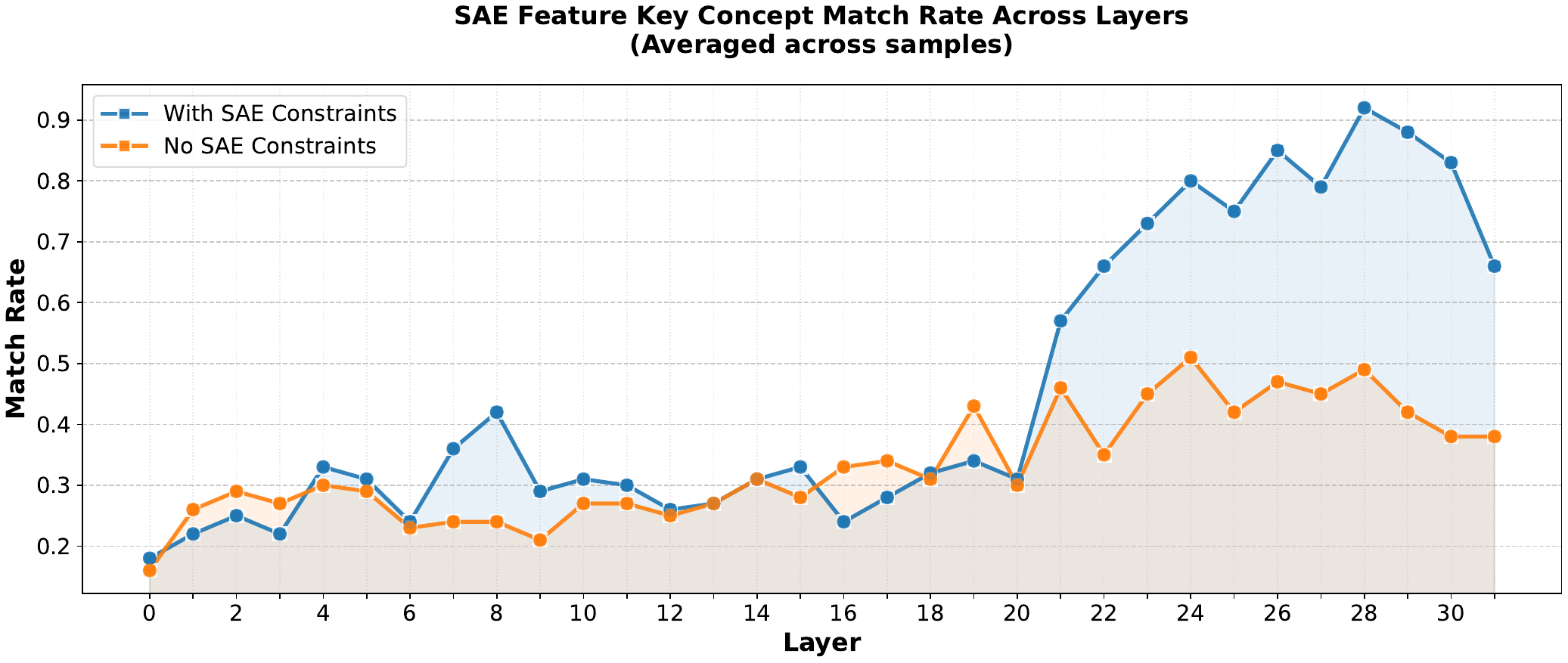}
        \caption{CLIP match rate for visual tokens with and without SAE constraints with LLaMA-3.1-8B-Instruct model. }
        \label{fig:matchrate_clip_llama}
    \end{subfigure}

    \caption{Matching rate for CLIP visual encoder.}
\end{figure}

\begin{figure}[!h]
    \centering
    
    \begin{subfigure}[b]{.7\textwidth}
        \centering
        \includegraphics[width=\textwidth]{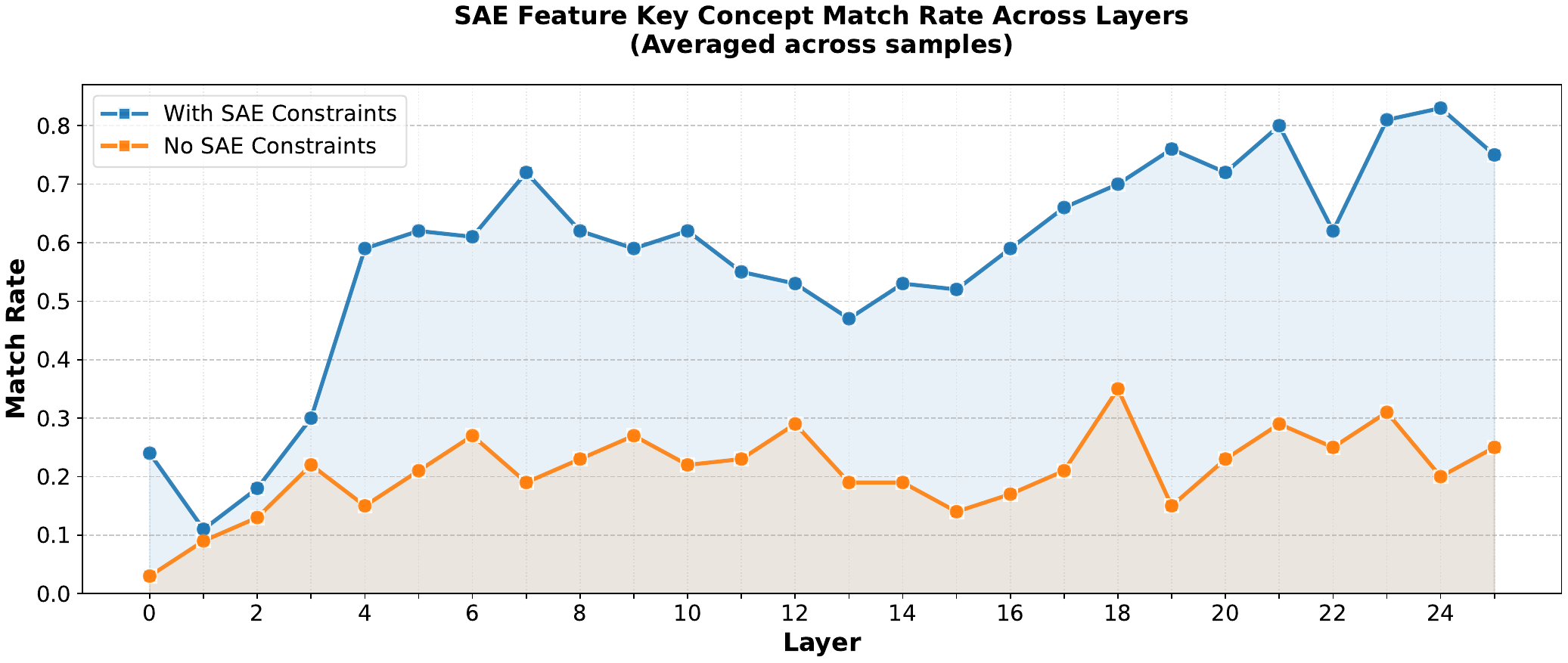}
        \caption{I-JEPA match rate for visual tokens with and without SAE constraints with Gemma-2-2B-it LLM model. }
        \label{fig:matchrate_jepa_gemma2b}
    \end{subfigure}

    \begin{subfigure}[b]{.7\textwidth}
        \centering
        \includegraphics[width=\textwidth]{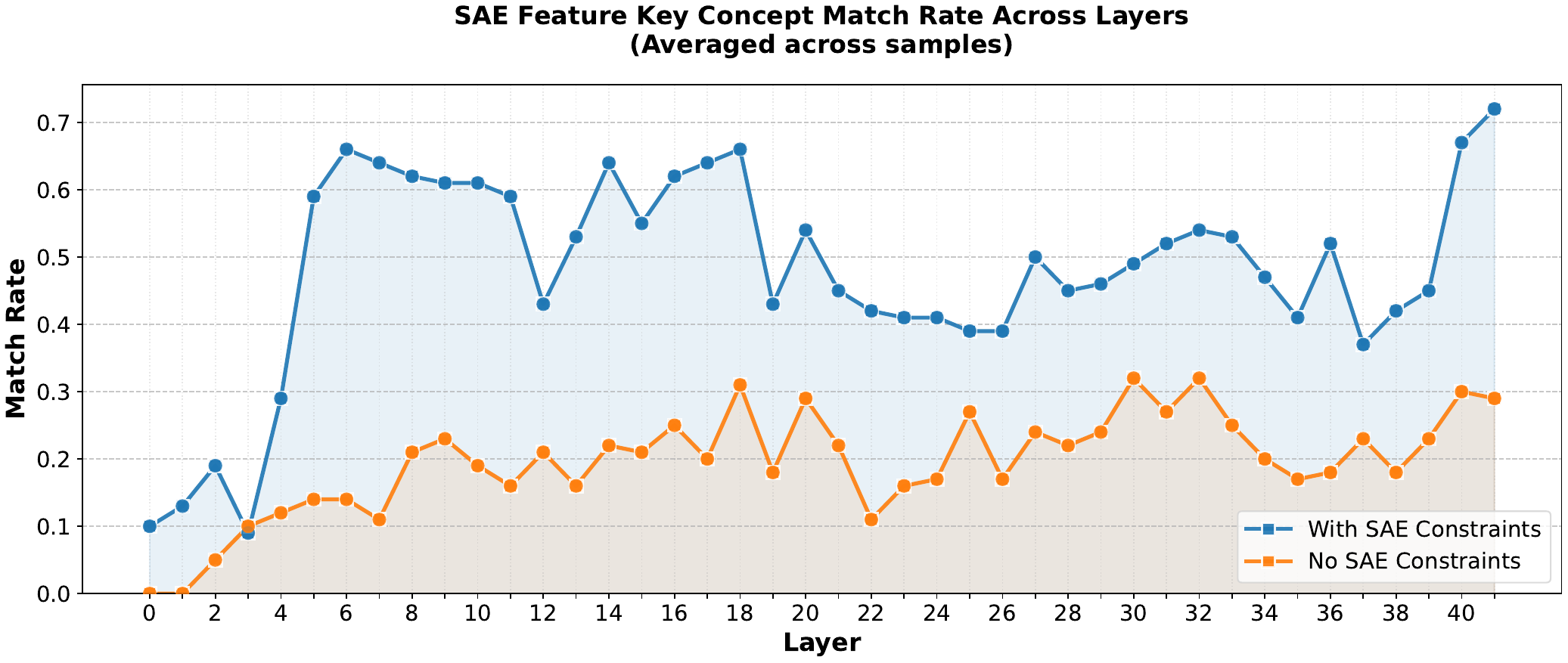}
        \caption{I-JEPA match rate for visual tokens with and without SAE constraints with Gemma-2-9B-it model. }
        \label{fig:matchrate_jepa_gemma9b}
    \end{subfigure}

    \begin{subfigure}[b]{.7\textwidth}
        \centering
        \includegraphics[width=\textwidth]{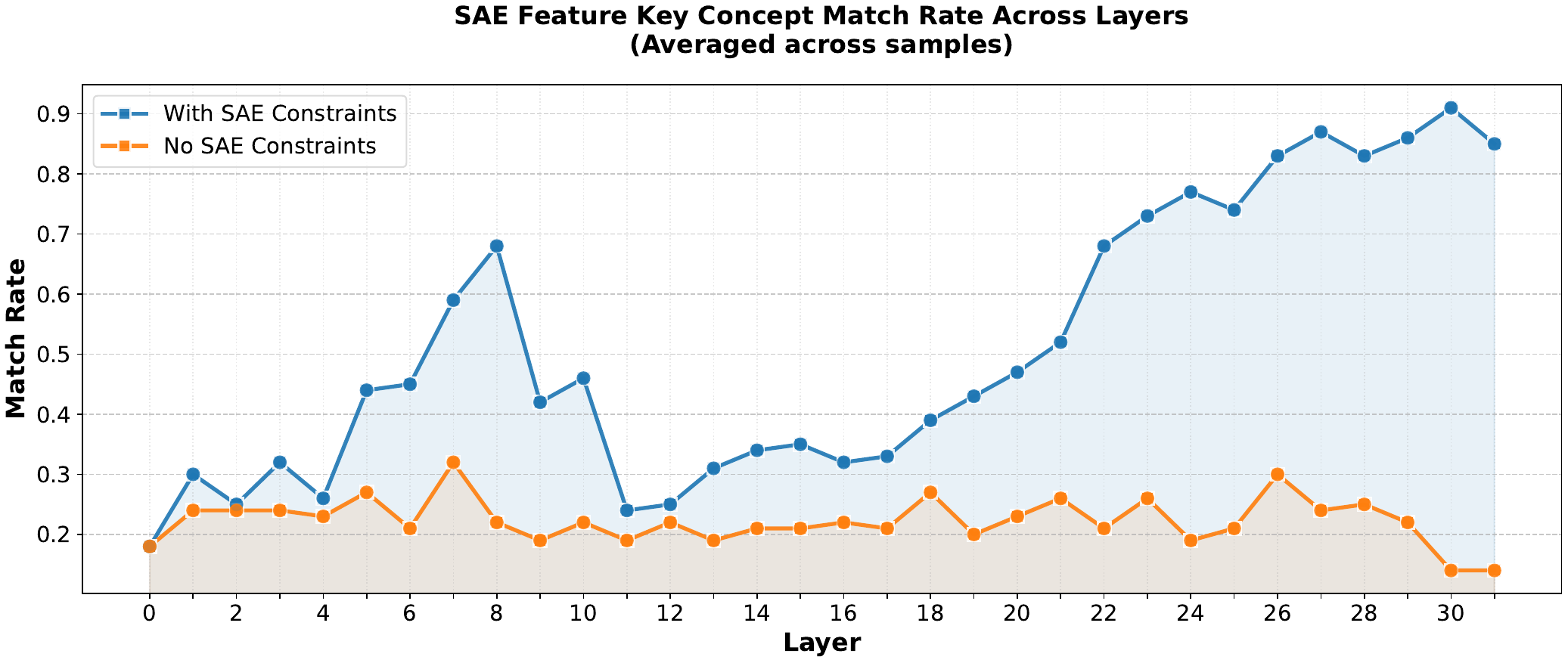}
        \caption{I-JEPA match rate for visual tokens with and without SAE constraints with LLaMA-3.1-8B-Instruct model. }
        \label{fig:matchrate_jepa_llama}
    \end{subfigure}

    \caption{Matching rate for I-JEPA visual encoder.}
\end{figure}

\clearpage
\section{Implementation Details}
\label{appendix:implm_details}

\paragraph{Models and Architecture} Our experiments utilize pre-trained SAEs from \cite{lieberum2024gemmascopeopensparse} and LLaMA Scope \cite{he2024llamascope}. We analyze the residual stream of Gemma 2 (2B and 9B) using $16k$-width SAEs, and for LLaMA, we employ the $8\times$ expansion factor SAEs. 

For the visual backbone, We employ visual encoders from three distinct model families:
\textbf{CLIP-ViT-L/14} \cite{radford2021LearningTV},
\textbf{DINOv2-ViT-L/14} \cite{oquab2024dinov}, and
\textbf{I-JEPA-ViT-H/14} \cite{Assran2023SelfSupervisedLF}. Each optimizes for different objectives: contrastive image-text alignment, self-supervised visual feature learning, and predictive visual modeling, respectively. All three models use a patch size of $14 \times 14$ and process $224 \times 224$ inputs into 256 visual patch tokens. While CLIP-ViT-L/14 and DINOv2-ViT-L/14 append an additional [CLS] token (resulting in 257 total tokens), I-JEPA-ViT-H/14 outputs only the 256 patch tokens.
This architectural consistency enables direct comparison within our unified representation framework.

\paragraph{Training and Computational Cost} The visual projector is trained in two stages while keeping the vision encoder and LLM frozen. The total computational requirement for full pre-training and fine-tuning is approximately 200 GPU hours for smaller LLM models and 290 GPU hours for larger architectures. We utilize A100 GPUs. For evaluations regarding Matching Rate, Reconstruction, and Sparsity, we utilize 1000 samples and employ the Gemini-2.5-flash-lite model. For Spatial Localization Accuracy calculation, we use 380 images sampled from the COCO dataset.

\paragraph{Steering and Intervention Parameters} We perform visual steering by intervening spatially on selected visual patches and temporally at every decoding step in the first 3 layers for Gemma models and the first 2 layers of LLaMA model, found empirically most effective. For both removal and replacement operations, we use a steering coefficient of 10. When training the projector with VISTA's SAE constraints, we apply the reconstruction loss to the first 5 layers of the LLM based on the performance-interpretability trade-off analysis in \cref{appendix:perf_interp}.

\section{Reconstruction and Sparsity Comparisons }
\label{appendix:reconst_spars}

\begin{figure}[!h]
    \centering
    
    \begin{subfigure}[b]{.9\textwidth}
        \centering
        \includegraphics[width=\textwidth]{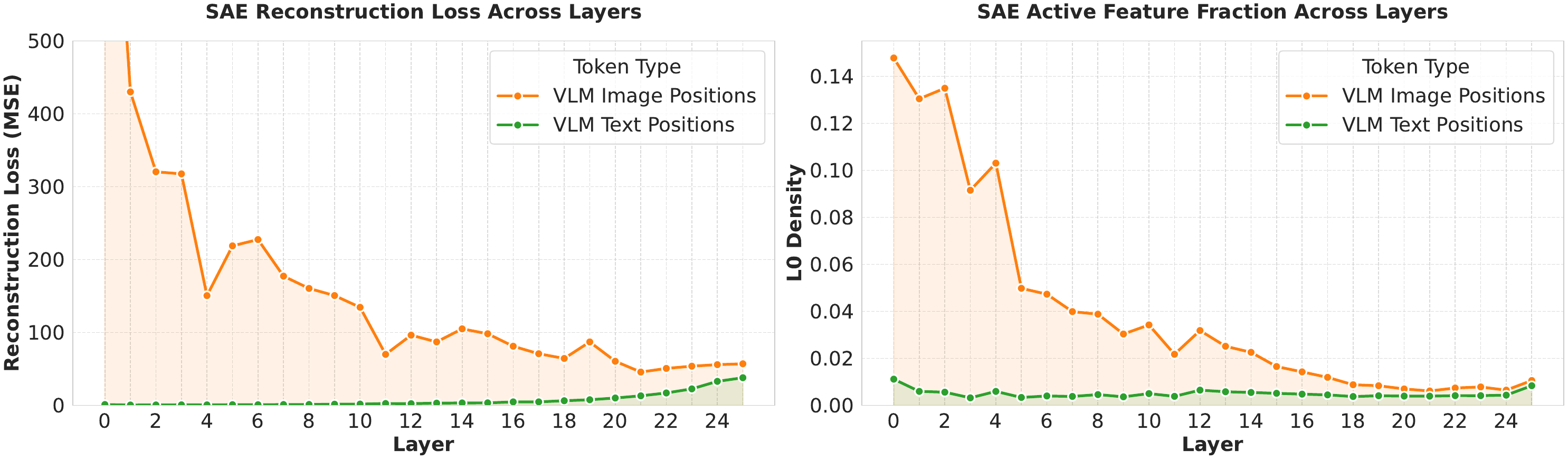}
        \caption{CLIP with Gemma-2-2B-it reconstruction without SAE constraints. }
        \label{fig:reconstr_sparsity_clip_wsae}
    \end{subfigure}
    \hfill  
    \begin{subfigure}[b]{.9\textwidth}
        \centering
        \includegraphics[width=\textwidth]{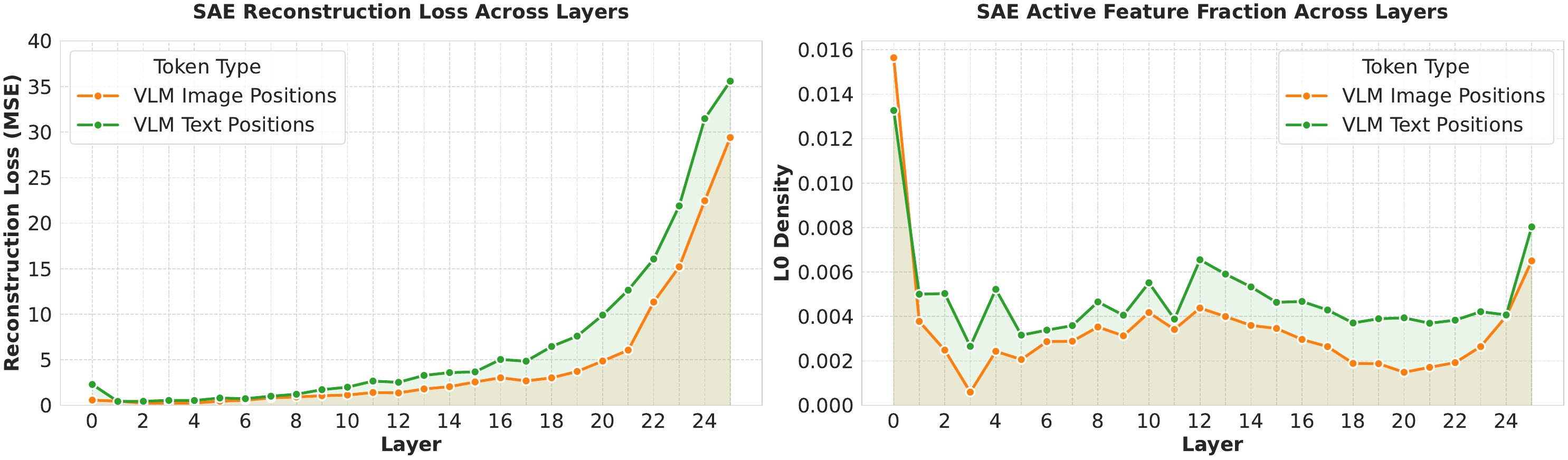}
        \caption{CLIP with Gemma-2-2B-it reconstruction with SAE constraints. }
        \label{fig:reconstr_sparsity_clip_nosae}
    \end{subfigure}

\caption{CLIP with Gemma-2-2B-it reconstruction and sparsity.}
\end{figure}

\begin{figure*}[!t]
    \centering
    
    \begin{subfigure}[b]{.9\textwidth}
        \centering
        \includegraphics[width=\textwidth]{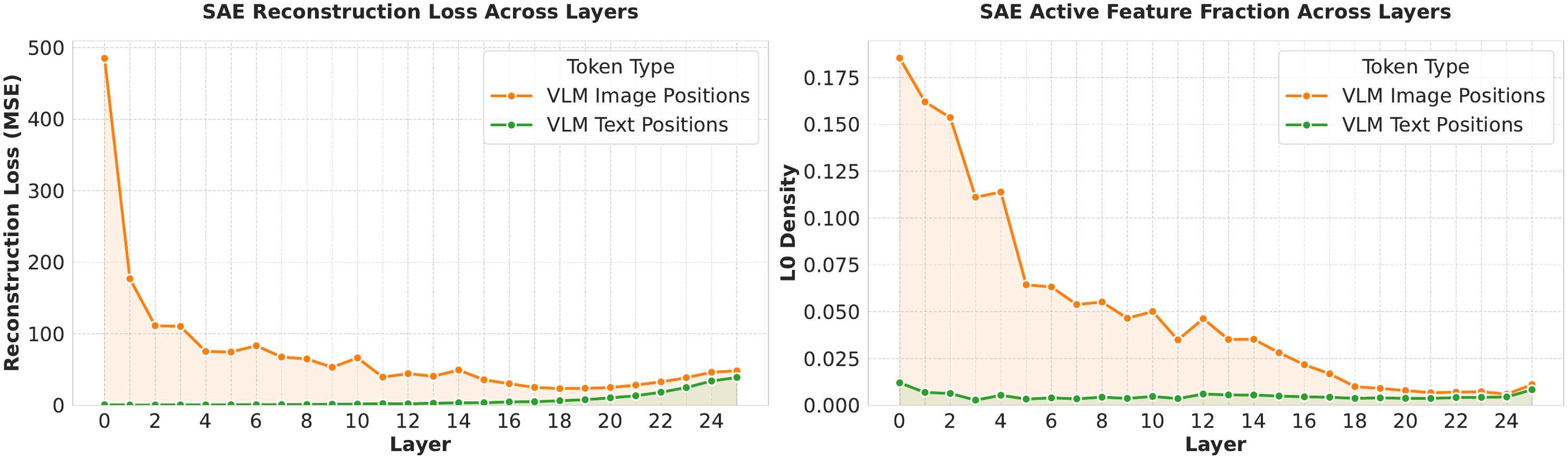}
        \caption{I-JEPA with Gemma-2-2B-it reconstruction without SAE constraints. }
        \label{fig:reconstr_sparsity_jepa_wsae}
    \end{subfigure}
    \hfill  
    \begin{subfigure}[b]{.9\textwidth}
        \centering
        \includegraphics[width=\textwidth]{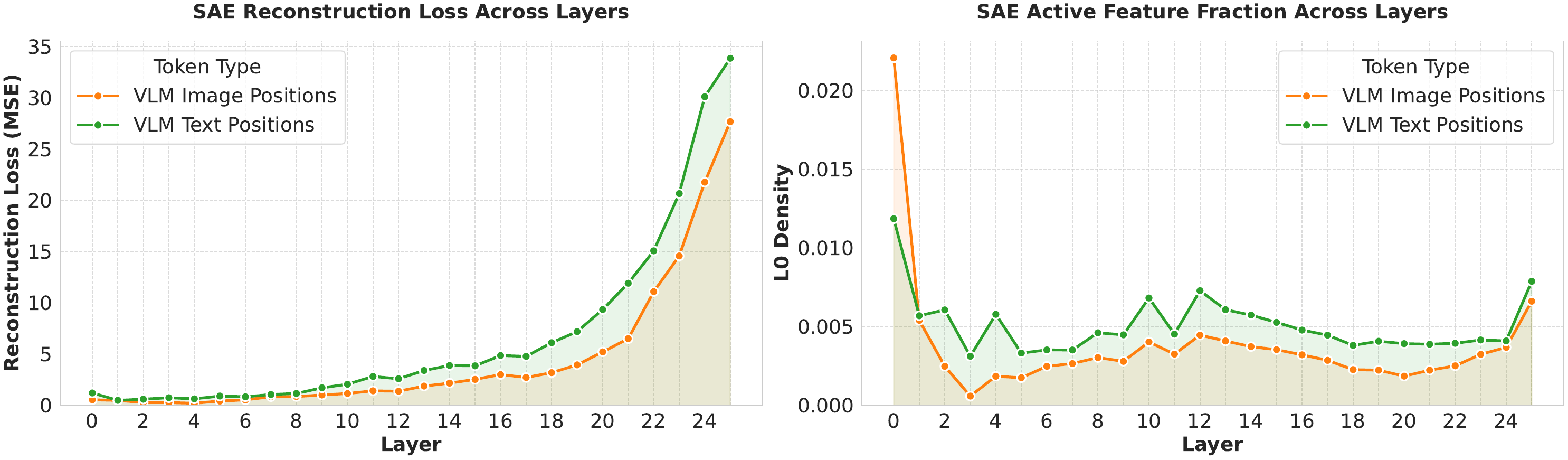}
        \caption{I-JEPA with Gemma-2-2B-it reconstruction with SAE constraints. }
        \label{fig:reconstr_sparsity_jepa_nosae}
    \end{subfigure}

\caption{I-JEPA with Gemma-2-2B-it reconstruction and sparsity.}

\end{figure*}

\begin{figure*}[!t]
    \centering
    
    \begin{subfigure}[b]{.9\textwidth}
        \centering
        \includegraphics[width=\textwidth]{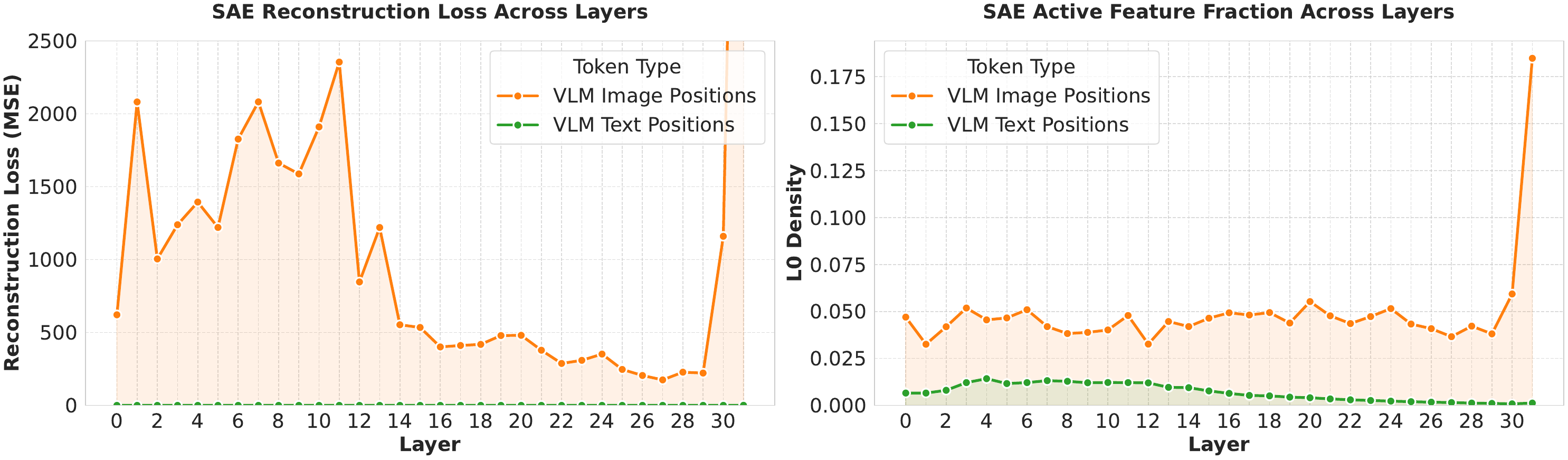}
        \caption{DINOv2 with LLaMA-3.1-8B-Instruct reconstruction without SAE constraints}
        \label{fig:reconstr_sparsity_dinov2_llama_nosae}
    \end{subfigure}
    \hfill  
    \begin{subfigure}[b]{.9\textwidth}
        \centering
        \includegraphics[width=\textwidth]{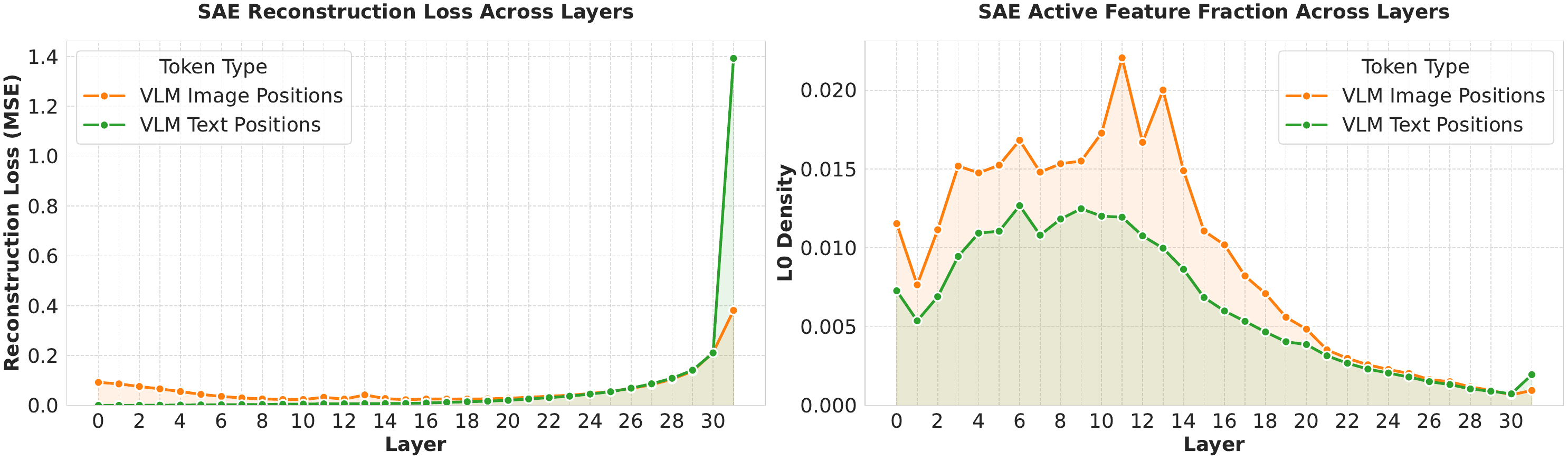}
        \caption{DINOv2 with LLaMA-3.1-8B-Instruct reconstruction with SAE constraints. }
        \label{fig:reconstr_sparsity_dinov2_llama_wsae}
    \end{subfigure}

\caption{DINOv2 with LLaMA-3.1-8B-Instruct reconstruction and sparsity.}
\end{figure*}

\begin{figure*}[!t]
    \centering
    
    \begin{subfigure}[b]{.9\textwidth}
        \centering
        \includegraphics[width=\textwidth]{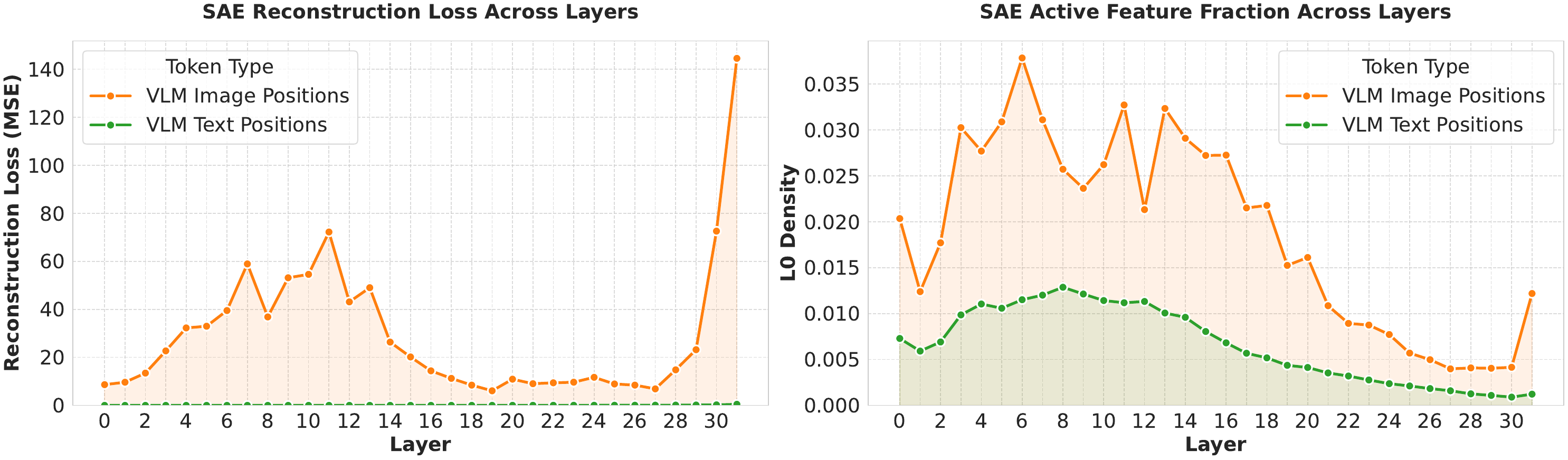}
        \caption{CLIP with LLaMA-3.1-8B-Instruct reconstruction without SAE constraints}
        \label{fig:reconstr_sparsity_clip_llama_nosae}
    \end{subfigure}
    \hfill  
    \begin{subfigure}[b]{.9\textwidth}
        \centering
        \includegraphics[width=\textwidth]{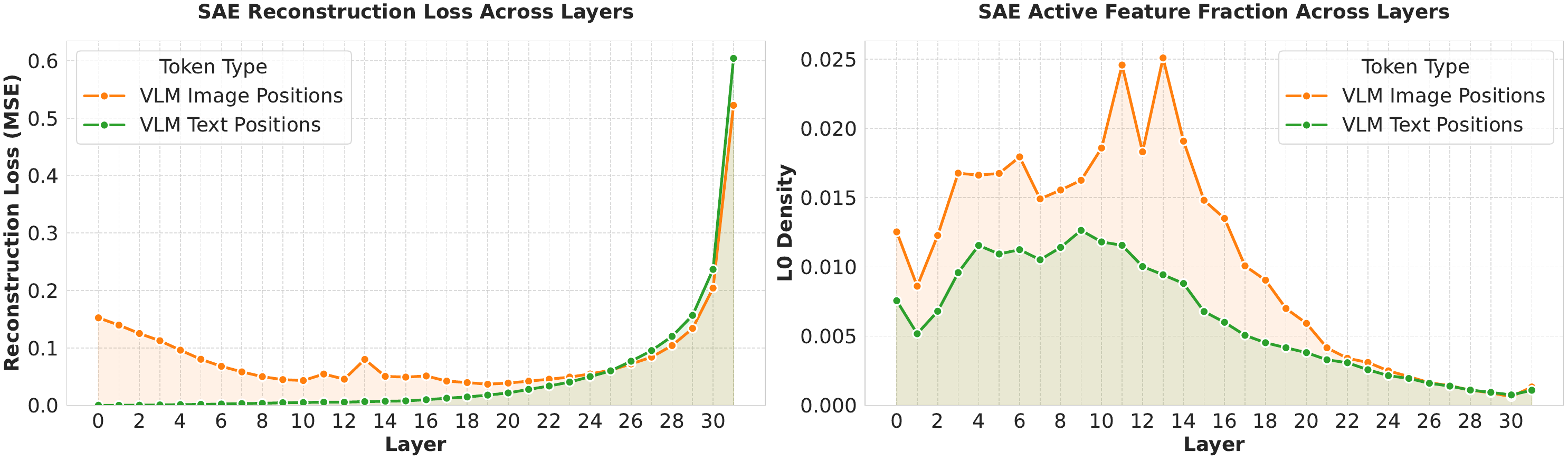}
        \caption{CLIP with LLaMA-3.1-8B-Instruct reconstruction with SAE constraints. }
        \label{fig:reconstr_sparsity_clip_llama_wsae}
    \end{subfigure}

\caption{CLIP with LLaMA-3.1-8B-Instruct reconstruction and sparsity.}
\end{figure*}

\begin{figure*}[!t]
    \centering
    
    \begin{subfigure}[b]{.9\textwidth}
        \centering
        \includegraphics[width=\textwidth]{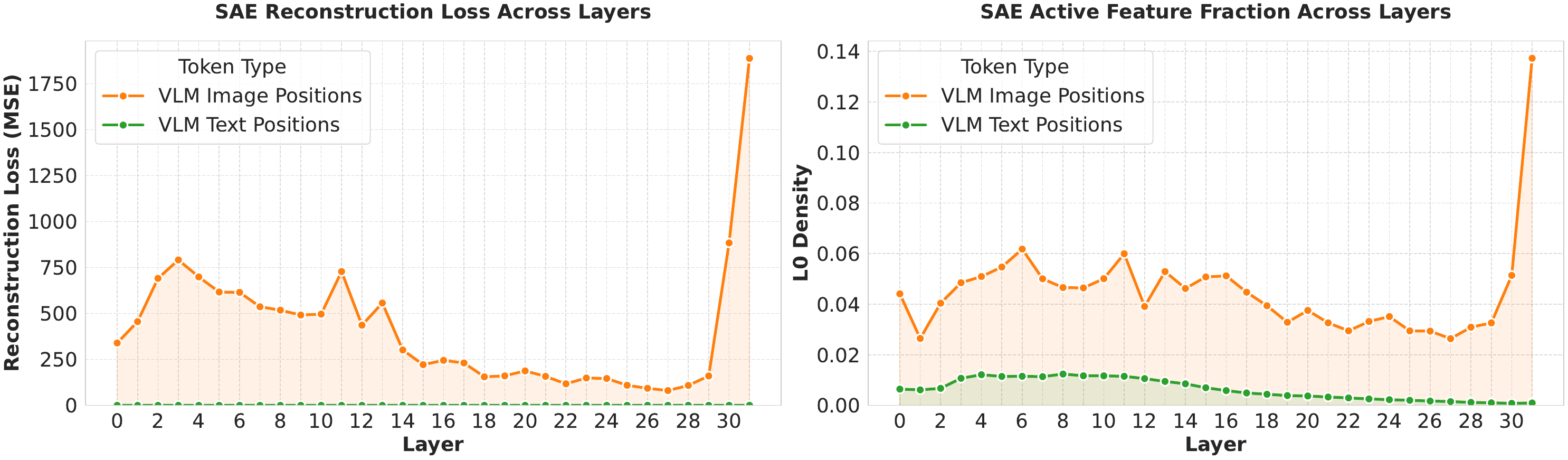}
        \caption{I-JEPA with LLaMA-3.1-8B-Instruct reconstruction without SAE constraints}
        \label{fig:reconstr_sparsity_jepa_llama_nosae}
    \end{subfigure}
    \hfill  
    \begin{subfigure}[b]{.9\textwidth}
        \centering
        \includegraphics[width=\textwidth]{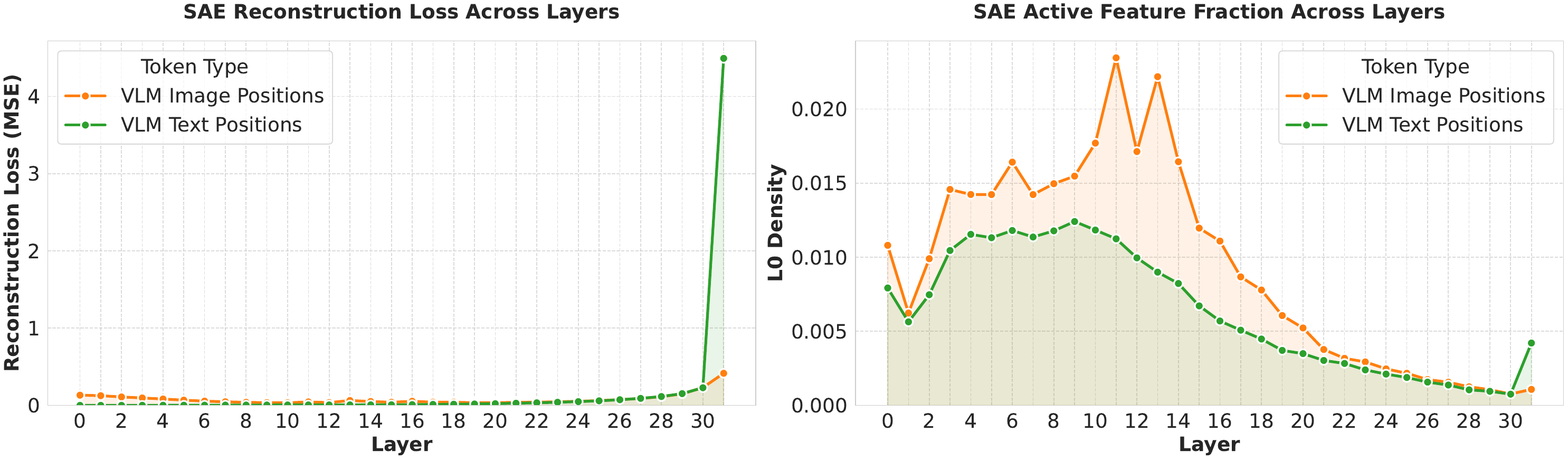}
        \caption{I-JEPA with LLaMA-3.1-8B-Instruct reconstruction with SAE constraints. }
        \label{fig:reconstr_sparsity_jepa_llama_wsae}
    \end{subfigure}

\caption{I-JEPA with LLaMA-3.1-8B-Instruct reconstruction and sparsity.}
\end{figure*}

\begin{figure*}[!t]
    \centering
    
    \begin{subfigure}[b]{.9\textwidth}
        \centering
        \includegraphics[width=\textwidth]{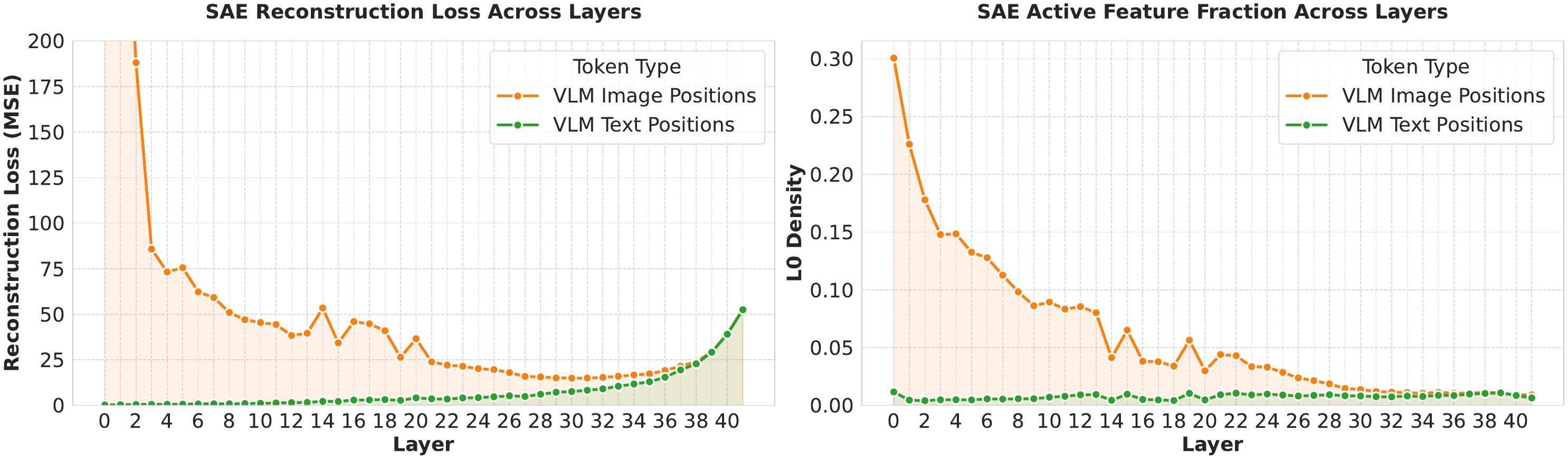}
        \caption{DINOv2 with Gemma-2-9B-it reconstruction without SAE constraints}
        \label{fig:reconstr_sparsity_dinov2_gemma9b_nosae}
    \end{subfigure}
    \hfill  
    \begin{subfigure}[b]{.9\textwidth}
        \centering
        \includegraphics[width=\textwidth]{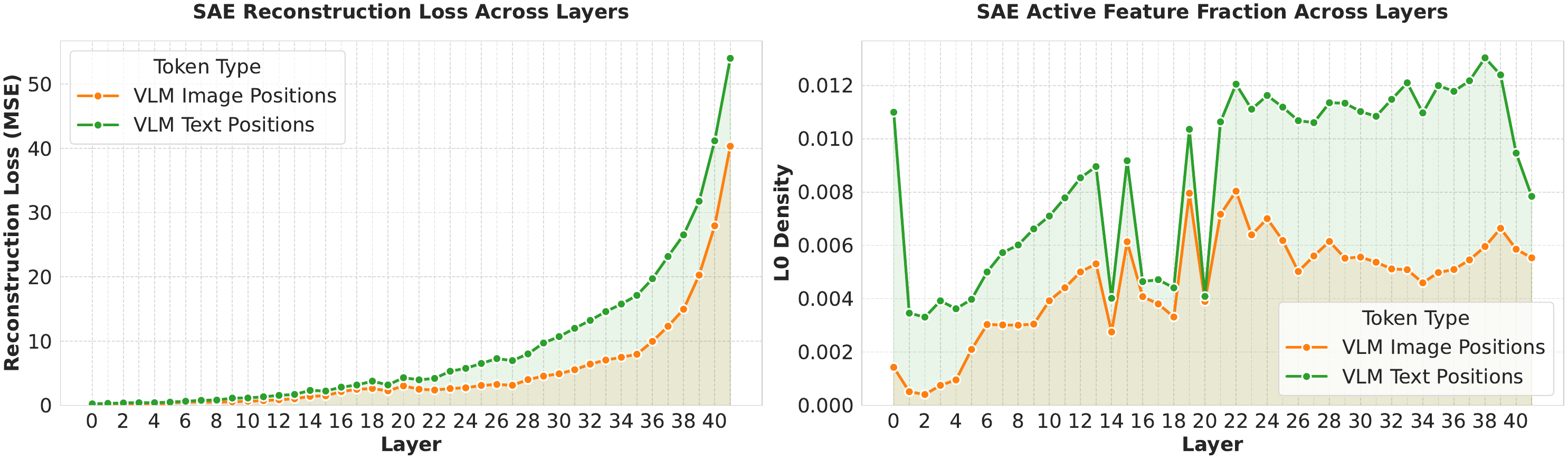}
        \caption{DINOv2 with Gemma-2-9B-it reconstruction with SAE constraints. }
        \label{fig:reconstr_sparsity_dinov2_gemma9b_wsae}
    \end{subfigure}

\caption{DINOv2 with Gemma-2-9B-it reconstruction and sparsity.}
\end{figure*}

\begin{figure*}[!t]
    \centering
    
    \begin{subfigure}[b]{.9\textwidth}
        \centering
        \includegraphics[width=\textwidth]{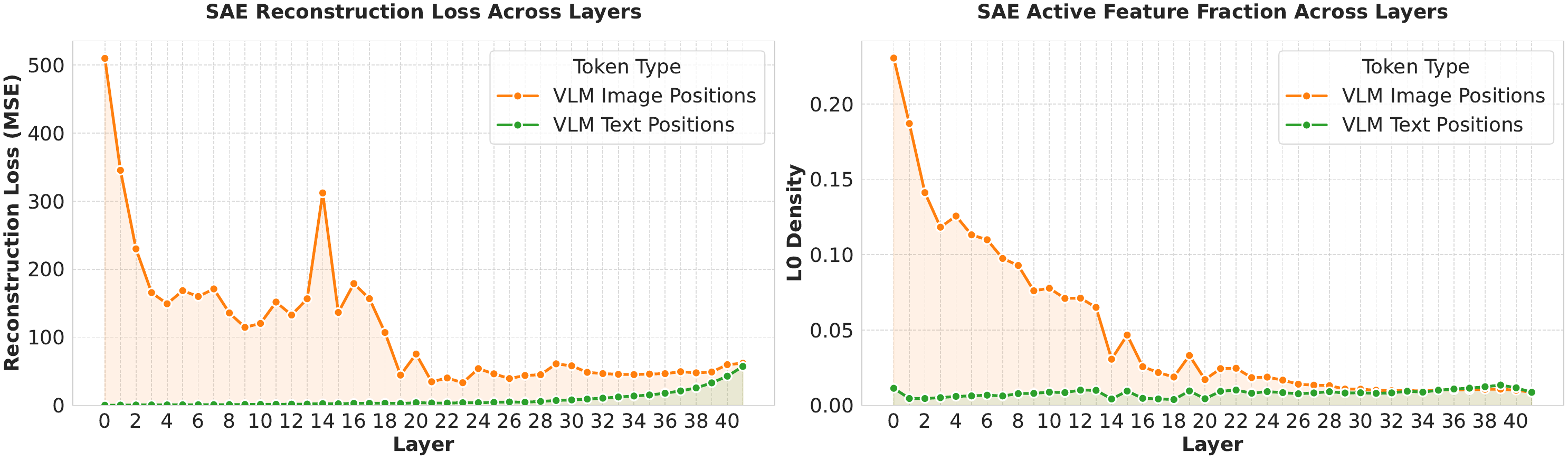}
        \caption{CLIP with Gemma-2-9B-it reconstruction without SAE constraints}
        \label{fig:reconstr_sparsity_clip_gemma9b_nosae}
    \end{subfigure}
    \hfill  
    \begin{subfigure}[b]{.9\textwidth}
        \centering
        \includegraphics[width=\textwidth]{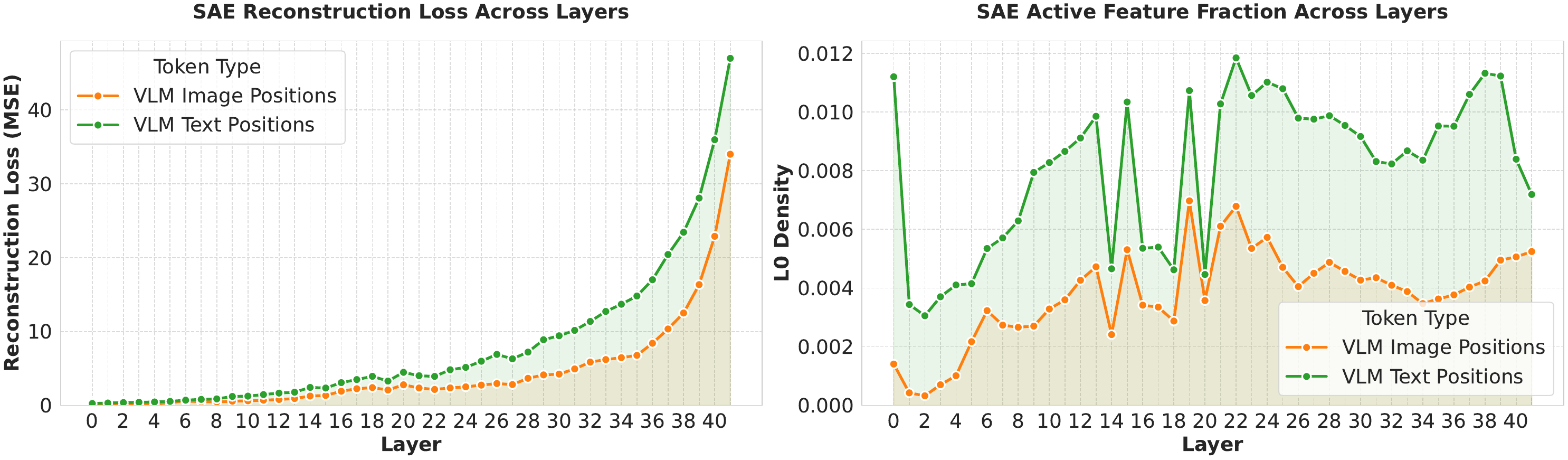}
        \caption{CLIP with Gemma-2-9B-it reconstruction with SAE constraints. }
        \label{fig:reconstr_sparsity_clip_gemma9b_wsae}
    \end{subfigure}

\caption{CLIP with Gemma-2-9B-it reconstruction and sparsity.}
\end{figure*}

\begin{figure*}[!t]
    \centering
    \begin{subfigure}[b]{.9\textwidth}
        \centering
        \includegraphics[width=\textwidth]{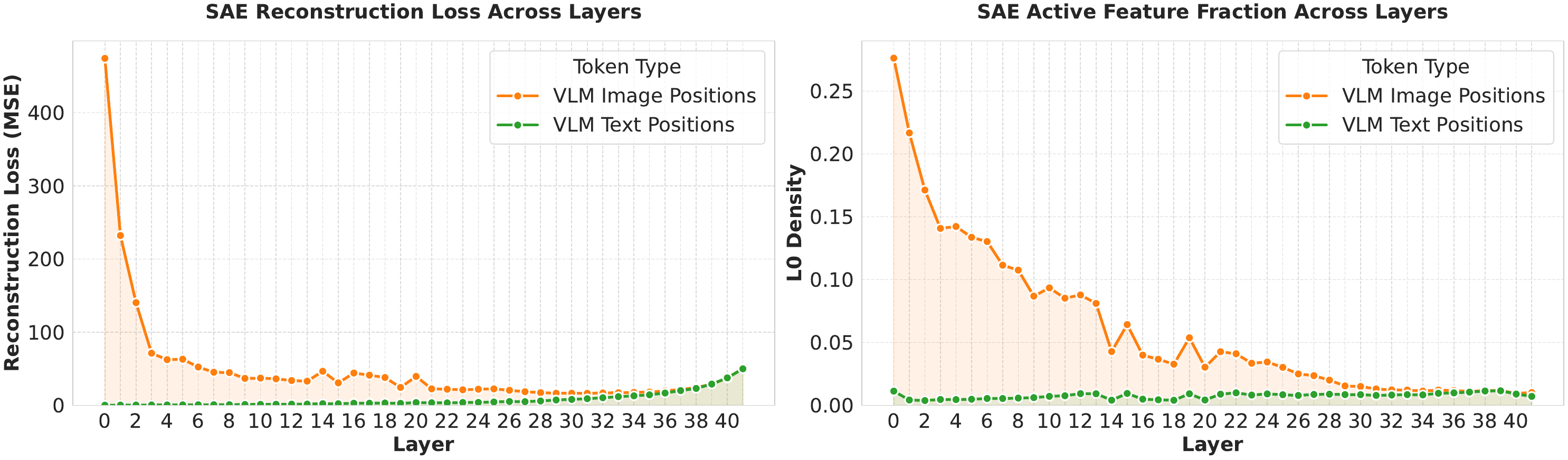}
        \caption{JEPA with Gemma-2-9B-it reconstruction without SAE constraints}
        \label{fig:reconstr_sparsity_jepa_gemma9b_nosae}
    \end{subfigure}
    \hfill  
    \begin{subfigure}[b]{.9\textwidth}
        \centering
        \includegraphics[width=\textwidth]{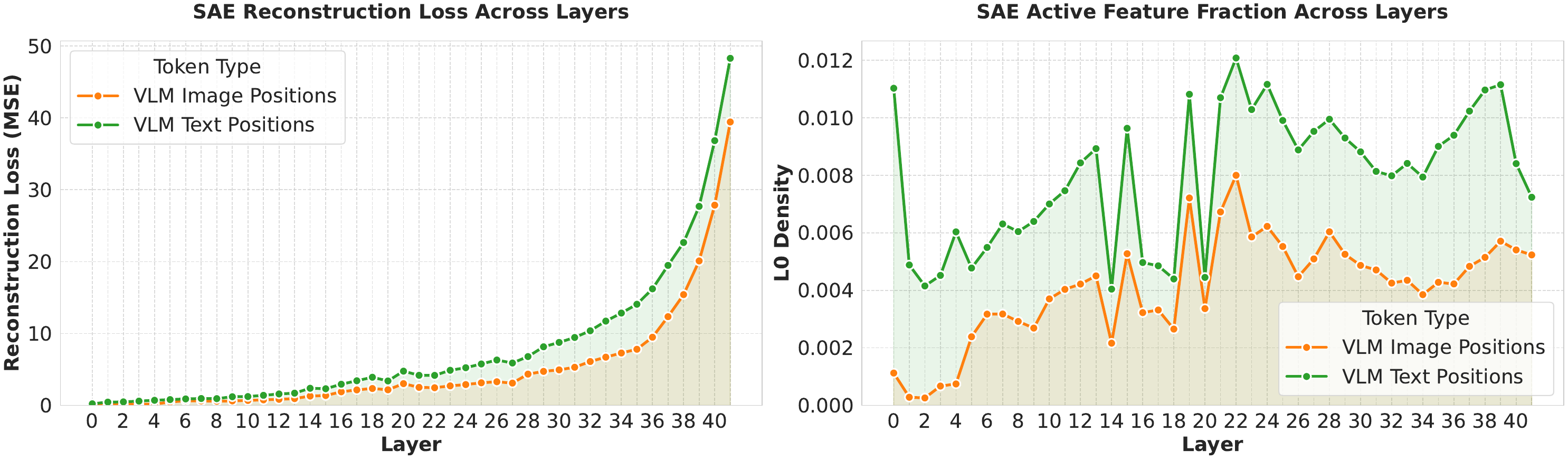}
        \caption{JEPA with Gemma-2-9B-it reconstruction with SAE constraints. }
        \label{fig:reconstr_sparsity_jepa_gemma9b_wsae}
    \end{subfigure}
\caption{JEPA with Gemma-2-9B-it reconstruction and sparsity.}
\end{figure*}

\clearpage
\section{Active Visual Tokens across Layers}
\label{appendix:active_vtokens}

\begin{figure}[!h]
    \centering
    
    \begin{subfigure}[b]{0.45\textwidth}
        \centering
        \includegraphics[width=\textwidth]{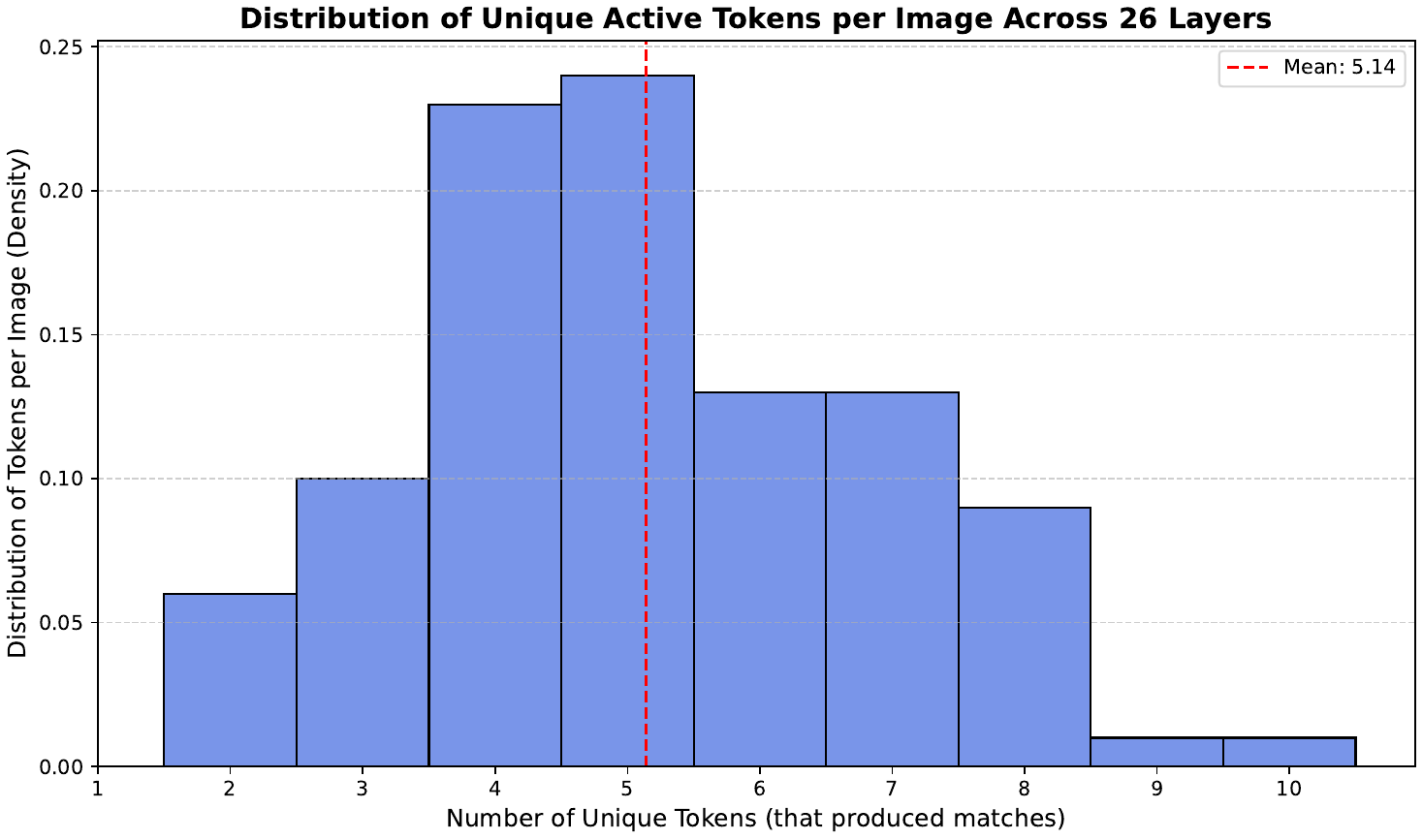}
        \caption{CLIP average activating tokens across layers with Gemma-2-2B-it. }
        \label{fig:steering_orig}
    \end{subfigure}
    \begin{subfigure}[b]{0.45\textwidth}
        \centering
        \includegraphics[width=\textwidth]{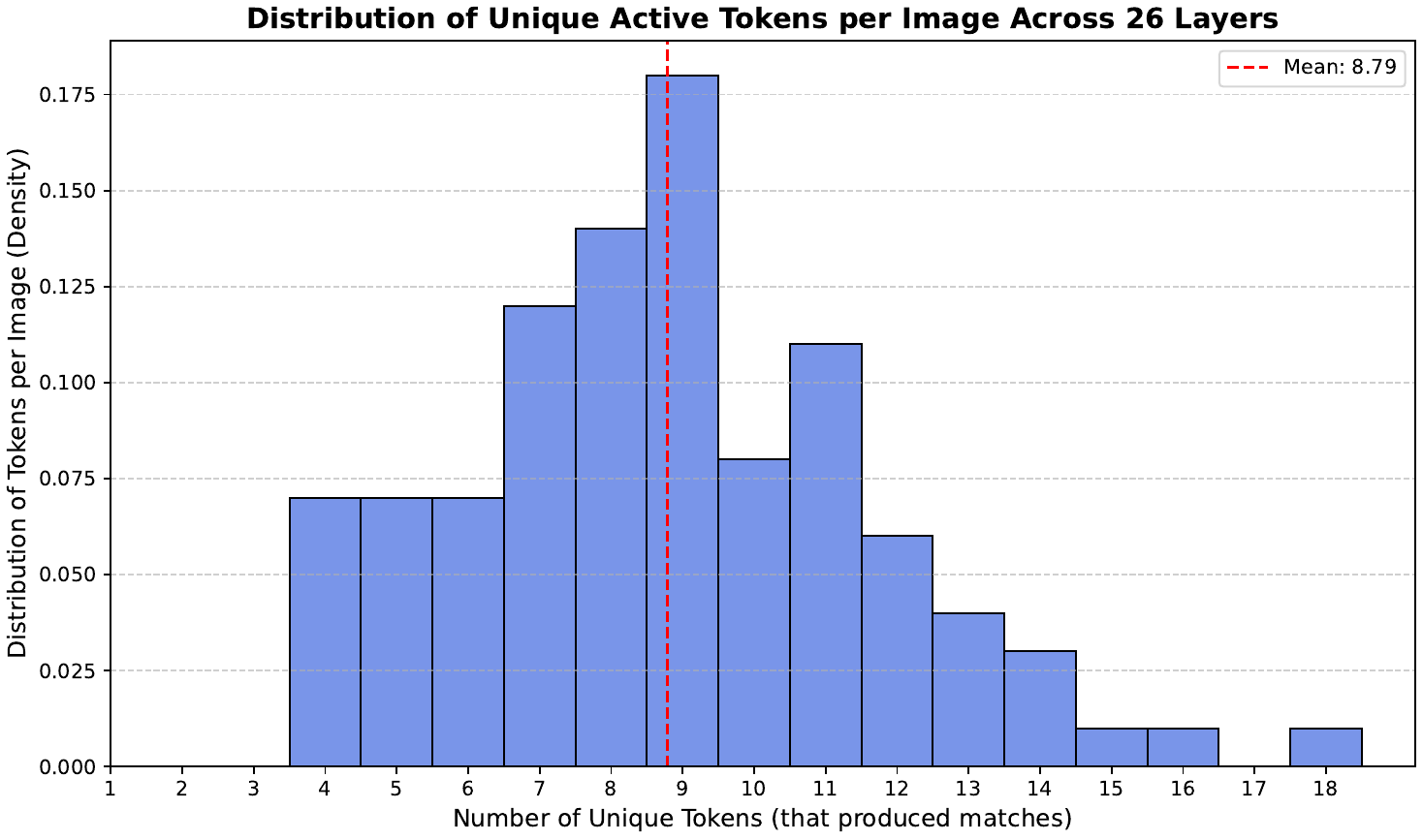}
        \caption{DINOv2 average activating tokens across layers with Gemma-2-2B-it. }
        \label{fig:steering_steered}
    \end{subfigure}
    
    \begin{subfigure}[b]{0.45\textwidth}
        \centering
        \includegraphics[width=\textwidth]{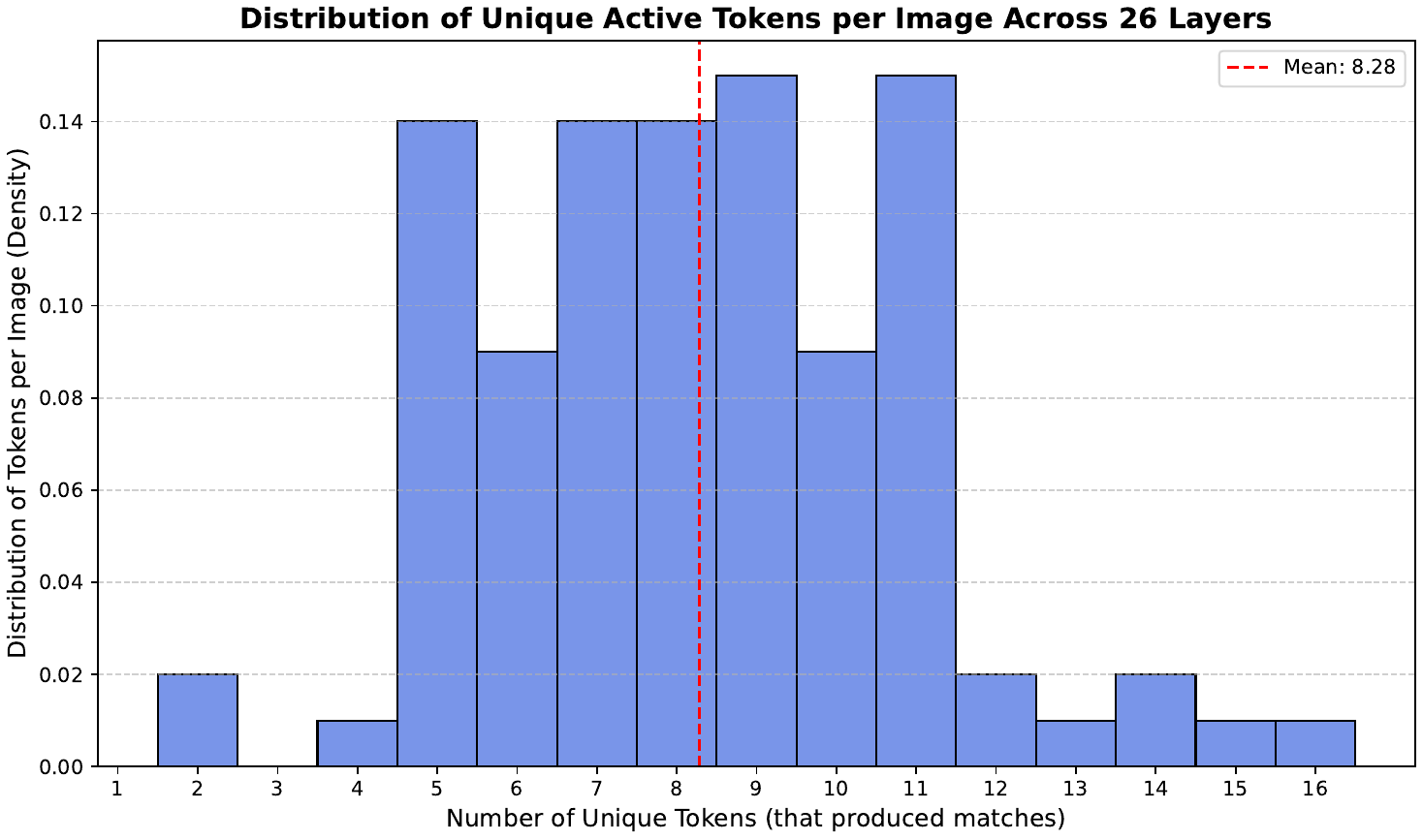}
        \caption{I-JEPA average activating tokens across layers with Gemma-2-2B-it. }
        \label{fig:steering_steered}
    \end{subfigure}
    \label{fig:avg_act_tokens_clip}

\caption{}
\label{fig:avg_act_tokens_all}
\end{figure}

We analyse for each image whether maximally activating tokens are different or not across layers. We noticed that using CLIP visual model the number of active visual features across layers is very small, with the mean around 5. On the contrary, for I-JEPA and DINOv2 we observe that the mean of the activating tokens across layers for an image is on average around 8 and 9 respectively meaning that with I-JEPA and DINOv2 more concepts per image are activated. This is shown in \cref{fig:avg_act_tokens_all}. 

\clearpage
\section{Qualitative Examples of Visual Steering with CLIP and DINOv2}
\label{appendix:qualitative_steering}


\begin{figure*}[!h]
    \centering
    \begin{subfigure}[t]{0.24\textwidth}
        \centering
        \includegraphics[width=\linewidth]{figures/cat_coach.jpg}
        \vspace{2pt}
        \scriptsize
        \textbf{Q:} What is shown in this image? \\
        \textbf{Original:} The image shows a cat sitting on a red and white striped curtain. The cat is looking at the camera. \\
        \textbf{Remove ``cat'':} The image shows a cat sitting on a red and white striped curtain. The cat is looking at the camera. \\
        \textbf{Replace with ``dog'':} The image shows a cat sitting on a red and white striped curtain. The cat is looking at the camera.
    \end{subfigure}
    \hfill
    \begin{subfigure}[t]{0.24\textwidth}
        \centering
        \includegraphics[width=\linewidth]{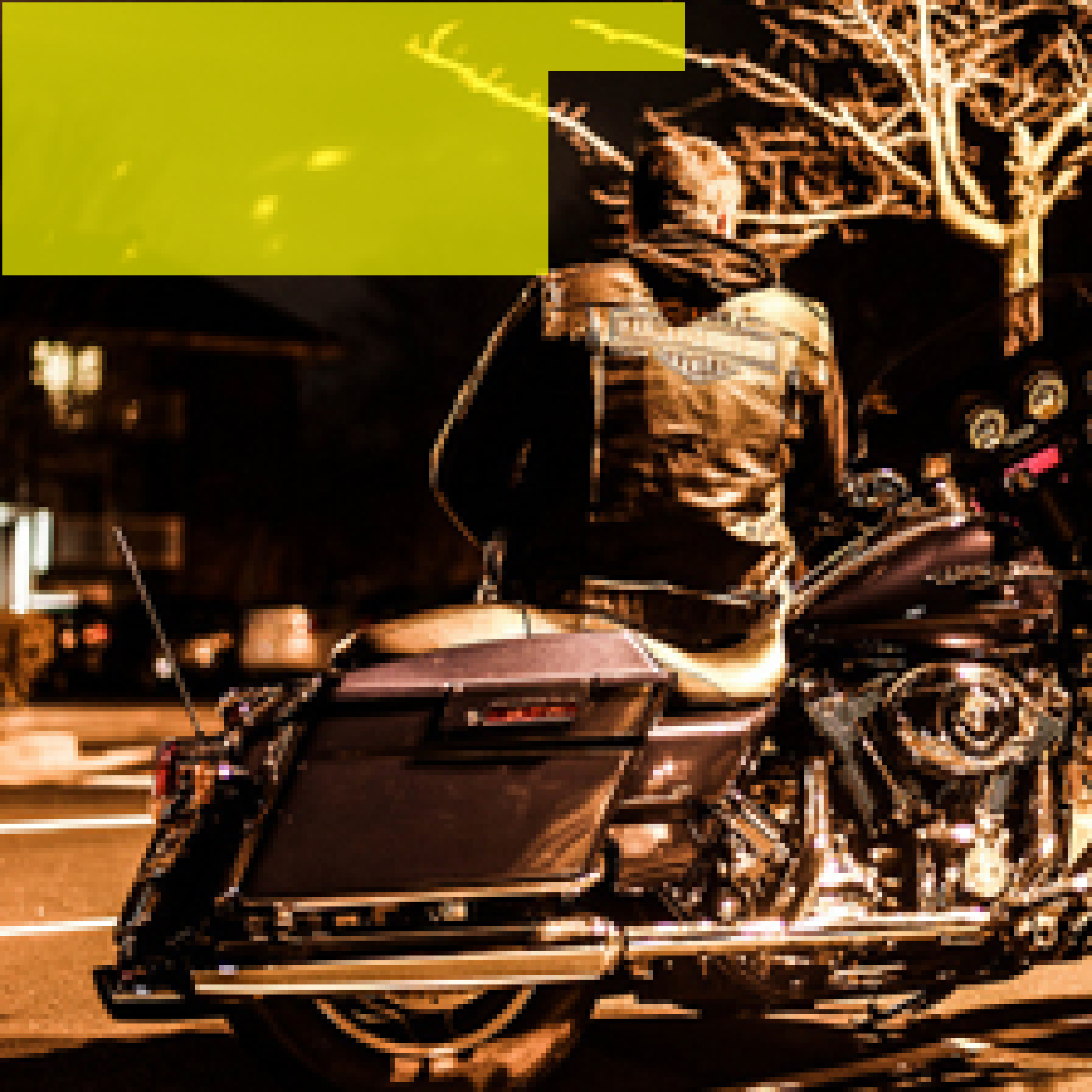}
        \vspace{2pt}
        \scriptsize
        \textbf{Q:} Describe the time of the day in the image. \\
        \textbf{Original:} The image shows a motorcycle parked on the side of the road at night.. \\
        \textbf{Remove ``night'':} The image shows a motorcycle parked on the side of the road at night. \\
        \textbf{Replace with ``morning'':} The image shows a motorcycle parked on the side of the road at night.

    \end{subfigure}
    \hfill
    \begin{subfigure}[t]{0.24\textwidth}
        \centering
        \includegraphics[width=\linewidth]{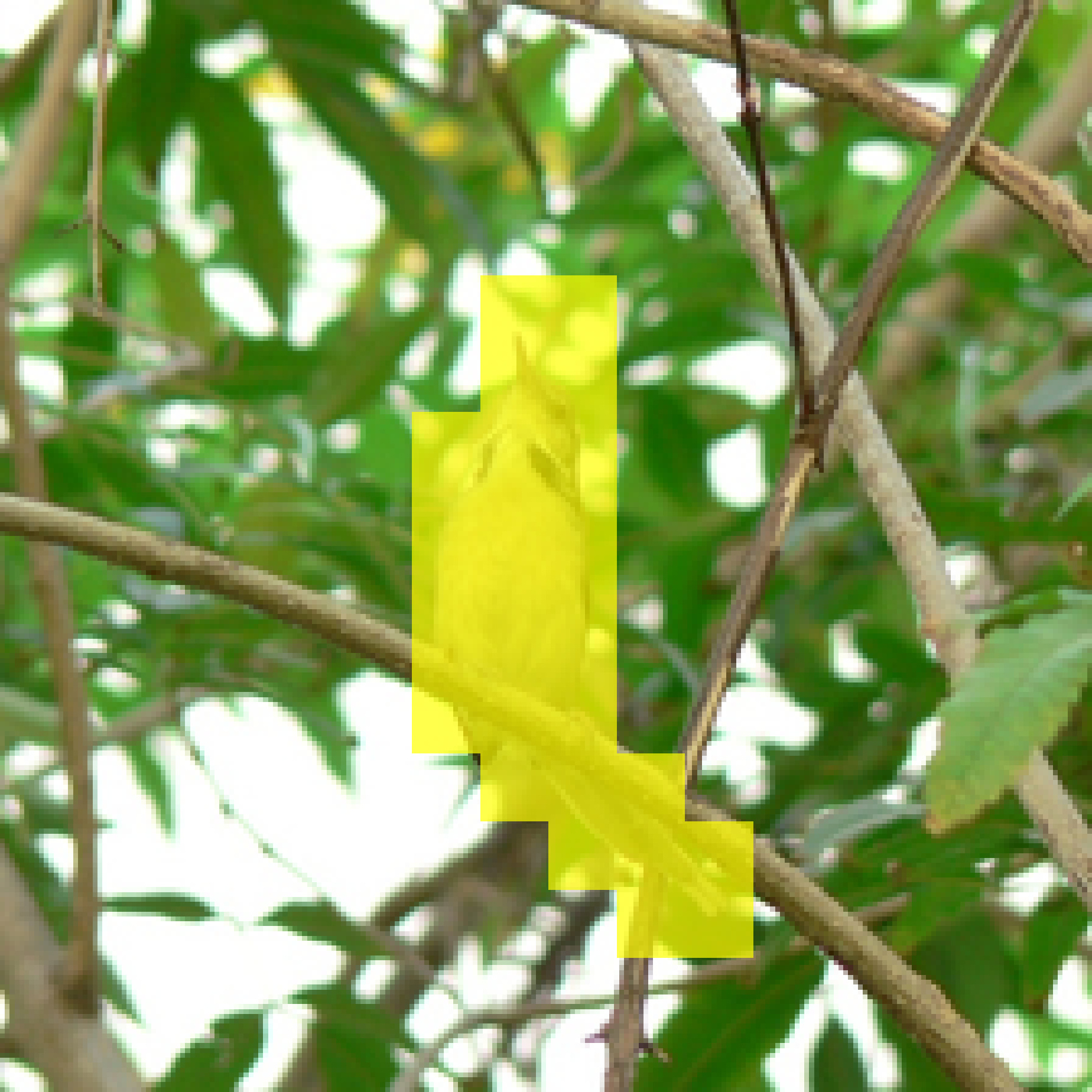}
        \vspace{2pt}
        \scriptsize
        \textbf{Q:} What is shown in this image? \\
        \textbf{Original:} Bird is sitting on a branch. Bird is a green color.\\
        \textbf{Remove ``bird'':} Bird is sitting on a branch. Bird is a green color. \\
        \textbf{Replace with ``cat'':} Bird is sitting on a branch. Bird is a green color.
    \end{subfigure}
    \hfill
    \begin{subfigure}[t]{0.24\textwidth}
        \centering
        \includegraphics[width=\linewidth]{figures/girl_eating_cake.jpg}
        \vspace{2pt}
        \scriptsize
        \textbf{Q:} What is shown in this image? \\
        \textbf{Original:} A girl is eating a donut. She is sitting on a chair and has a big smile on her face. \\
        \textbf{Remove ``chocolate'':} The image shows a little girl with a big smile and a big cupcake. She is sitting on a chair and eating a cupcake. \\
        \textbf{Replace with ``bread'':} The image shows a girl with a big smile and a little girl with a big smile. The girl is eating a donut.
    \end{subfigure}
    \caption{Qualitative example of visual steering with CLIP visual encoder and Gemma-2-2B-it LLM, selecting precise patches in the image and changing model's understanding of what is contained in them. Local steering with CLIP visual encoder is not effective. }
    \label{fig:steering_example_clip}
\end{figure*}

\begin{figure*}[!h]
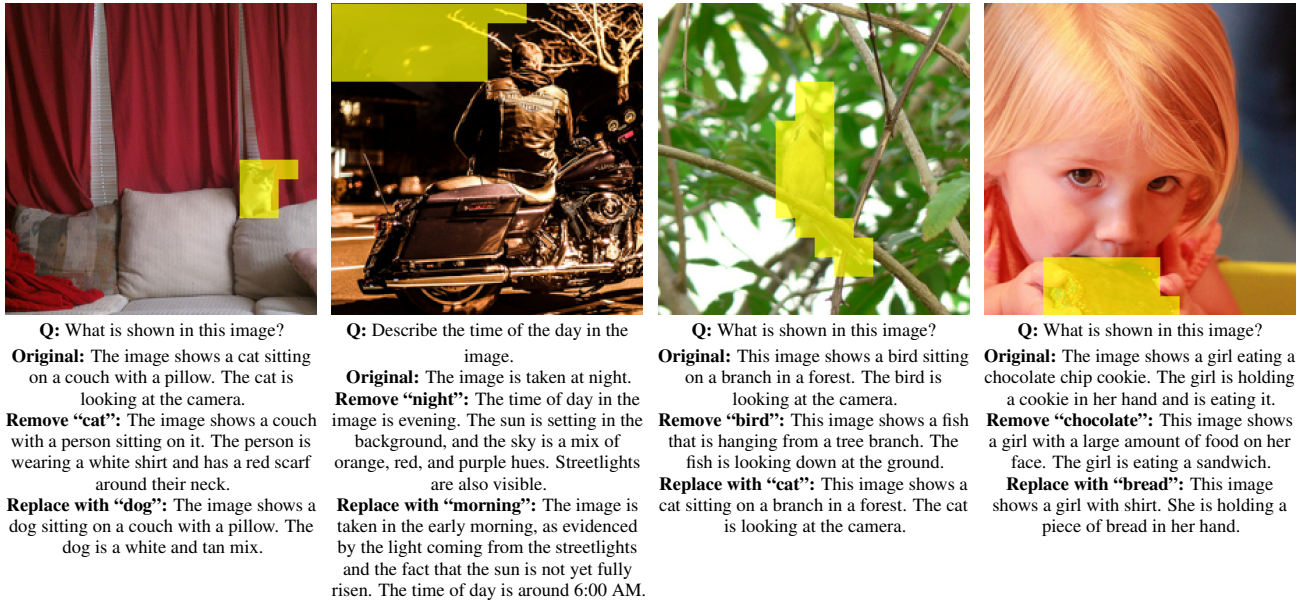

    \centering
    \begin{subfigure}[t]{0.24\textwidth}
        \centering
        \includegraphics[width=\linewidth]{figures/cat_coach.jpg}
        \vspace{2pt}
        \scriptsize
        \textbf{Q:} What is shown in this image? \\
        \textbf{Original:} The image shows a cat sitting on a couch with a pillow. The cat is looking at the camera. \\
        \textbf{Remove ``cat'':} The image shows a couch with a person sitting on it. The person is wearing a white shirt and has a red scarf around their neck. \\
        \textbf{Replace with ``dog'':} The image shows a dog sitting on a couch with a pillow. The dog is a white and tan mix.
    \end{subfigure}
    \hfill
    \begin{subfigure}[t]{0.24\textwidth}
        \centering
        \includegraphics[width=\linewidth]{figures/motorbike_night_man.jpg}
        \vspace{2pt}
        \scriptsize
        \textbf{Q:} Describe the time of the day in the image. \\
        \textbf{Original:} The image is taken at night. \\
        \textbf{Remove ``night'':} The time of day in the image is evening. The sun is setting in the background, and the sky is a mix of orange, red, and purple hues. Streetlights are also visible. \\
        \textbf{Replace with ``morning'':} The image is taken in the early morning, as evidenced by the light coming from the  streetlights  and the fact that the sun is not yet fully risen. The time of day is around 6:00 AM.

    \end{subfigure}
    \hfill
    \begin{subfigure}[t]{0.24\textwidth}
        \centering
        \includegraphics[width=\linewidth]{figures/bird_tree.jpg}
        \vspace{2pt}
        \scriptsize
        \textbf{Q:} What is shown in this image? \\
        \textbf{Original:} This image shows a bird sitting on a branch in a forest. The bird is looking at the camera.\\
        \textbf{Remove ``bird'':} This image shows a fish that is hanging from a tree branch. The fish is looking down at the  ground. \\
        \textbf{Replace with ``cat'':} This image shows a cat sitting on a branch  in a forest. The cat is looking at the camera.
    \end{subfigure}
    \hfill
    \begin{subfigure}[t]{0.24\textwidth}
        \centering
        \includegraphics[width=\linewidth]{figures/girl_eating_cake.jpg}
        \vspace{2pt}
        \scriptsize
        \textbf{Q:} What is shown in this image? \\
        \textbf{Original:} The image shows a girl eating a chocolate chip cookie. The girl is holding a cookie in her hand and is eating it. \\
        \textbf{Remove ``chocolate'':} This image shows a girl with a large amount of
 food on her face. The girl is eating a sandwich. \\
        \textbf{Replace with ``bread'':} This image shows a girl with shirt. She is holding a piece of bread in her hand.
    \end{subfigure}
    \label{fig:steering_example_dinov2_appendix}
    \caption{Qualitative example of visual steering with DINOv2 visual encoder and Gemma-2-2B-it LLM, selecting precise patches in the image and changing model’s understanding of what is contained in them. Local steering with DINOv2 visual encoder is effective. }
\end{figure*}

\section{Performance - Interpretability Trade-off}
\label{appendix:perf_interp}

We compare how additional constraints on SAE layers affect both interpretability and benchmark performance. To aggregate interpretability metrics, we averaged the matching rate across all evaluated layers. For performance, we standardized all benchmarks to a 0-100 percentage scale. While GQA, POPE, and LLaVA-Bench natively occupy this range, the MME total score is normalized against its theoretical maximum of 2800 points. Furthermore, we select the POPE F1-score over accuracy to robustly penalize hallucination bias. The final score is reported as the arithmetic mean of these standardized values. We use DINOv2 with Gemma-2-2B-it for this ablation as it is our primary configuration throughout the paper. Running the full SAE-layer sweep across every encoder - LLM combination and across all layer subsets of the 26-layer model is computationally prohibitive, so we restrict our analysis to representative subsets rather than an exhaustive search.

\begin{figure}[!h]
    \centering
    \includegraphics[width=.75\linewidth]{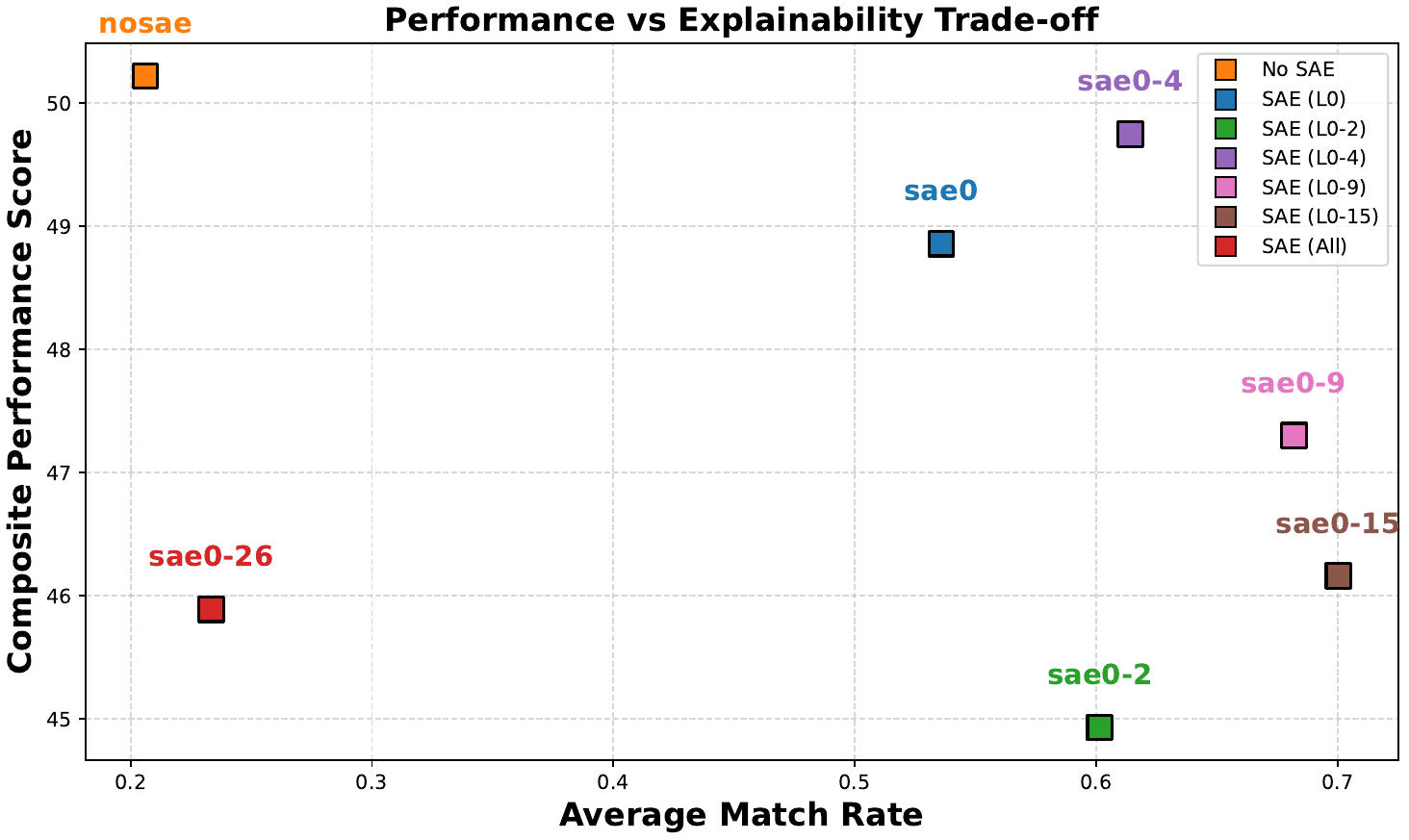}
    \caption{Performance-Interpretability Trade-off. Increasing the number of SAE-constrained layers improves interpretability up to a threshold. The closest performance to not including any SAE constraints is achieved including the SAEs from layer 0 to 4. }
    \label{fig:perf_interp_tradeoff}
\end{figure}

\begin{figure}[!h]
    \centering
    \includegraphics[width=.75\linewidth]{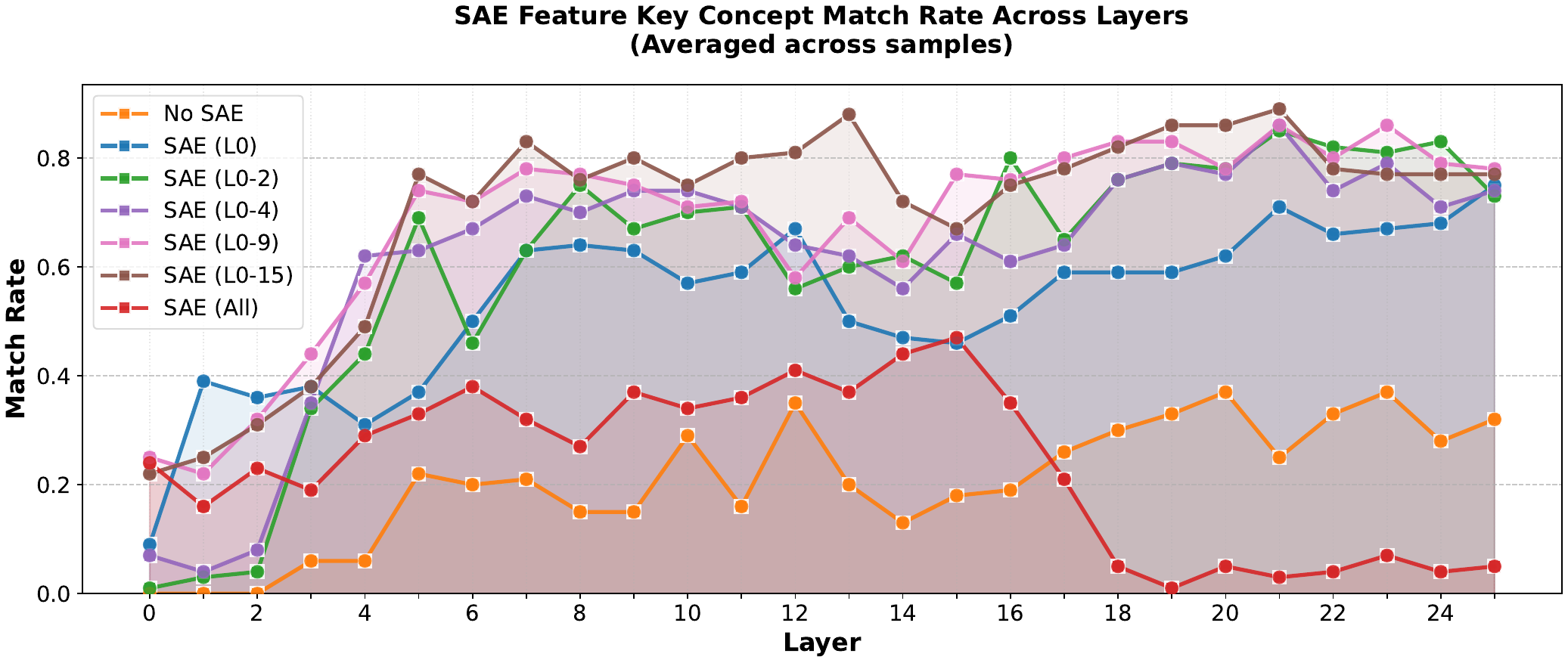}
    \caption{Matching Rate across layers for different SAE constraints. Increasing the number of SAE-constrained layers improves interpretability up to a threshold. Including all layers for SAE constraints reduces the matching rate in the last layers. }
    \label{fig:match_rate_abl}
\end{figure}

As shown in \cref{fig:perf_interp_tradeoff,fig:match_rate_abl} increasing the number of SAE-constrained layers improves interpretability up to a threshold. While applying SAEs to all layers causes a slight regression in interpretability compared to the peak, specifically reducing it in the last layers, it remains superior to the unconstrained baseline. Regarding performance, the configuration that includes SAE constraints on the first 5 layers yields results closest to the baseline (no SAE constraints) while preserving a good level of interpretability. 

Note that this trade-off is architecture-dependent; for instance, adding SAEs in the LLaMA-based model actually improves performance compared to the slight decrease seen in Gemma-2-2B, and the gap is reduced with Gemma-2-9B as seen in \cref{appendix:performance}. We therefore adopt the five-layer configuration as a representative heuristic across architectures. While a more optimal combination of SAEs likely exists for each specific architecture, our core finding that SAE constraints substantially improve interpretability with limited performance cost is robust across the configurations we tested.

\newpage

\section{Spatial Localization with Different LLM Backbones}
\label{appendix:spatial_loc_abl}

We compute the spatial localization accuracy scores across different visual encoders and LLM backbones (\cref{tab:localization_accuracy_sae_only}). DINOv2 consistently demonstrates high spatial alignment. While I-JEPA also maintains consistent alignment, its accuracy remains lower than that of DINOv2. Conversely, CLIP exhibits the lowest localization accuracy across all LLMs, with scores that fluctuate depending on the specific language backbone.

\begin{table}[ht]
\centering
\caption{Spatial Localization Accuracy Scores across different visual encoders and LLM backbones.}
\label{tab:localization_accuracy_sae_only}
\resizebox{0.35\textwidth}{!}{
\begin{tabular}{@{}llc@{}}
\toprule
\textbf{Visual Encoder} & \textbf{LLM Backbone} & \textbf{SLA} \\
\midrule
\multirow{3}{*}{DINOv2} & Gemma-2-2B & 0.927 \\
 & Gemma-2-9B & 0.918 \\
 & LLaMA-3.1-8B & 0.913 \\
\midrule
\multirow{3}{*}{CLIP} & Gemma-2-2B & 0.201 \\
 & Gemma-2-9B & 0.642 \\
 & LLaMA-3.1-8B & 0.261 \\
\midrule
\multirow{3}{*}{I-JEPA} & Gemma-2-2B & 0.702 \\
 & Gemma-2-9B & 0.717 \\
 & LLaMA-3.1-8B & 0.770 \\
\bottomrule
\end{tabular}
}
\end{table}




\section{Steering with Different LLM Backbones}
\label{appendix:steering_backbones_abl}

We evaluate the robustness of the VISTA steering mechanism across three distinct LLMs for all visual encoders. As shown in \cref{tab:llm_ablation_vista_allencoders}, steering successfully works across different LLM backbones for DINOv2 model with similar Mean Score result. For the other two visual encoders I-JEPA generally performs better than CLIP for all LLMs as it maintains a higher localization accuracy. 

\begin{table*}[!h]
\centering
\caption{Quantitative ablation analysis of steering with and without SAEs across DINOv2, CLIP, and I-JEPA visual encoders using Gemma-2-2B, Gemma-2-9B, and LLaMA-3.1-8B backbones.}
\label{tab:llm_ablation_vista_allencoders}
\resizebox{0.6\textwidth}{!}{
\begin{tabular}{@{}ll cccc c cccc@{}}
\toprule
\multirow{2.5}{*}{\textbf{Method}} & \multirow{2.5}{*}{\textbf{\begin{tabular}[l]{@{}l@{}}LLM\\Backbone\end{tabular}}} & \multicolumn{4}{c}{\textbf{Remove Operation}} & & \multicolumn{4}{c}{\textbf{Replace Operation}} \\ 
\cmidrule(lr){3-6} \cmidrule(lr){8-11}
 & & \textbf{S2} & \textbf{S1} & \textbf{S0} & \textbf{MS} & & \textbf{S2} & \textbf{S1} & \textbf{S0} & \textbf{MS} \\ 
\midrule
\multicolumn{11}{c}{\textbf{Baseline (No SAE)}} \\
\midrule
\multirow{3}{*}{DINOv2} & Gemma-2-2B & 10 & 2 & 88 & 0.22 & & 3 & 0 & 97 & 0.06 \\
 & Gemma-2-9B & 11 & 0 & 89 & 0.22 & & 2 & 0 & 98 & 0.04 \\
 & LLaMA-3.1-8B & 0 & 0 & 100 & 0.00 & & 1 & 0 & 99 & 0.02 \\
\addlinespace
\multirow{3}{*}{CLIP} & Gemma-2-2B & 13 & 20 & 67 & 0.46 & & 16 & 6 & 78 & 0.38 \\
 & Gemma-2-9B & 5 & 2 & 93 & 0.12 & & 1 & 1 & 98 & 0.03 \\
 & LLaMA-3.1-8B & 0 & 0 & 100 & 0.00 & & 1 & 0 & 99 & 0.02 \\
\addlinespace
\multirow{3}{*}{I-JEPA} & Gemma-2-2B & 8 & 18 & 74 & 0.34 & & 20 & 9 & 71 & 0.49 \\
 & Gemma-2-9B & 8 & 11 & 81 & 0.27 & & 6 & 1 & 93 & 0.13 \\
 & LLaMA-3.1-8B & 4 & 2 & 94 & 0.10 & & 2 & 0 & 98 & 0.04 \\
\midrule
\multicolumn{11}{c}{\textbf{VISTA (With SAE)}} \\
\midrule
\multirow{3}{*}{DINOv2} & Gemma-2-2B & 47 & 49 & 4 & 1.43 & & 59 & 28 & 13 & 1.46 \\
 & Gemma-2-9B & 54 & 38 & 8 & \textbf{1.46} & & 62 & 25 & 13 & 1.49 \\
 & LLaMA-3.1-8B & 47 & 35 & 18 & 1.29 & & 63 & 30 & 7 & \textbf{1.56} \\
\addlinespace
\multirow{3}{*}{CLIP} & Gemma-2-2B & 18 & 4 & 78 & 0.40 & & 7 & 4 & 89 & 0.18 \\
 & Gemma-2-9B & 13 & 10 & 77 & 0.36 & & 24 & 14 & 62 & 0.62 \\
 & LLaMA-3.1-8B & 4 & 0 & 96 & 0.08 & & 6 & 6 & 88 & 0.18 \\
\addlinespace
\multirow{3}{*}{I-JEPA} & Gemma-2-2B & 8 & 25 & 67 & 0.41 & & 17 & 47 & 36 & 0.81 \\
 & Gemma-2-9B & 12 & 19 & 69 & 0.43 & & 28 & 35 & 37 & 0.91 \\
 & LLaMA-3.1-8B & 26 & 21 & 53 & 0.73 & & 36 & 36 & 28 & 1.08 \\
\bottomrule
\end{tabular}
}
\end{table*}

\section{Out-of-Distribution Robustness}
\label{appendix:ood_robustness}

Our training data (CC3M and LLaVA instruction tuning data) is predominantly photo-centric. To test whether VISTA's alignment generalizes beyond this distribution, we evaluated on 500 images from DomainNet \cite{peng2019moment}, spanning visually distinct domains such as clipart, sketches, paintings, and infographics. These domains are OOD from the projector's training data.

We report reconstruction error, sparsity, and matching rate on this OOD set, comparing against in-distribution results. As shown in \cref{tab:ood_recon_sparsity,tab:ood_match_rate}, all three metrics remain consistent across distributions, indicating that VISTA's cross-modal alignment is a structural property that transfers beyond the training distribution.

\begin{table}[h]
\centering
\caption{Reconstruction error and sparsity on OOD DomainNet images compared to text tokens. Metrics remain comparable, indicating visual tokens continue to inhabit the text SAE manifold under distribution shift.}
\label{tab:ood_recon_sparsity}
\begin{tabular}{@{}lcc@{}}
\toprule
\textbf{Modality} & \textbf{Reconstruction} & \textbf{Sparsity} \\
\midrule
Text & 6.92 & 0.0048 \\
Image (OOD) & 6.49 & 0.0042 \\
\bottomrule
\end{tabular}
\end{table}

\begin{table}[h]
\centering
\caption{Matching rate on in-distribution versus OOD (DomainNet) images. VISTA preserves most of its interpretability under significant distribution shift.}
\label{tab:ood_match_rate}
\begin{tabular}{@{}lc@{}}
\toprule
\textbf{Image Data} & \textbf{Match Rate} \\
\midrule
In-Distribution & 0.63 \\
OOD (DomainNet) & 0.58 \\
\bottomrule
\end{tabular}
\end{table}

We attribute this robustness to two factors. First, VISTA learns a structural mapping between two frozen representation spaces rather than memorizing training examples, so it inherits the base vision encoder's zero-shot generalization. Second, because DINOv2 already clusters semantically related images (e.g., a sketch of a dog and a photo of a dog) in nearby regions of its feature space, the linear projector preserves this clustering when mapping into the text SAE space. Consequently, concepts learned on photographic training data transfer naturally to non-photographic domains.

\section{Visual vs Text-Only Steering Comparison}
\label{appendix:visual_vs_text_steering}

To verify that VISTA's interventions modify the model's perception of visual content rather than hijacking the language generation pathway, we compare three intervention regimes on the ``Cat on a couch'' image from \cref{fig:steering_example_dinov2}, using the ``-cat +dog'' direction in all cases:

\begin{itemize}
    \item \textbf{VISTA (visual tokens steered):} The model outputs \textit{``A dog sitting on a couch with a pillow''}. Scene context is preserved because unsteered background patches remain intact.
    \item \textbf{Text-only steering, no image:} Without the image, the model outputs generic text such as \textit{``A dog running in a park''} failing to describe the actual scene (couch, pillow).
    \item \textbf{Prompt steering with image present:} Applying the ``-cat +dog'' vector to the text tokens while the image is provided causes hallucinations such as \textit{``The dog is playing in the grass with a cat next to it''}. The original concept is not removed and hallucinated content appears.
\end{itemize}

Only VISTA produces a coherent, scene-preserving modification of the perceived content. This three-way comparison provides direct causal evidence that VISTA modifies the model's perception of specific visual patches rather than steering the language pathway.

\end{document}